\documentclass{ucetd}
\usepackage{subfigure,epsfig,amsfonts}
\usepackage{amsmath}
\usepackage{amssymb}
\usepackage{amsthm}
\usepackage{booktabs}       
\usepackage{comment}
\usepackage{color}
\usepackage{empheq}
\usepackage{tikz}
\usepackage{framed}
\usepackage{siunitx}
\usepackage{pifont}
\usepackage{enumitem}
\usetikzlibrary{calc,shapes.arrows,arrows.meta,shapes.misc}
\usepackage[square,sort,comma,numbers]{natbib}
\usepackage{ifthen}
\usepackage{usebib}
\newbibfield{title} 
\newbibfield{author} 
\newbibfield{address} 
\newbibfield{conf} 
\usepackage[pagebackref=true,breaklinks=true,letterpaper=true,bookmarks=false,hidelinks]{hyperref}
\usepackage{cleveref}

\usepackage{upgreek}
\usepackage{bm}
\makeatletter
\newcommand{\bfgreek}[1]{\bm{\@nameuse{#1}}}
\makeatother

\def\ie{\emph{i.e.}}
\def\eg{\emph{e.g.}}

\def\etal{\emph{et al.}}

\def\optimizer{\texttt{optimizer}}

\def\ungamma{\widetilde\gamma}

\def\e{\mathrm{e}}

\newcommand{\cmt}[1]{\iffalse #1 \fi}
\newcommand{\cmark}{\ding{51}}%

\def\dotcup{\dot{\cup}}

\newcommand{\Cat}[1]{\mathrm{Cat}(#1)}
\newcommand{\Bern}[1]{\mathrm{Bern}(#1)}

\def\amp{\&}
\def\ndash{-}
\def\dash{-\,\,}

\DeclareMathOperator*{\argmax}{arg\,max}

\def\vx{\mathbf{x}}
\def\vy{\mathbf{y}}
\def\vz{\mathbf{z}}
\def\vg{\mathbf{g}}
\def\vb{\mathbf{b}}

\def\vp{\mathbf{p}}
\def\vq{\mathbf{q}}
\def\vt{\mathbf{t}}
\def\vW{\mathbf{W}}
\def\vX{\mathbf{X}}
\def\vZ{\mathbf{Z}}

\def\softmax{\mathrm{softmax}}

\def\DKL{D_{\mathrm{KL}}}
\def\DE{D_{\mathrm{E}}}

\def\DIS{D_{\mathrm{IS}}}

\definecolor{gray}{gray}{0.5}

\newcommand*{\pol}{\mathrm{pol}}
\newcommand*{\rot}{\mathrm{rot}}

\newcommand*{\old}{\text{(old)}}
\newcommand*{\new}{\text{(new)}}

\usepackage{color}
\definecolor{myblue}{rgb}{1, 1, .8}

\newlength\mytemplen
\newsavebox\mytempbox

\makeatletter
\newcommand\mybluebox{%
    \@ifnextchar[
       {\@mybluebox}%
       {\@mybluebox[0pt]}}

\def\@mybluebox[#1]{%
    \@ifnextchar[
       {\@@mybluebox[#1]}%
       {\@@mybluebox[#1][0pt]}}

\def\@@mybluebox[#1][#2]#3{
    \sbox\mytempbox{#3}%
    \mytemplen\ht\mytempbox
    \advance\mytemplen #1\relax
    \ht\mytempbox\mytemplen
    \mytemplen\dp\mytempbox
    \advance\mytemplen #2\relax
    \dp\mytempbox\mytemplen
    \colorbox{myblue}{\hspace{1em}\usebox{\mytempbox}\hspace{1em}}}

\makeatother

\title{Discovery of Visual Semantics by Unsupervised and Self-Supervised Representation Learning}
\author{Gustav Martin Larsson}
\department{Computer Science}
\division{Physical Sciences}
\degree{Doctor of Philosophy}
\date{August 2017}

\epigraph{Epigraph Text}

\begin{document}
\maketitle

\makecopyright


\tableofcontents
\listoffigures
\listoftables

\acknowledgments

I want to start by thanking my primary advisor, Gregory Shakhnarovich. You have
been a great mentor and teacher, attentive to both personal needs and professional
development. Above all else, you gave me what I sought the most from
an advisor: an environment where hard work is rewarded with success and
fulfillment.

I also want to thank my co-advisor and advisor for my Master's degree, Yali Amit. You and I have spent countless of hours discussing research and machine learning, which has helped me gain invaluable experience, intuition, and perspective. In this dissertation, Yali was my primary advisor for the work in \cref{sec:smm}.

I also owe a great deal to my close collaborator Michael Maire, who has been
involved in almost every aspect of this dissertation and has in many ways been
a deputy advisor to Greg. I appreciate everything that you have done for me.

Furthermore, I want to thank my entire committee, Gregory Shakhnarovich, Yali Amit, Michael Maire, and Risi Kondor. Anne Rogers, thank you for making the job of a teaching assistant fun and fulfilling. I also appreciate that I could always turn to you with problems, big and small. A final thanks to all the students, faculty, administrative and technical staff at both University of Chicago and TTI-Chicago.

\vspace{0.3cm}
\noindent
Chenan, the support that you have given me is hard to put into words and more
than I ever intended to put on your shoulders. You have accepted my problems as
ours, and given me a strength that I do not have alone. When things get tough and I despair and retreat, you are reasoned, levelheaded, and get straight to finding a solution.

Ett stort tack till mina f\"or\"aldrar Mats och Gunnel, syskon Loe och Erik, sl\"akt och v\"anner hemma i Sverige. Loe, du och David var ett ov\"arderligt st\"od i Chicago. Farfar, du var den f\"orste som uppt\"ackte och uppmuntrade mitt intresse f\"or matematik. Tack.

\abstract
The success of deep learning in computer vision is rooted in the ability of deep networks to scale up model complexity as demanded by challenging visual tasks. As complexity is increased, so is the demand for large amounts of labeled data to train the model. This is associated with a costly human annotation effort. Modern vision networks often rely on a two-stage training process to satisfy this thirst for training data: the first stage, \textit{pretraining}, is done on a general vision task where a large collection of annotated data is available. This primes the network with semantic knowledge that is general to a wide variety of vision tasks. The second stage, \textit{fine-tuning}, continues the training of the network, this time for the target task where annotations are often scarce. The reliance on supervised pretraining anchors future progress to a constant human annotation effort, especially for new or ever-changing domains. To address this concern, with the long-term goal of leveraging the abundance of cheap unlabeled data, we explore methods of unsupervised pretraining. In particular, we propose to use self-supervised automatic image colorization.

We begin by evaluating two baselines for leveraging unlabeled data for representation learning. One is based on training a mixture model for each layer in a greedy manner. We show that this method excels on relatively simple tasks in the small sample regime. It can also be used to produce a well-organized feature space that is equivariant to cyclic transformations, such as rotation. Second, we consider autoencoders, which are trained end-to-end and thus avoid the main concerns of greedy training. However, its per-pixel loss is not a good analog to perceptual similarity and the representation suffers as a consequence. Both of these methods leave a wide gap between unsupervised and supervised pretraining.

As a precursor to our improvements in unsupervised representation learning, we develop a novel method for automatic colorization of grayscale images and focus initially on its use as a graphics application. We set a new state-of-the-art that handles a wide variety of scenes and contexts. Our method makes it possible to revitalize old black-and-white photography, without requiring human effort or expertise. In order for the model to appropriately re-color a grayscale object, it must first be able to identify it. Since such high-level semantic knowledge benefits colorization, we found success employing the two-stage training process with supervised pretraining. This raises the question: If colorization and classification both benefit from the same visual semantics, can we reverse the relationship and use colorization to benefit classification?

Using colorization as a pretraining method does not require data annotations, since labeled training pairs are automatically constructed by separating intensity and color. The task is what is called \textit{self-supervised}. Colorization joins a growing family of self-supervision methods as a front-runner with state-of-the-art results. We show that up to a certain sample size, labeled data can be entirely replaced by a large collection of unlabeled data. If these techniques continue to improve, they may one day supplant supervised pretraining altogether. We provide a significant step toward this goal.

As a future direction for self-supervision, we investigate if multiple proxy tasks can be combined to improve generalization in the representation. A wide range of combination methods is explored, both offline methods that fuse or distill already trained networks, or online methods that actively train multiple tasks together. On controlled experiments, we demonstrate significant gains using both offline and online methods. However, the benefits do not translate to self-supervision pretraining, leaving the question of multi-proxy self-supervision an open and interesting problem.

\mainmatter

\chapter*{Introduction}
\addcontentsline{toc}{chapter}{Introduction}

Supervised versus unsupervised learning is a central debate in machine
learning. Supervision in computer vision typically refers to training images
that are annotated with additional information, such as class label (\eg,
``dog''). To solve a prediction task, some annotated samples are required to
define the task. \textit{Representation learning} is the act of learning
useful features from input samples and can be helpful prior to prediction training,
especially if labeled samples are scarce. This
\textit{pretraining} procedure can be either supervised or unsupervised. The choice between the two is in our
experience straightforward: Supervise \textit{if} you have the means to supervise. That is, if
you have data annotations, using them typically offers substantial improvements
over using the data as raw unlabeled data. The ``if'' is important, since it
may be costly or impractical. Having humans annotate a large set of data is
error-prone, expensive, and could lead to over-specialization
to the pretraining task.  Constantly annotating new data for an autonomous
robot, as the environment changes, can be impractical and limit its autonomy.
The prospect of an unsupervised method that is able to leverage the abundance of raw sensory data would
allow vision systems to seamlessly adapt to new situations and promote progress
in new domains.  We argue that this is a worthwhile pursuit. In this
dissertation, we investigate options for unsupervised representation learning
and present promising progress toward the goal of reducing the reliance on
labeled data.

Taking a step back, two of the primary sub-areas of machine learning are supervised and unsupervised learning.
The former, \textit{supervised learning}, takes training data of
inputs and outputs, with the goal of learning a model that can correctly
predict outputs given inputs. The loss driving the learning is its ability to
correctly predict outputs, or perform a closely related surrogate task.  The latter,
\textit{unsupervised learning}, has a less clearly defined goal and can be used
for several reasons. Given input samples, it may be used to find interesting
structure in the data, to generate synthetic samples, or to learn a function that
maps the data into a useful feature space that hopefully benefits downstream
supervised learning. The last goal can also be achieved through supervised learning itself, by
training on a supervised task similar to the downstream task
(\textit{transfer learning}).  The performance gap between unsupervised and
supervised representation learning has in computer vision tilted in favor of
supervised methods, and the gap has become increasingly wide as networks and
sample sizes have grown.

However, a new family of methods, \textit{self-supervision}, is reversing this
trend of a widening gap by greatly improving representation learning on
unlabeled data. The technique involves manufacturing a supervised prediction
task on raw data, similar to playing a game with yourself. For instance, an
agent (\eg, child) picks up a ball and places it on a slanted surface,
guessing its rolling trajectory. The correct outcome is subsequently observed
and if the prediction was wrong the agent can learn from its mistake.

\Cref{chp:unsupervised} discusses two distinct methods of traditional unsupervised learning.
Both methods build multi-layer feed-forward convolutional networks, which has
become standard practice in computer vision. The main difference between the two methods lies in the
way they are trained: one layer at a time (greedy) or all layers together (end-to-end). The former
method, which is the primary contribution of the chapter, works well on
small-scale problems and can cleverly deal with latent transformations, such as
rotations. However, neither of the two methods succeed in providing significant
benefits in the increasingly complex challenges in computer vision.

In \cref{chp:colorization}, we take a break from representation learning and
focus on automatic colorization: a technique to bring color to old black-and-white photographs.
With the assumption that colorization requires semantic knowledge, we build a
new algorithm inspired by work in semantic segmentation. We accommodate the
multimodal nature of color by predicting distributions over a color space. This
rich prediction representation captures the notion that a grayscale image can
be associated with several plausible colorizations and can be used in
post-processing. We also present a novel memory-efficient training method using
sparsely sampled color targets, with applications beyond colorization to any
image-to-image prediction task. Finally, we establish new quantitative
colorization benchmarks to promote further progress.

The broader implications of automatic colorization become clear in
\cref{chp:selfsup}, where we show it to be a strong
self-supervised proxy task for representation learning. The difficulty
and importance of high-level semantics in colorization drives a general-purpose
representation that can be used for other visual tasks, such as classification
and semantic segmentation. We demonstrate state-of-the-art results on both when
supervised pretraining is forgone. A series of stress tests on supervised
pretraining (smaller sample size, reduced label space, label noise, etc.)
sheds further light on the similarities between supervised and self-supervised
learning.

In \cref{chp:multiproxy}, we explore the possibility of using multiple proxy
tasks as a future direction.  We evaluate both \textit{offline} and
\textit{online} methods.  Offline is what we call methods that combine already
trained networks without being aware of how the networks were originally
trained. We evaluate network concatenation and a novel method of
task-agnostic model distillation. Online is a method that actively trains
multiple proxy tasks jointly. We identify key issues that may occur during joint
training, such as unbalanced gradients, and propose effective solutions.
Through a variety of controlled experiments based on supervised
pretraining, we show success for both offline and online methods.  Despite
promising results, our techniques interestingly do not move the needle in the
context of self-supervision. However, we hope that our investigation lays
the groundwork for future work in making self-supervision compete with
supervised pretraining.

\setcounter{chapter}{0}
\chapter{Unsupervised Representation Learning}
\label{chp:unsupervised}

We evaluate two traditional methods for unsupervised representation learning,
establishing baselines for future chapters and showing how difficult it is
to leverage unlabeled data.

\section{Introduction}

The main contribution of this chapter is a greedy method of training a
multi-layer network presented in \cref{sec:smm}. The method is based on
representing the data distribution of image patches or intermediate feature patches as
mixture models. Training is driven by trying to maximize the likelihood of
unlabeled training data under this model. Once a network has been trained, a
modest amount of labeled data can be used at the top layer to solve visual
tasks. We show competitive results on simple tasks, such as variations on
MNIST. The feature space can also be organized to be equivariant to cyclic
transformations, which we demonstrate by successfully training a model on MNIST
with random rotations.  However, due to its greedy training, it fails to
scale up to the size required by modern computer vision architectures.

To address the concerns of greedy training, we evaluate in
\cref{sec:autoencoder} autoencoders, an unsupervised model trained end-to-end.
This allows top layers to signal corrections they would like to see in lower
layers, using back-propagation of errors. The model is phrased as a compression
problem with an encoder and a decoder, where the encoder can subsequently be used
as a feature extractor. A good representation intended to solve high-level
tasks should be invariant to low-level semantics and help organize high-level
semantics into features. The reconstruction loss, operating on a per-pixel
level, encourages the opposite by putting a large price on perceptually
insignificant differences.

The purpose of this chapter is primarily to establish a baseline and give
context to self-supervised learning by describing two predecessors.
Additionally, we offer the following contributions:

\begin{itemize}
    \item A method to build multi-layer feature extractors using mixture
        models. It is trained greedily and can be made equivariant to cyclic
        transformations. It works particularly well in the small-sample regime.
    \item Shortcut Autoencoder, a novel technique for efficient training of deep
        autoencoders.
\end{itemize}

\section{Stacked Mixture Models}
\label{sec:smm}

We present a framework for building multi-layer networks, with the option of
an organized feature space equivariant to rotation or other cyclic
transformations. The work is an extension of the parts-based model developed by
Bernstein \& Amit~\cite{bernstein2005part}. On the surface, it resembles a
convolutional neural network (CNN), with several layers and weight sharing
across spatial translations. However, training is layer-wise and unsupervised.
This means, it uses no back-propagation and class-annotated data is only needed
at the final classification layer. Each layer is based on a mixture model trained on
unlabeled data. We demonstrate that this model works well on simple tasks, such
as variations on MNIST, and is particularly competitive when annotated
data is scarce.




\subsection{Related Work}

Our starting point is the \textit{parts model} by Bernstein \&
Amit~\cite{bernstein2005part}, using mixture models to learn unsupervised
visual representations. Coates~\etal~\cite{coates2010analysis} similarly use
$k$-means clustering to learn dictionaries of parts. At test time, both of
these models are analogous to the convolutional filters in a CNN. However,
training is both greedy and unsupervised, making it fast and cheap to estimate
parameters. In this section, we outline similar unsupervised representation
learning methods, with a particular focus on its contemporary methods from 2014,
when our work was developed.

The Restricted Boltzmann Machine (RBM)~\cite{smolensky1986information} is an
undirected graphical model forming a bipartite graph between visible and hidden
binary units.  The bipartite structure, which makes it a \textit{restricted}
version of a Boltzmann machine~\cite{hinton1986learning}, ensures that visible
units are conditionally independent of hidden units and can be sampled directly;
the same is true reversing the roles of hidden and visible units. Inference of
the joint probability between hidden and visible units is slightly harder, but
can be done using repeated Gibbs sampling.  The process involves alternating
between sampling visible units conditioned on hidden, and hidden units
conditioned on visible, until convergence. The models are trained using
gradient methods on the log-likelihood. To do this properly, for
each iteration of parameter updates, an inner loop of Gibbs sampling is
required. However, it has been shown that training can be successfully
performed even when the Gibbs sampling has not converged. This technique is
called Contrastive Divergence~\cite{hinton2002cd}.

There are two primary multi-layer extensions of the RBM. The first one is called Deep
Belief Networks (DBN)~\cite{hinton2006reducing}. After training an initial RBM, a
second one is trained with the hidden layers of the first one now as visible
units of the next. This creates a series of RBMs trained sequentially in a
greedy fashion. The only layers in the finished model that form an RBM are the
two top layers, since after those two layers units are sampled conditioned on the
hidden units using directed graphical models. The finished model can be
fine-tuned end-to-end, by effectively turning it into an autoencoder with tied
weights~\cite{hinton2006reducing}. The DBN can also be made into a
convolutional DBN~\cite{lee2011unsupervised}.
The second multi-layer extension is the Deep Boltzmann Machine (DBM)~\cite{salakhutdinov2009deep},
which preserves the undirectedness in each layer, making inference more
complicated and computationally more intensive. This is often alleviated
by using similar methods of training a DBN as initial pretraining, followed by
sampling strategies.

The standard way of using an RBM (or DBN/DBM) for classification is to
use the top-layer's hidden units as features fed to a linear classifier. There are
also ways to make the discriminative training part of the RBM
training~\cite{larochelle2008classification}. Fine-tuning discriminatively is
also an option, effectively turning the RBM into a feed-forward network. The point of
the RBM training is in such case to pretrain the network.

\begin{figure}
    \begin{center}
    \begin{minipage}[b]{0.19\linewidth}
        \begin{center}
            \includegraphics[width=\textwidth]{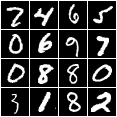} \\
            \emph{basic}
        \end{center}
    \end{minipage}
    \hfill
    \begin{minipage}[b]{0.19\linewidth}
        \begin{center}
            \includegraphics[width=\textwidth]{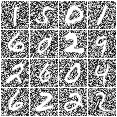} \\
            \emph{bg-rand}
        \end{center}
    \end{minipage}
    \hfill
    \begin{minipage}[b]{0.19\linewidth}
        \begin{center}
            \includegraphics[width=\textwidth]{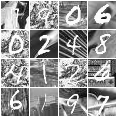} \\
            \emph{bg-img}
        \end{center}
    \end{minipage}
    \hfill
    \begin{minipage}[b]{0.19\linewidth}
        \begin{center}
            \includegraphics[width=\textwidth]{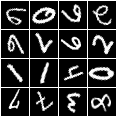} \\
            \emph{rot}
        \end{center}
    \end{minipage}
    \hfill
    \begin{minipage}[b]{0.19\linewidth}
        \begin{center}
            \includegraphics[width=\textwidth]{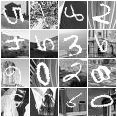} \\
            \emph{bg-img-rot}
        \end{center}
    \end{minipage}
    \end{center}
    \caption[Variations of MNIST examples]{
        \textbf{Variations of MNIST examples.}
        The variations include added background images or noise and random rotations. All variants are
        divided into 10,000 training, 2,000 validation, and 60,000 testing samples.
    }
    \label{fig:mnist-variations-examples}
\end{figure}

There are many more flavors of RBMs. We discuss briefly a few that relate
specifically to our test benchmarks.
Nair \& Hinton proposed something reminiscent of mixture models for
RBMs~\cite{nair2009implicit}, with separate sets of parameters for each mixture
component. This work was extended by Sohn~\etal~\cite{sohn2013pgbm} in
Point-wise Gated Boltzmann Machines (PGBM), by allowing the mixture components
to be specified per pixel.  This is realized by having a binary gate for each
pixel that determines which component is active. This framework can be used to
learn task-relevant and task-irrelevant features, which is particularly suited
for MNIST with cluttered background (see \cref{fig:mnist-variations-examples}). The
task-irrelevant features can subsequently be discarded by only feeding the
task-relevant activations into a classifier, making the model more robust to
noise.

The Transformation-Invariant RBM (TIRBM)~\cite{sohn2012learning} can learn
invariance to transformations such as rotation, scale, and translation. It does
this by introducing a vector of hidden units indicating transformation and
regularized to be one-hot. Each element is associated with a pre-defined linear
transformation matrix that operates on the input samples before fed into the
regular RBM weights. The model is a single-layer RBM with no generalization
to multiple layers.

We also compare results to the Stacked Denoising Autoencoder
(SDAE)~\cite{vincent2010stacked}. It is also a greedy and unsupervised model,
that for each layer trains a single-layer autoencoder. It fixes the encoder as the
first layer and continues to train on the subsequent layer. It injects noise
into the input, while keeping the clean sample as the reconstruction target.
Today, it would be more common to train the autoencoder end-to-end and use
dropout~\cite{srivastava2014dropout} as a regularizing noise injection. We
evaluate this approach in \cref{sec:autoencoder}.

Since our work in 2014, there have been several attempts at building features
equivariant to rotations using CNNs trained
end-to-end~\cite{cheng2016rifd,dieleman2016exploiting,cohen2016group,worrall2017harmonic,zhou2017oriented}.
Although not presented in our tables, this has further improved
state-of-the-art on benchmarks with randomly rotated samples. However, these
methods are fully supervised and are not relevant to unsupervised
representation learning, unlike the RBM-based methods.

\subsection{Parts Model}
\begin{figure}
    \begin{center}
    \begin{minipage}[b]{0.32\linewidth}
        \begin{center}
            \includegraphics[width=\textwidth]{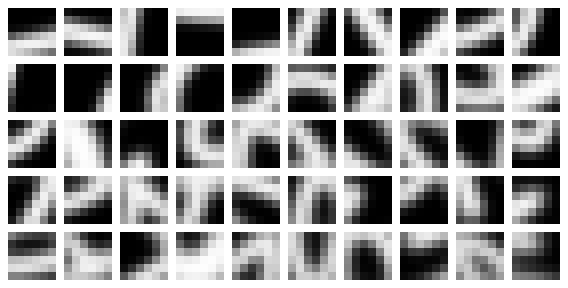} \\
            \emph{basic}
        \end{center}
    \end{minipage}
    \hfill
    \begin{minipage}[b]{0.32\linewidth}
        \begin{center}
            \includegraphics[width=\textwidth]{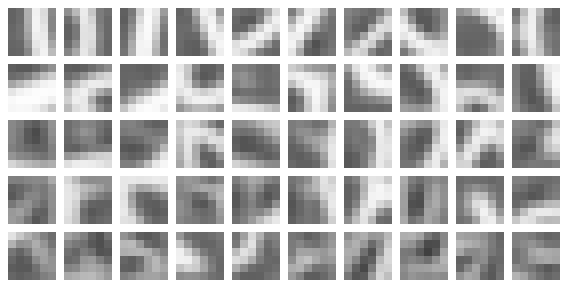} \\
            \emph{bg-rand}
        \end{center}
    \end{minipage}
    \hfill
    \begin{minipage}[b]{0.32\linewidth}
        \begin{center}
            \includegraphics[width=\textwidth]{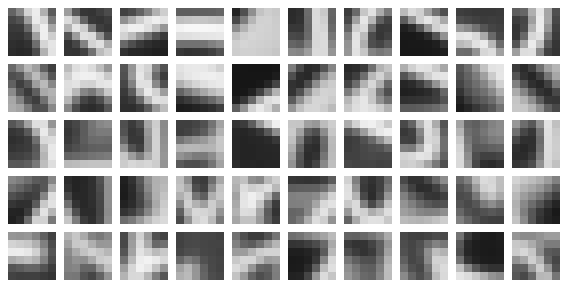} \\
            \emph{bg-img}
        \end{center}
    \end{minipage}
    \end{center}
    \caption[Rotatable parts on MNIST]{
        \textbf{Rotatable parts on MNIST.}
        Examples of 6-by-6 convolutional parts learned using a mixture models with latent rotation on variations of MNIST (see \cref{fig:mnist-variations-examples}). Only the canonical rotation (that happens to have the leading index) is displayed for each part.
    }
    \label{fig:parts-mnist}
\end{figure}

We begin by reviewing the \textit{parts model} by Bernstein \&
Amit~\cite{bernstein2005part}, presenting it in slightly more general terms.
The final model is a feed-forward network composed of several layers:
\[
    f = f_L \circ \dots \circ f_1
\]
The model is trained in a greedy manner, which means $f_1$ is trained first
without regard for any other layer, then $f_2$ without changing $f_1$, and so forth. To formulate
$f_l$, we associate each layer with a stationary (translation-invariant)
distribution over a small spatial window. If, for instance, $\mathrm{dom}(f_l)
\subset \mathbb{R}^{H \times W \times F}$ (height, width, feature channels),
then the distribution describes how to draw samples from $\mathbb{R}^{S \times
S \times F}$ representing any spatial region of the incoming representation. We
model the distribution over each patch as a mixture model.  The latent mixture
component can be thought of as defining what type the patch represents from a
dictionary of parts. At low levels, a part can represent a specific type of
edge (see \cref{fig:parts-mnist}) and at higher levels it could be part of
a semantic object.  This is different from the distributed representation of an
RBM, where each hidden unit represents only one of several active aspects of the visible units. Each
mixture component can be modeled with a variety of distributions. Although the
framework is general, we focus primarily on products of Bernoulli
distributions.

The distribution allows us to sample patches of intermediate representations.
However, our model is only generative in this local sense, since we do not
describe how to sample an entire image with overlapping patches (unlike for instance the
POP model~\cite{amit2007pop}).  Luckily, sampling is not our primary interest and
instead we focus on the representation that we can learn through this process.
The goal of the layer is to transform the incoming signal in a way that will
eventually benefit discriminative tasks. This typically means summarizing the
co-occurrence patterns of low-level features into a single high-level feature
(\eg, this group of pixels describes a line). The hope is that the
estimated distribution has captured this underlying structure in a succinct and
meaningful representation.  Hinton describes this method of representation
learning as ``procrastination''~\cite{hinton2007recognize}, since a supervised
task is not involved until the very last layer and there is an element of hope
that it will eventually be useful.


The statistical model so far given input gives back a likelihood, not a set of
features. We need an activation function, $g$, that takes the likelihood $p_i =
p(X \mid Z=i)$ for each component $i$ and converts it to a feature ready to be
passed to the next layer. To discuss concrete suggestions for $g$, we must be
specific about the type of distribution.

\textbf{Bernoulli.}
Our main focus is when the distribution is a product of Bernoulli distributions
\begin{align} \label{eq:prod-bern}
    p(X \mid Z = z) = \prod_{d} p(x_d; \theta_{z,d}) = \prod_d \theta_{z,d}^{x_d} (1 - \theta_{z,d})^{(1-x_d)},
\end{align}
where $d$ indexes spatial and feature dimensions together. The activation function then forms a
one-hot vector of the maximum likelihood component:
\begin{equation} \label{eq:bern-argmax}
    [g(p)]_i = \begin{cases}1 & \text{if } p_i = \max_j p_j\\
                    0 & \text{otherwise} \end{cases}
\end{equation}
This creates a sparse binary feature representation. In order to make it more
robust to local spatial translations, we spread the 1's to nearby locations using
morphological dilation.

\textbf{Gaussian.} It can also be based on Gaussian mixture models (GMM), with
a range of options for parameterization of the covariance matrix.
Coates~\etal~\cite{coates2010analysis} lend inspiration for $g$,
who similarly built an unsupervised feature extractor using $k$-means clustering. The
activation function they call \textit{hard} is the same as
\eqref{eq:bern-argmax}, with $p_i$ defined as the negative distances to
centroid $i$.  A second one is called \textit{triangle}, which assigns soft scores
to each component and then passes it through a rectifier. The soft score is
designed so that roughly half of the features are rectified. An analogous
activation for GMM would be:
\begin{equation}
    [g(x)]_i = \max\{0, \log p_i - \frac 1F \sum_j \log p_j\}
\end{equation}
%




\subsection{Latent Transformation}

The mixture model's latent space can be extended to be a product of
latent component \textit{and} latent transformation. The latent space thus
holds information not just about which part from a dictionary, but also which
instantiation of the part (\eg, orientation, polarity, etc.). We focus on
transformations that can be expressed as products of cyclic groups. For
instance, we could have a group with $R$ rotations around the part's center. Let
$\Phi : \mathcal{X} \mapsto \mathcal{X}$ be a $\frac{360}{R}$-degree rotation
around the center. The transformation $\Phi$ then generates the cyclic rotational
group. If $x \in \mathcal{X}$ denotes a single sample, let $\vx \in
\mathcal{X}^R$ denote a \textit{complete} sample that contains the complete
group's transformations of $x$:
\begin{equation}
    \vx = [x, \Phi x, \Phi^2 x, \dots, \Phi^{R-1}x]
\end{equation}
We associate $\vx$ with a permutation, $\sigma = (0\, 1\, \dots\, R -
1)$. For a direct product of two cyclic groups, the permutation $\sigma$ will
be the product of the two permutations.  Let $A$ denote the permutation matrix
corresponding to $\sigma$ and let $A(i,j)$ denote indexing the matrix.

We refer to the latent joint variable as $(Z, W)$, where $Z$ represents
component and $W$ permutation. The model for a Bernoulli mixture model with
latent transformation becomes
\begin{align} \label{eq:latent-model}
    (Z, W) &\sim \Cat{\pi}, \\
    X_{w',d} \mid Z=z, W=w &\sim \Bern{ \mu_{z, A({w,w'}), d}}, \quad\quad \forall w', d 
\end{align}
where $w'$ indexes which transformation of the complete sample and $d$ the
sample feature (over channels and both spatial dimensions). Note that the
random variable $X$ models the complete sample. We now see the purpose of $A$:
permuting the parameter space of the complete sample depending on the latent
transformation.  Next, we describe how to estimate the parameters $\pi$ and $\mu$.

\subsection{Permutation EM} \label{sec:permutation-em}

Expectation Maximization (EM) is used to train the original parts model. The
latent permutation mixture model can be trained with an adaptation that we call
Permutation EM. 
We describe how this is done specifically for the Bernoulli
mixture model, although a Gaussian mixture model is closely analogous. We
present the modifications briefly and assume that the reader is familiar with
EM. For more details and derivations, see \cref{apx:permutation-em}.  

\textbf{E-Step.} The responsibilities $\gamma^{(n)}$ for sample $n$, are computed over the joint latent variable $(Z, W)$ as
\begin{align}
    \gamma^{(n)}_{z,w}    = \frac{\ungamma^{(n)}_{z,w}}{ \displaystyle\sum_{z',w'} \ungamma^{(n)}_{z',w'}}, \qquad
     \quad\ungamma^{(n)}_{z,w} = \pi_{z,w} \prod_{w',d} p(x^{(n)}_{A(w,w'),d}; \mu_{z,w',d}),
\end{align}
where, $p$ is defined in \eqref{eq:prod-bern}. To avoid numerical underflow, it is better to compute and store logarithms of the unnormalized responsibilities $\ungamma$ and use the log-sum-exp trick.

\textbf{M-Step.} The parameters are updated as follows:
\begin{align}
    \pi_{z,w} \leftarrow \frac 1N \displaystyle\sum_n \gamma^{(n)}_{z,w} \qquad
    \mu_{z,w,d} \leftarrow \frac{\displaystyle\sum_{n,w'} \gamma^{(n)}_{z,w'} x_{A(w, w'),d}}{\displaystyle\sum_{n,w'} \gamma^{(n)}_{z,w'}}
\end{align}

In words, the E-step computes responsibilities over best matched component \textit{and}
transformation. The complete sample is permuted according to a latent
transformation, which synchronizes the samples of various shape \textit{and} pose into sharp
models in the M-step. In the case of latent rotation, the responsibilities learn the rotation
of each sample relative to an arbitrary canonical rotation.


Since the latent model is a distribution over \textit{complete} samples,
extracting features strictly requires building the complete sample
and evaluating the likelihood of each component and permutation. However, this
is costly and tedious as a feature extractor. Instead, we ignore that we
trained the parts using complete samples, and instead consider each part and
transformation as a separate feature. We then use \eqref{eq:bern-argmax} as we
normally would to code features. The feature space can of course still be interpreted with an
association between parts and its transformed siblings. 

%


\begin{figure}
    \centering
    \input{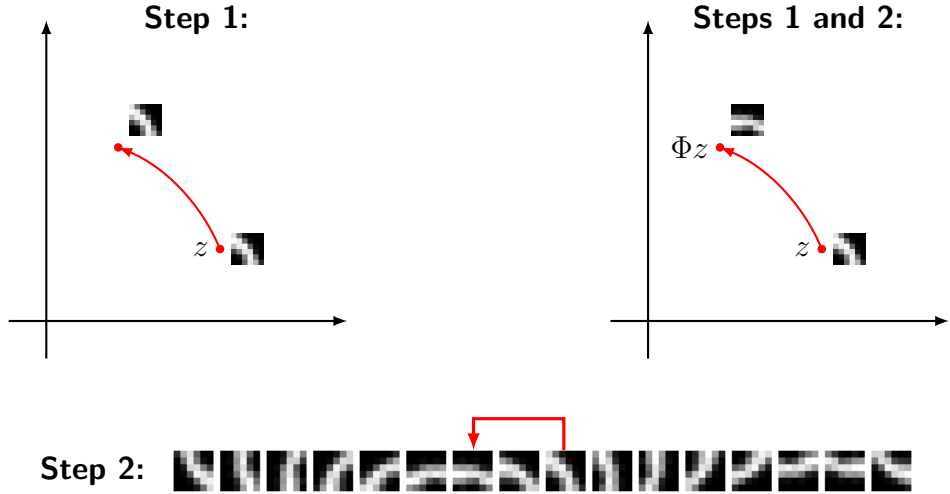}
    \caption[Feature transformation]{\textbf{Feature transformation.}
        Transformation ($\Phi$) of a feature space is a two-step process: (1) spatial transformation, and (2) permutation of the feature space. The figure depicts a 45-degree counter-clockwise rotation around the origin of a feature space with latent rotation ($R=16$). 
    }
    \label{fig:two-step-rotation}
\end{figure}

\subsection{Implementing Transformation}

A few notes on how to implement the transformation $\Phi$, specifically for
latent rotation. At the first layer, $\Phi$ simply
becomes an image rotation operation on the patches (transformation and resampling, typically with
bilinear interpolation).  However, if the patch is used in an intermediate
layer and each feature represents an oriented structure, this is not enough to
correctly transform the feature space. We present two options.

First, the transformation is performed as a two-step process: transforming and
permuting.  The first step is to transform the spatial dimensions, using the
same technique as in the first layer. The difference is that instead of 1 or 3
color channels, we now typically have a lot more feature channels. However, more profoundly,
unlike color pixels, the features represent local structure beyond their center
pixel.  This structure is typically not rotation
invariant, so if we rotate a feature 45 degrees, we also need to rotate what
the feature represents by 45 degrees.  To make this correction we 
permute the feature space, assuming that the incoming feature space is
equivariant to rotation. These two steps are illustrated in
\cref{fig:two-step-rotation}.

An alternative to this method is to rotate the original input and separately
pass them through all previous layers. This has the advantage that we do not
need to do further transforming or permuting. We can even use
this layer on top of features that are not organized to be equivariant to
rotation.  The model will discover on its own how to rotate features to
correspond best to the transformed inputs.  This method is convenient even if
the incoming feature space can be rotated.  For instance, in the first layer we
may use edges with 8 orientations, while in the second layer we may use parts with 32
orientations. We could write the correct permutation to synchronize the 8 and
32 orientations. However, the analytically transformed feature space may not
perfectly correspond to transformation in the input space, due to the lower
fidelity of 8 orientations.  It is possible to choose different methods for
different layers. If two subsequent layers have the same number of latent orientations, then the analytical approach
becomes appropriate and efficient. Regardless of the method used during
training, the finished feature space can always be analytically transformed.

A final note on resampling. When we rotate a point around the patch center, it
may not end up exactly on the feature grid. We need to resample after
transformation, which can be done with techniques like nearest neighbor or
bilinear interpolation. However, even simpler, if your extraction layer is
followed by a pooling layer you can simply check if the new point is within the
pooling region (equivalent to nearest neighbor in the post-pooling feature
space). We use this technique at layers that use pooling and nearest neighbor
at layers that do not use pooling.






\subsection{Classification}

So far, we have only discussed unsupervised representation learning.
Classification is done by training a global (patch covering the entire input)
mixture model for each class $y$ separately, storing the parameters in $\theta_y$. The
same unlabeled training procedure is performed as before, except each model is trained
only on samples from a specific class.  With $C$ separately trained top layers,
a final class prediction is done exactly like feature extraction, where the
maximum likelihood part is identified:
\begin{equation} \label{eq:mm-argmax}
    \hat y = \argmax_y \log p(X \mid Y = y; \theta_y)
\end{equation}
One of the benefits of this model is that new classes can be added without
having to re-train existing classes. Each separately trained model for each
class can itself be a mixture model of several components. In our main
experiments, we use $M=1$. This does not require EM and parameters are
estimated by simple averaging. Even with such simple supervised training step,
we do well thanks to the unsupervised representation from the previous layers.

\subsection{Independence Assumption} \label{sec:independence-assumption}

The beauty of the mixture model is that the assumption of conditional
independence does not imply the assumption of independence. This allows more
complex dependence structure, while still being cheap to train and evaluate.
However, having said that, the conditional independence assumption will still
be violated by natural image data.
There is always strong correlation between proximal features, even conditioned on the component.
There may also be to a lesser extent long-range correlation in certain
situations.\footnote{In MNIST, where digits are centered, if the top of a 1 is
located in the top-right corner, chances are it is a tilted 1 and the bottom
will be located in the bottom-left corner. This creates a long-range
correlation.} 
The reason for the independence
assumption is that it makes training and inference straightforward.

One of the main concerns of breaking this assumption is that likelihood
estimates will be wrong. This is true and likelihoods will in fact be grossly
under-estimated.  However, what we are interested in is only if the likelihoods
compare well between components, as required by \eqref{eq:mm-argmax}. It turns
out that the under-estimations is fairly even across classes, or results would
suffer. If a situation appears where class likelihoods are not balanced, it may
help to standardize the log-likelihoods with statistics estimated from your training data.


With latent transformation, the violation becomes even clearer. The model
assumes according to \eqref{eq:latent-model} that two transformations are
conditionally independent given the joint latent variable. The likelihood of
each individual segment of the complete sample will be nearly perfectly
correlated with all the other transformations. This will under-estimate the
likelihoods further, although again without disruption to our end goal.

\subsection{Layer Types}
\label{sec:mm-layers}
Before describing the models that we evaluate, we outline the layers that we use as building blocks:
\begin{enumerate}[leftmargin=2cm]
    \item[\textbf{Edges}] Parameterless oriented binary edges defined
        in~\cite{bernstein2005part}. The output contains 8 features (cardinal
        and diagonal directions) per color channel based on intensity
        differences. Since a Bernoulli mixture layer requires binary input, this
        provides a bridge between the real-valued input and the rest of the binary network.
    \item[\textbf{Mixture}] Bernoulli mixture model layer. At the final task layer, the patch
        size is set to be equal to the input size, making it effectively
        non-convolutional. Otherwise, the patch size is 6-by-6. The value
        $M$ (mixture components) is left as an option. When used as
        an intermediate layer, we reduce spatial resolution by pooling features in
        4-by-4 disjoint regions using element-wise OR (max) operations.
    \item[\textbf{RotMix}] Bernoulli mixture model with latent orientation.
        Analogous to \textit{Mixture}, with built-in pooling and the same patch
        size.  Can also be used as a final task layer.  Options include $R$
        (rotations) and $M$ (mixture components). The number of parts at test time is $R \cdot M$.
    \item[\textbf{SVM}] Multi-class linear SVM. This is evaluated as a discriminative
        alternative to log-likelihood comparisons.
\end{enumerate}
\begin{table}
\newcommand{\tb}{\textbf}
\newcommand{\xa}[1]{$\num[round-mode=places,round-precision=1]{#1}\phantom{\;\,\pm\num[round-mode=places,round-precision=1]{0.0}}$}
\newcommand{\xae}[2]{$\mathbf{#2}\phantom{\;\,\pm\num[round-mode=places,round-precision=1]{0.0}}$}
\newcommand{\xb}[2]{$\num[round-mode=places,round-precision=1]{#1}\pm\num[round-mode=places,round-precision=1]{#2}$}
\newcommand{\xc}{-\qquad}
\newcommand{\xd}[2]{-\qquad}
\newcommand{\xe}[3]{$\textbf{#3}\pm\num[round-mode=places,round-precision=1]{#2}$}
\begin{center}

    \begin{tabular*}{\linewidth}{l|rrr|rr}
\toprule
Model                                         & \textit{basic}               & \textit{bg-rand}             & \textit{bg-img}              & \textit{rot}             & \textit{bg-img-rot} \\ \midrule
        RBM~\cite{sohn2012learning}           & \xc{}                        & \xc{}                        & \xc{}                        & \xa{15.6}                & \xa{54.00}\\
RBM~\cite{sohn2013pgbm}                       & \xc{}                        & \xa{11.39}                   & \xa{15.42}                   & \xc{}                    & \xa{49.89}\\
DBM-3~\cite{larochelle2007empirical}          & \xb{3.11}{0.15}              & \xb{6.73}{0.22}              & \xb{16.31}{0.32}             & \xb{10.30}{0.27}         & \xb{47.39}{0.44}\\
SDAE-3~\cite{vincent2010stacked}              & \xb{2.61}{0.14}              & \xb{8.52}{0.24}              & \xb{16.68}{0.33}             & \xb{8.76}{0.29}          & \xb{43.76}{0.43}\\
SDAE-3 (Linear SVM)~\cite{vincent2010stacked} & \xb{2.63}{0.14}              & \xb{11.32}{0.28}             & \xb{14.55}{0.31}             & \xb{10.00}{0.26}         & \xb{42.07}{0.43}\\
SDAE-3 (Kernel SVM)~\cite{vincent2010stacked} & \xb{2.57}{0.14}              & \xb{10.16}{0.26}             & \xb{14.06}{0.30}             & \xb{8.64}{0.25}          & \xb{39.07}{0.43}\\
PGBM~\cite{sohn2013pgbm}                      & \xc{}                        & \xa{7.27}                    & \xa{13.33}                   & \xc{}                    & \xa{45.45}\\
Supervised PGBM~\cite{sohn2013pgbm}           & \xc{}                        & \xa{6.87}                    & \xa{12.85}                   & \xc{}                    & \xa{44.67}\\
PGBM + DN-1~\cite{sohn2013pgbm}               & \xc{}                        & \xa{6.08}                    & \xa{12.25}                   & \xc{}                    & \xa{36.76}\\
TIRBM~\cite{sohn2012learning}                 & \xc{}                        & \xc{}                        & \xc{}                        & \xae{4.2}{4.2}           & \xa{35.50}\\ \midrule
Plain                                         & \xb{1.913200}{0.027989}      & \xb{7.883600}{0.049758}      & \xb{10.049200}{0.071483}     & \xb{11.762000}{0.080528} & \xb{36.241600}{0.140088}   \\
Plain-SVM                                     & \xe{1.314000}{0.022091}{1.3} & \xe{5.972000}{0.051284}{6.0} & \xe{8.032800}{0.051871}{8.0} & \xb{9.092400}{0.043514}  & \xb{33.356400}{0.189113}   \\
Oriented                                      & \xd{3.852800}{0.157590}      & \xd{10.907200}{0.167459}     & \xd{18.955600}{0.988735}     & \xb{5.511600}{0.148202}  & \xe{24.973600}{0.800533}{25.0}\\
    \bottomrule
\end{tabular*}
\end{center}
    \caption[Results on variations of MNIST]{
        \textbf{Variations of MNIST.} (Error rate, \%)
    Evaluation on variations of MNIST (see~\cref{fig:mnist-variations-examples}).
    10,000 samples are used for training and 60,000 for testing. As a result,
    the first column \textit{basic} should not be compared with regular
    MNIST results that train on 60,000 and test on 10,000. Our models in the
    bottom three rows are described in \cref{sec:mm-results}.
}
\label{tab:mnist}
\end{table}


\def\dash{-\phantom{.0}}
\begin{table}
\newcommand{\xa}[1]{$\num[round-mode=places,round-precision=1]{#1}\phantom{\;\,\pm\num[round-mode=places,round-precision=1]{0.0}}$}
\newcommand{\xb}[2]{$\num[round-mode=places,round-precision=1]{#1}\pm\num[round-mode=places,round-precision=1]{#2}$}
\newcommand{\xc}{-\quad\quad\quad\quad}
\newcommand{\xd}[2]{\color{gray}{\xb{#1}{#2}}}
\newcommand{\xe}[3]{$\mathbf{#3}\pm\num[round-mode=places,round-precision=1]{#2}$}
\begin{center}
    \newcommand{\tb}{\textbf}
    \begin{tabular*}{\linewidth}{l@{\,\,\,}c|rrr|rr}
    \toprule
        Model                               && \textit{basic}               & \textit{bg-rand}               & \textit{bg-img}                & \textit{rot}                   & \textit{bg-img-rot} \\ \midrule \multicolumn{6}{l}{1000 samples/class (full)} \\ \midrule
        CNN                                 && \xe{1.770000}{0.050990}{1.8} & \xe{4.250000}{0.192354}{4.3}   & \xe{8.040000}{0.355528}{8.0}   & \xb{10.360000}{0.313688}       & \xb{31.890000}{0.146287}\\
        CNN (+rot.\ aug.)                   && \xc                          & \xc                            & \xc                            & \xe{4.0125}{0.1473728}{4.0}    & \xe{19.150000}{0.318198}{19.2}\\
        Our model                           && \xb{1.913200}{0.027989}      & \xb{7.883600}{0.049758}        & \xb{10.049200}{0.071483}       & \xb{5.511600}{0.148202}        & \xb{24.973600}{0.800533}         \\ \midrule \multicolumn{6}{l}{100 samples/class} \\ \midrule
        Semi-sup.\ PGBM~\cite{sohn2013pgbm} &\cmark& \xc                          & \xb{12.00}{0.80}               & \xb{20.30}{0.20}               & \xc                            & \xb{59.20}{0.70}  \\
        CDBN~\cite{lee2011unsupervised}     &\cmark&${}^\dagger$\xe{2.62}{0.12}{2.6}              & \xc                            & \xc                            & \xc                       & \xc  \\
        CNN                                 && \xb{4.700000}{0.130384}      & \xe{10.180000}{0.258070}{10.2} & \xb{17.110000}{0.446542}       & \xb{24.980000}{0.444522}       & \xb{53.020000}{0.939468} \\
        CNN (+rot.\ aug.)                   && \xc                          & \xc                            & \xc                            & \xb{11.5000}{0.176068}         & \xe{31.310000}{0.811419}{31.3} \\
        Our model                           &\cmark& \xe{2.636000}{0.113903}{2.6} & \xe{10.164000}{0.100100}{10.2} & \xe{14.730400}{0.381811}{14.7} & \xe{7.754400}{0.755834}{7.8}   & \xb{35.297600}{2.083907}           \\ \midrule \multicolumn{6}{l}{10 samples/class} \\ \midrule
        CNN                                 && \xb{15.590000}{0.445421}     & \xb{24.420000}{0.853581}       & \xb{34.150000}{2.767128}       & \xb{58.940000}{0.677791}       & \xb{75.510000}{0.428252}\\
        CNN (+rot.\ aug.)                   && \xc                          & \xc                            & \xc                            & \xb{26.230000}{0.393192}       & \xb{54.670000}{1.038075} \\
        Our model                           &\cmark& \xe{8.844500}{0.722049}{8.8} & \xe{21.055000}{1.137081}{21.1} & \xe{29.713200}{1.626599}{29.7} & \xe{16.364400}{0.919581}{16.4} & \xe{48.430800}{2.421398}{48.4}              \\
    \bottomrule
        \multicolumn{7}{l}{Check mark (\cmark) means all 10,000 training samples are used as additional unlabeled data} \\
    \end{tabular*}
\end{center}
    \caption[Small sample size]{
        \textbf{Small sample size.} (Error rate, \%)
        The sample size is varied from full (1,000
        samples/class) to 10 samples/class. However, in the non-full
        cases, our models are allowed to observe the rest of the 10,000 samples
        as unlabeled data. This can be leveraged using pretraining (\eg, our model)
        or semi-supervision (\eg, PGBM~\cite{sohn2013pgbm}). Our model refers
        to \textit{Plain} (see \cref{sec:mm-results}) for the three columns on the left with no rotation and
        \textit{Oriented} for the two columns on the right with rotation. The baseline
        CNN models do not use additional unlabeled samples. Since our rotational
        model is designed to work well for randomly rotated classes, we also augment
        the CNN training data with random rotations (``+rot.\ aug.''). The CNN
        is designed to be as similar as possible to our model. The biggest differences are in training
        and that our model has binary and not floating point activations. Our
        model performs well in all settings and particularly well when labeled sample size
        is small. (${}^\dagger$evaluated on the
        original MNIST with 10,000 samples and not \textit{basic} on 60,000 samples)
    }
\label{tab:mnist-sample-size}
\end{table}


\subsection{Experiments}
\label{sec:mm-results}

We consider three models, each with three layers: 
\begin{enumerate}[leftmargin=2.7cm]
    \item[\textbf{Plain}]
        1. Edges, 2. Mixture ($S=6, M = 1280$), 3. Mixture ($M=1$)
    \item[\textbf{Plain-SVM}]
        1. Edges, 2. Mixture ($S=6, M = 1280$), 3. SVM
    \item[\textbf{Oriented}]
        1. Edges, 2. RotMix ($S=6, M = 40, R=32$), 3. RotMix ($M=1, R=16$)
\end{enumerate}
The models are evaluated on variations on MNIST~\cite{larochelle2007empirical},
a set of derivatives that introduce various forms of background
clutter and/or random rotations. It designates 10,000 samples for
training, 2,000 for validation, and 60,000 for testing. This differs from
the original MNIST, which has 60,000 samples for training and 10,000
for testing. Examples can be seen in \cref{fig:mnist-variations-examples}.

Results and comparisons to contemporary work are summarized in
\cref{tab:mnist}. \textit{Plain} provides strong results across all
benchmarks. Switching the last layer to a discriminatively trained linear SVM
consistently improves results. However, remember that this is less flexible in
terms of adding new classes and does not not work well for randomly rotated samples.

By using mixtures with latent orientation
(\textit{Oriented}), the results on MNIST with random rotations can be
improved even further.  TIRBM, which is the only other model that explicitly
deals with random rotations, does slightly better for rotated MNIST with clean
backgrounds. However, in the case of cluttered background images, we offer a
10-point improvement over their results. Our strong results on background clutter are also shown by
beating all versions of PGBM, which has per-pixel gates specifically designed to deal
with noise and clutter. It is worth noting however that RBMs typically do not use
convolutional weight sharing, which puts them at a disadvantage. The
Convolutional DBN (CDBN) performs on par with our results as seen
in \cref{tab:mnist-sample-size}. However, our model offers more straightforward
training and inference.

We also compare against a 3-layer CNN, designed to be closely analogous to our
network in \cref{tab:mnist-sample-size}. We see that it performs better than our
mixture model classifiers, but fairly close to our SVM-based results in \cref{tab:mnist}. If we
reduce sample size from 1,000 per class to 100 and 10 per class, our model
becomes increasingly competitive and outperforms the CNN. In order to give the CNN a more fair
comparison on rotated MNIST, we also try augmenting data using random
rotations. This greatly improves CNN results and it excels with enough samples.
However, in the small-sample regime at only 10 samples per class, we still do
better.

\begin{figure}
    \begin{center}
    \begin{minipage}[b]{0.47\linewidth}
        \begin{center}
            Mixture on \emph{basic\phantom{g}}\\
            \includegraphics[width=\textwidth]{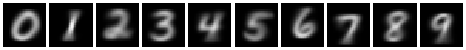} \\
            RotMix on \emph{rot\phantom{g}}\\
            \includegraphics[width=\textwidth]{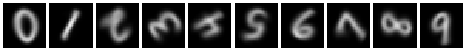} \\
            Corrected RotMix on \emph{rot\phantom{g}}\\
            \includegraphics[width=\textwidth]{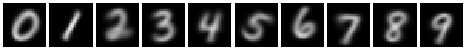} \\
        \end{center}
    \end{minipage}
    \hfill
    \begin{minipage}[b]{0.47\linewidth}
        \begin{center}
            Mixture on \emph{bg-img}\\
            \includegraphics[width=\textwidth]{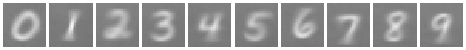} \\
            RotMix on \emph{bg-img-rot}\\
            \includegraphics[width=\textwidth]{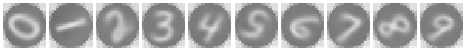} \\
            Corrected RotMix on \emph{bg-img-rot}\\
            \includegraphics[width=\textwidth]{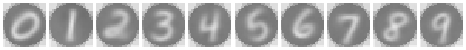} \\
        \end{center}
    \end{minipage}
    \end{center}
    \caption[Class-level mixture model on MNIST]{
        \textbf{Class-level mixture model on MNIST.} Top row is a Mixture layer (see \cref{sec:mm-layers})
        with a single component trained on correctly rotated samples. The
        middle row is a RotMix layer with a single component and 16 latent
        rotations trained on randomly rotated data. The bottom row is the same
        as the middle row, except 5 correctly rotated samples have been used to
        designate the canonical rotation.
    }
    \label{fig:mixture-mnist}
\end{figure}

Examples of trained parts (layer 2) can be seen in \cref{fig:parts-mnist} and
trained classes (layer 3) can be seen in \cref{fig:mixture-mnist}. We note that a
task-level mixture with latent orientations is able to train even cleaner versions of the
digits on rotated samples than a plain mixture is on regular samples. This is
because the \textit{RotMix} layers can account for natural rotational
variation, such as the tilt of the digit 1 (see \cref{fig:mixture-mnist}).
Source code for all experiments is publicly
available.\footnote{\url{https://github.com/gustavla/parts-net}}

\section{Autoencoder}
\label{sec:autoencoder}

We have seen that stacked mixture models can be effective on simple benchmarks,
such as variations on MNIST. However, the network contains only two
parameterized layers and is difficult to grow deeper due
to the greedy training. We shift our attention to another unsupervised model
that can be trained end-to-end and thus allows deeper models to be trained to convergence.

The autoencoder is a feed-forward network with an encoder ($f$) and a decoder ($g$). The
encoder can for instance be inspired by a regular convolutional feed-forward network used for 
a visual classification task. The value returned by the encoder is then fed through a decoder
to produce a reconstruction $\hat{x} = g(f(x))$. This
is trained end-to-end using gradient descent on a per-pixel reconstruction loss:
\begin{equation} \label{eq:autoencoder-loss}
    \ell(x) = \| x - \hat{x} \|^2
\end{equation}
%
%
The autoencoder should contain an information bottleneck
(most obviously controlled by the size of the encoded vector) or the task becomes trivial. The
network must economize and learn how to represent high-level input with as
few bits as possible. In the ideal case, the encoding represents local
coordinates on a low-dimensional manifold that perfectly captures the data
distribution. This does not happen and the main problem stems from the
reconstruction loss, which operates on a per-pixel level with no regard to
high-level semantics. Two samples that are perceptually similar to humans may have a large $L_2$
difference. This uncompromising nature of the squared $L_2$ loss forces the encoding
to devote much informational real estate to low-level information.

The primary purpose of this section is to evaluate autoencoders for
pretraining, in order to establish a baseline for subsequent chapters.  Their
poor ability to capture high-level semantics results in modest improvements
over using no pretraining at all.
In order to improve training of our baseline, we propose the Shortcut
Autoencoder that converges well and improves pretraining results.

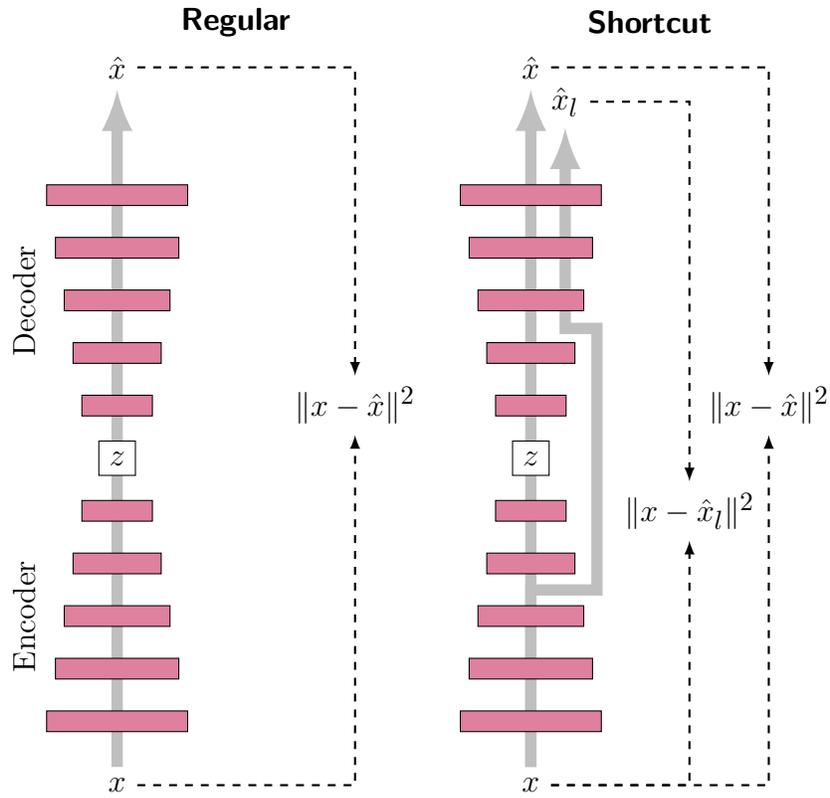
\begin{figure}
    \centering

\begin{tikzpicture}[thick,>=latex,scale=1]
    \tikzset{main line/.style={very thick}}
    \tikzstyle{flow arrow}=[->,line width=1.5mm,lightgray]
    \tikzstyle{loss arrow}=[->,dashed]
    \tikzstyle{layer}=[rectangle,draw,thick,fill=purple!50,thin,minimum width=1.5cm,minimum height=5]

    \foreach \col in {1, ..., 2} {
        \ifthenelse{\col=9}{
            \pgfmathsetmacro{\x}{\col * 5.5}
        }{
            \pgfmathsetmacro{\x}{\col * 5.5}
        }
        \pgfmathsetmacro{\dy}{0.7}
        \pgfmathsetmacro{\layers}{5}
        \pgfmathsetmacro{\maxwidth}{60}
        \pgfmathsetmacro{\minwidth}{20}

        \node (x) at (\x, -0.15) {$x$};
        \pgfmathsetmacro{\y}{2*\dy*(\layers+1)+1}
        \node (xhat) at (\x, \y) {$\hat{x}$};

        \draw[flow arrow] (x.north) -| (xhat);

        \ifthenelse{\col=2}{
            \node (xhat2) at ([xshift=13,yshift=-13]\x, \y) {$\hat{x}_l$};

            \draw[flow arrow] (x.north) -| ([yshift=67]x.north) -|  ([xshift=25,yshift=166]x.north) -| (xhat2);
        }{}

        \foreach \b in {1, ..., \layers} {
            \pgfmathsetmacro{\y}{\dy*\b}
            \pgfmathsetmacro{\width}{\maxwidth-(\maxwidth-\minwidth)/(\layers+1)*\b}
            \node[layer,minimum width=\width] (layer-\b) at (\x, \y) {};

            \pgfmathsetmacro{\y}{\dy*(2+2*\layers-\b)}

            \node[layer,minimum width=\width] (layer+\b) at (\x, \y) {};
        }

        \pgfmathsetmacro{\y}{\dy*(\layers+1)}
        \node[layer,fill=white!50,minimum height=7,minimum width=7] (embedding) at (\x, \y) {$z$};
        \node (loss) at ([xshift=90,yshift=20]embedding) {$\|x - \hat{x}\|^2$};

        \ifthenelse{\col=1}{
            \pgfmathsetmacro{\y}{2+2*\dy*\layers+1}
            \node[] at ([xshift=45]\x, \y) {\textsf{\textbf{Regular}}};

            \node[rotate=90] at ([xshift=-15]layer-3.west) {Encoder};
            \node[rotate=90] at ([xshift=-15]layer+3.west) {Decoder};
        }{}

        \ifthenelse{\col=9}{
            \node (loss2) at ([xshift=60,yshift=-20]embedding) {$\|z_l - \hat{z}_l\|^2$};

            \draw[loss arrow] ([yshift=5]layer-3.north) -| (loss2);
            \draw[loss arrow] ([yshift=-5]layer+3.south) -| (loss2);

            \pgfmathsetmacro{\y}{2+2*\dy*\layers+1}
            \node[] at ([xshift=45]\x, \y) {\textsf{\textbf{Ladder}}};
        }{}

        \ifthenelse{\col=2}{
            \node (loss2) at ([xshift=60,yshift=-20]embedding) {$\|x - \hat{x}_l\|^2$};

            \draw[loss arrow] (xhat2.east) -| (loss2);
            \draw[loss arrow] (x.east) -| (loss2);

            \pgfmathsetmacro{\y}{2+2*\dy*\layers+1}
            \node[] at ([xshift=45]\x, \y) {\textsf{\textbf{Shortcut}}};
        }{}

        \draw[loss arrow] (xhat.east) -| (loss);
        \draw[loss arrow] (x.east) -| (loss);
    }

\end{tikzpicture}

    \caption[Autoencoder schematics]{
        \textbf{Autoencoder schematics.}
        Schematics of the two autoencoders we consider for pretraining.
        Regular autoencoders pass samples through an encoder and then a
        decoder, attaching a reconstruction loss. The depth is double that of
        the encoder, which can make it difficult to train properly.  We propose
        \textit{shortcut autoencoders}, that separately passes samples through
        shortcuts, adding additional auxiliary reconstruction losses that
        converge faster.
    }
    \label{fig:autoencoder}
\end{figure}

\subsection{Related Work}

Using autoencoders to facilitate the training of neural networks has been
around since 1987~\cite{ballard1987modular}. 20 years later, a similar idea in
a more familiar form was developed by Bengio~\etal~\cite{bengio2007greedy}:
Greedy layer-wise training of a DBN by fitting shallow autoencoders
(single-layer encoder, single-layer decoder). The idea was inspired by Hinton
\& Salakhutdinov~\cite{hinton2006reducing}, who used RBMs for
layer-wise pretraining. Incidentally, the network that they trained this way was a
multi-layer autoencoder for the purpose of dimensionality reduction.
With improved techniques of training deep neural networks from scratch,
layer-wise pretraining fell out of favor. None of the major milestone networks
(AlexNet~\cite{alexnet}, VGG-16~\cite{vgg16}, ResNet~\cite{resnet}) have since
used such network priming. Attempts at making use of autoencoders in
semi-supervised settings has
continued~\cite{socher2011semi,valpola2015neural,rasmus2015semi,zhao2015whatwhere},
but has largely alluded widespread success.

There are many variations on the regular autoencoder.
Weights between the encoder and the decoder can be either separate
or tied~\cite{vincent2008extracting}. The Denoising
Autoencoder~\cite{vincent2008extracting,vincent2010stacked}, a precursor
to dropout~\cite{srivastava2014dropout}, improves generalization by randomly zeroing out input features.
One of the benefits of autoencoders (as opposed to for
instance Principal Component Analysis) is the ease of experimenting
with new losses and regularizations, such as adding sparsity regularization of
the latent space~\cite{ng2011sparse}.
In Transforming Autoencoders~\cite{hinton2011transforming}, the input image and
the target image are different according to a transformation whose parameters
are fed to the network. This promotes a more organized feature space.
Regular autoencoders are not suited for generating samples, since we would need
to know how to meaningfully sample the latent distribution. This is addressed
in Variational Autoencoders~\cite{kingma2014auto}, where the encoder instead of
a latent vector produces sufficient statistics of a distribution that are then
used to sample the latent vector.  The statistics are regularized toward
preset values (\eg, unit normal), so that latent vectors
can be sampled directly from this distribution. Another attempt at improving autoencoders for
generative purposes trains the autoencoder to be one link in a Markov
chain~\cite{bengio2014deep}.

\subsection{Shortcut Autoencoder}

We have seen the loss of a regular autoencoder in \eqref{eq:autoencoder-loss}.
This can be non-trivial to train due to the total depth of combining both
encoder and decoder. Let $\ell_k$ denote the reconstruction error when passing a
sample through only the $k$ first encoding layers and then the corresponding
decoding layers:
\begin{equation}
    \ell_k(x) = \| x - \hat{x}_k \|,\qquad \hat{x}_k = (g_1 \circ \dots \circ g_k \circ f_k \circ \dots \circ f_1)(x),
\end{equation}
where $f_k$ is the $k$:th layer of the encoder and $g_k$ the corresponding
decoder layer. A regular autoencoder is thus $\ell = \ell_L$, where $L$ is the
number of layers. The Shortcut Autoencoder adds one or several auxiliary
$\ell_l$ to the objective:
\begin{equation}
    \ell(x) = \ell_L(x) + \lambda \sum_{k \in \mathcal{L}} \ell_k(x)
\end{equation}
The $\mathcal{L}$ defines which layers get shortcut losses (does not need to be
exhaustive) and $\lambda$ regulates their importance (we use $\lambda = 1$).
The auxiliary losses can be implemented by building separate truncated networks
that share weights with the full loss ($\ell_L$) and the other shortcut losses. The 
shallow losses train faster and make it easier for the deeper paths to converge.

There are a few approaches that resemble this technique.
Bengio~\etal~\cite{bengio2007greedy} used shallow autoencoders to train a deep
network in a greedy manner. This starts by training for $\ell_1$ and freezing the parameters of $f_1$. Next, a shallow encoder is trained on top of the second layer, with
$f_1(x)$ as the input. This is different, although reminiscent, to training for
$\ell_2$ without updating the parameters of layer 1. If this had been done, it would have
been a close sequential analog of our proposed method. Shortcut connections in
autoencoders are not new~\cite{valpola2015neural,mao2016image}. However,
the paths in those works are part of the network architecture with aggregation of the main path
and the shortcut paths, for instance using element-wise means~\cite{mao2016image}.
An alternative to shortcuts is to try to make activations synchronized between
the encoder and the decoder through $L_2$ losses. This idea of synchronizing
activations is similarly used in FitNets~\cite{romero2014fitnets}, where the
activations of a deep model tries to mimic that of a shallow. We did not find
success with such regularization and results were consistently worse. Our
approach of not promoting synchronization explicitly, but instead implicitly by
requiring subsequent layers to deal with both paths, was inspired by our work
on fractal networks~\cite{larsson2017fractalnet}.

\subsection{Baseline Results}

\begin{table}
\begin{center}

\newcommand{\xa}[2]{$\num[round-mode=places,round-precision=1]{#1}$}
    \begin{tabular*}{\linewidth}{l@{\qquad}|@{\extracolsep{\fill}}rr}
        \toprule
        Method                         & Classification (\%mAP) & Segmentation (\%mIU) \\ \midrule
        No pretraining                 & \xa{46.15}{a231}       & \xa{23.50}{a228}     \\ \midrule
        Autoencoder                    & \xa{50.77}{mu105}      & \xa{25.99}{a256}     \\
        Shortcut Autoencoder           & \xa{53.29}{mu104}      & \xa{28.66}{a255}     \\
        Layer-wise $k$-means~\cite{krahenbuhl2016datadriven,donahue2016adversarial}    & \xa{56.6}{paper}                    & \xa{32.6}{paper}                            \\
        
        \bottomrule
    \end{tabular*}
\end{center}
    \caption[Autoencoders for pretraining]{
    \textbf{Autoencoders for pretraining.} Autoencoders may slightly help, but
    the benefits are modest. Adding shortcuts makes convergence faster and
    offers slightly improved results. However, the improvements are so
    modest that it is hard to know if meaningful semantic knowledge was
    internalized during the pretraining phase, or if the parameter statistics
    simply offered a better and more even initialization. We also include
    results from a method of using $k$-means to initialize the network. This is similar to
    the idea of using stacked mixture models (\cref{sec:smm}) as a starting
    point for end-to-end fine-tuning.
}
\label{tab:autoencoder}
\end{table}

We evaluate autoencoders as a method of pretraining by taking the encoder and
using it as a starting point for a downstream task. For this, we use two
standard benchmarks: VOC 2007 Classification and VOC 2012 Semantic
Segmentation. We revisit these benchmarks in \cref{chp:selfsup,chp:multiproxy},
providing more training and network details. For now, it is worth knowing that
we allow end-to-end fine-tuning of the pretrained network. This allows us to
directly compare results against randomly initializing the network.

The encoder uses AlexNet~\cite{alexnet} and in the case of shortcuts we place them at
\texttt{conv2}, \texttt{conv4}, \texttt{fc6}.  We do note use a dedicated
hidden representation $z$ as \cref{fig:autoencoder} suggests. We tried this,
but found that using \texttt{fc7} directly for this purpose yields slightly better
results. This still creates a bottleneck since 4,096 features is 2.6\% of the input
feature size. The notion of a single central bottleneck also lacks nuance.
Adding dropout, which we use, is another way of limiting information bandwidth
throughout the entire network (an idea rooted in DAE~\cite{vincent2008extracting} and expanded in Ladder
Networks~\cite{valpola2015neural}).

Results are shown in \cref{tab:autoencoder}. Regular autoencoders offer a
modest improvement over random initialization. This is somewhat improved by
adding shortcuts, which also makes the network converge faster.
However, it still does not perform better than using layer-wise $k$-means to
initialize the layers, which is similar to if we had initialized the network
with a stacked mixture model from \cref{sec:smm}. Both results are still not satisfactory and these
numbers serve to show how difficult it is to prime a network using unlabeled
data. We will revisit this in \cref{chp:selfsup} and show that self-supervision
is a much more effective method of unsupervised pretraining.

\setcounter{chapter}{1}
\chapter{Automatic Colorization}
\label{chp:colorization}

Leaving representation learning aside, we develop a method of automatic
colorization and evaluate its use as a graphics application: a way to
revitalize old black-and-white photographs. We return to unsupervised learning in
\cref{chp:selfsup}, where we will use our colorization network to drive
representation learning.

\section{Introduction}
\label{sec:intro}

Colorization of grayscale images is a simple task for the human imagination.
A human need only recall that sky is usually blue and grass is green; for
many objects, the mind is free to hallucinate any of several plausible colors.
The high-level comprehension required for this process is precisely why
the development of fully automatic colorization algorithms is challenging.
Colorization is therefore intriguing beyond its immediate practical utility in
many graphics applications.  Automatic colorization serves as a proxy measure
for visual understanding.  Our work makes this connection explicit; we unify a
colorization pipeline with the type of deep neural network architectures
driving advances in image classification and object detection.

Both our technical approach and focus on fully automatic results depart from
past work.  Given colorization's importance across multiple applications
(\eg, historical photographs and videos~\cite{tsaftaris2014novel}, artist
assistance~\cite{sykora2004unsupervised,qu2006manga}), much research strives
to make it cheaper and less time-consuming~\cite{
   welsh2002transferring,
   levin2004colorization,
   irony2005colorization,
   charpiat2008automatic,
   morimoto2009automatic,
   chia2011semantic,
   gupta2012image,
   deshpande2015learning,
   cheng2015deep}.
However, most methods still require some level of user input~\cite{
   levin2004colorization,
   sapiro2005inpainting,
   irony2005colorization,
   charpiat2008automatic,
   chia2011semantic,
   gupta2012image}.
Our work joins the relatively few recent efforts on developing fully automatic colorization
algorithms~\cite{morimoto2009automatic,deshpande2015learning,cheng2015deep}.
Some~\cite{deshpande2015learning, cheng2015deep} show promising results,
especially on photos of certain typical scene types (\eg, beaches and
landscapes).  However, their success is limited on complex images with
foreground objects.

\begin{figure}[!t]
   \hspace{-0.35cm}
   \input{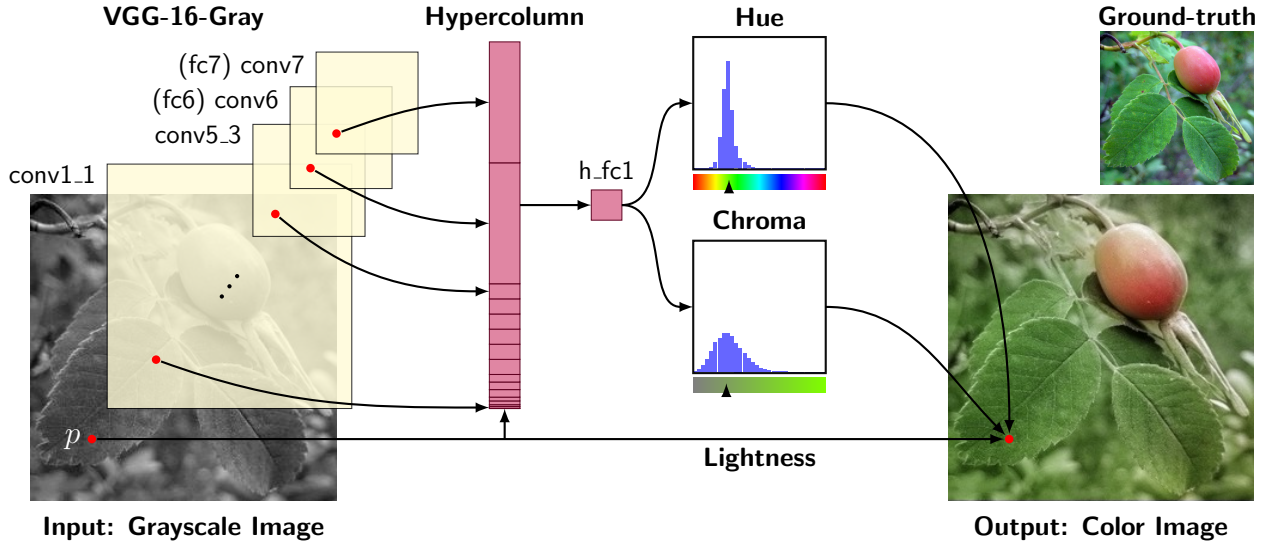}
   \vspace{-0.2cm}
    \caption[System overview]{
   \textbf{System overview.}
      We process a grayscale image through a deep convolutional
      architecture~(VGG-16)~\cite{vgg16} and take spatially localized multilayer
      slices (hypercolumns)~\cite{MYP:ACCV:2014,mostajabi2015feedforward,hariharan2015hypercolumns},
      as per-pixel descriptors.  We train our system end-to-end for the task
      of predicting hue and chroma distributions for each pixel $p$ given its
      hypercolumn descriptor.  These predicted distributions determine color
      assignment at test time.
   }
   \label{fig:schematic}
\end{figure}

At a technical level, existing automatic colorization methods often employ
a strategy of finding suitable reference images and transferring their
color onto a target grayscale image~\cite{morimoto2009automatic,
deshpande2015learning}.  This works well if sufficiently similar reference
images can be found, which becomes increasingly difficult for unique grayscale
input images.  Such a strategy also requires availability and search of a
large repository of reference images at test time.  In contrast, our approach
is entirely free from database search and fast at test time.
Section~\ref{sec:related} provides a more complete view of prior methods,
highlighting differences with our proposed algorithm.

Our approach to fully automatic colorization converts two intuitive observations
into design principles.  First, semantic information matters.  In order to
colorize arbitrary images, a system must interpret the semantic composition
of the visual scene (what is in the image: faces, animals, cars, plants,
etc.) as well as localize scene components (where things are).  Recent
dramatic advances in image understanding rely on deep convolutional neural
networks (CNNs) and offer tools to incorporate semantic parsing and
localization into a colorization system.

Our second observation is that while some scene elements can be assigned a
single color with high confidence, other elements (\eg, clothes or cars)
may draw from many suitable colors.  Thus, we design our system to predict a
color histogram, instead of a single color, at every image location.
Figure~\ref{fig:schematic} sketches the CNN architecture we use to connect
semantics with color distributions by exploiting features across multiple
abstraction levels.  Section~\ref{sec:method} provides details.

Section~\ref{sec:color-experiments} reports extensive experimental validation of
our algorithm against competing methods~\cite{welsh2002transferring,
deshpande2015learning} in two settings:  fully (grayscale input only) and
partially (grayscale input and reference global color histogram provided)
automatic colorization.  Across every benchmark metric and every
dataset~\cite{xiao2010sun,patterson2014sun,imagenet}, our method achieves the
best performance.  Lacking an established standard colorization benchmark,
exhaustive comparison is necessary.  To ease this burden for future research,
we propose a new colorization benchmark on the ImageNet
dataset~\cite{imagenet}, which may simplify quantitative comparisons and drive
further progress.  Our system achieves such a drastic performance leap that
its fully automatic colorization output is superior to that of  prior methods
relying on additional information such as reference images or ground-truth
color histograms.

Section~\ref{sec:final} summarizes our contributions:
\begin{itemize}
    \item A novel technical approach to colorization, bringing semantic
        knowledge to bear using CNNs, and modeling color distributions.
    \item State-of-the-art performance across fully and partially automatic
        colorization tasks.
    \item A new ImageNet colorization benchmark.
\end{itemize}

\section{Related work}
\label{sec:related}

Previous colorization methods broadly fall into three categories:
scribble-based~\cite{
   levin2004colorization,
   huang2005adaptive,
   qu2006manga,
   yatziv2006fast,
   luan2007natural},
transfer~\cite{
   welsh2002transferring,
   irony2005colorization,
   tai2005local,
   charpiat2008automatic,
   morimoto2009automatic,
   chia2011semantic,
   gupta2012image},
and automatic direct prediction~\cite{
   deshpande2015learning,cheng2015deep}.

\emph{Scribble-based} methods, first introduced by
Levin~\etal~\cite{levin2004colorization}, require manually specifying
desired colors of certain regions.  These scribble colors are propagated under
the assumption that adjacent pixels with similar luminance should have similar
color, with the optimization relying on Normalized Cuts~\cite{
shi2000normalized}.  Users can interactively refine results via additional
scribbles.  Further advances extend similarity to
texture~\cite{qu2006manga,luan2007natural}, and exploit edges to reduce color
bleeding~\cite{huang2005adaptive}.

\emph{Transfer-based} methods rely on availability of related \emph{reference}
image(s), from which color is transferred to the target grayscale image.
Mapping between source and target is established automatically, using
correspondences between local descriptors~\cite{welsh2002transferring,
charpiat2008automatic,morimoto2009automatic}, or in combination with manual
intervention~\cite{irony2005colorization,chia2011semantic}.
Excepting~\cite{morimoto2009automatic}, reference image selection is at
least partially manual.

In contrast to these method families, our goal is \emph{fully automatic}
colorization.  We are aware of two recent efforts in this direction.
Deshpande~\etal~\cite{deshpande2015learning} colorize an entire image by
solving a linear system.  This can be seen as an extension of patch-matching
techniques~\cite{welsh2002transferring}, adding interaction terms for spatial
consistency.  Regression trees address the high-dimensionality of the system.
Inference requires an iterative algorithm.  Most of the experiments are
focused on a dataset (SUN-6) limited to images of a few scene classes, and
best results are obtained when the scene class is known at test time.  They
also examine another partially automatic task, in which a desired global color
histogram is provided.

The work of Cheng~\etal~\cite{cheng2015deep} is perhaps most related to ours.
It combines three levels of features with increasing receptive field:
   the raw image patch,
   DAISY features~\cite{tola2008fast}, and
   semantic features~\cite{long2015fully}.
These features are concatenated and fed into a three-layer fully connected
neural network trained with an $L_2$ loss.  Only this last component is
optimized; the feature representations are fixed.

Unlike~\cite{deshpande2015learning,cheng2015deep}, our system
does not rely on hand-crafted features, is trained end-to-end, and treats
color prediction as a histogram estimation task rather than as regression.
Experiments in Section~\ref{sec:color-experiments} justify these principles by
demonstrating performance superior to the best reported by~\cite{
deshpande2015learning,cheng2015deep} across all regimes.



Two concurrent efforts also present feed-forward networks trained end-to-end
for colorization.  Iizuka \& Simo-Serra~\etal~\cite{IizukaSIGGRAPH2016} propose
a network that concatenates two separate paths, specializing in global and
local features, respectively.  This concatenation can be seen as a two-tiered
hypercolumn; in comparison, our 16-layer hypercolumn creates a continuum
between low- and high-level features.  Their network is trained jointly for
classification (cross-entropy) and colorization ($L_2$ loss in Lab).  We
initialize, but do not anchor, our system to a classification-based network,
allowing for fine-tuning of colorization on unlabeled datasets. This is also
what enables us to use colorization for representation learning in \cref{chp:selfsup}.

Zhang~\etal~\cite{zhang2016colorful} similarly propose predicting color
histograms to handle multimodality.  Some key differences include their usage
of up-convolutional layers, deep supervision, and dense training.  In
comparison, we use a fully convolutional approach, with deep supervision
implicit in the hypercolumn design, and, as Section~\ref{sec:method} describes,
memory-efficient training via spatially sparse samples.

\section{Method}
\label{sec:method}

We frame the colorization problem as learning a function
$f : \mathcal{X} \to \mathcal{Y}$.  Given a grayscale image patch
$\vx \in \mathcal{X} = [0, 1]^{S \times S}$, $f$ predicts the color $\vy \in
\mathcal{Y}$ of its center pixel.  The patch size $S \times S$ is the
receptive field of the colorizer.  The output space $\mathcal{Y}$ depends
on the choice of color parameterization.  We implement $f$ according to the
neural network architecture diagrammed in \cref{fig:schematic}.

Motivating this strategy is the success of similar architectures for
semantic segmentation~\cite{
   farabet2013learning,
   long2015fully,
   chen2014semantic,
   hariharan2015hypercolumns,
   mostajabi2015feedforward}
and edge detection~\cite{
   MYP:ACCV:2014,
   GL:ACCV:2014,
   BST:CVPR:2015,
   SWWBZ:CVPR:2015,
   xie2015holistically}.
Together with colorization, these tasks can all be viewed as image-to-image
prediction problems, in which a value is predicted for each input pixel.
Leading methods commonly adapt deep convolutional neural networks
pretrained for image classification~\cite{imagenet,vgg16}.  Such classification
networks can be converted to \emph{fully convolutional} networks that
produce output of the same spatial size as the input, \eg~using the
shift-and-stitch method~\cite{long2015fully} or the more efficient {\em\`a
trous} algorithm~\cite{chen2014semantic}.  Subsequent training with a
task-specific loss fine-tunes the converted network.

Skip-layer connections, which directly link low- and mid-level features to
prediction layers, are an architectural addition beneficial for many
image-to-image problems.  Some methods implement skip connections
directly through concatenation layers~\cite{long2015fully,chen2014semantic},
while others equivalently extract per-pixel descriptors by reading localized
slices of multiple layers~\cite{MYP:ACCV:2014,mostajabi2015feedforward,
hariharan2015hypercolumns}.  We use this latter strategy and adopt the
recently coined \emph{hypercolumn} terminology~\cite{hariharan2015hypercolumns}
for such slices.

Though we build upon these ideas, our technical approach innovates on two
fronts.  First, we integrate domain knowledge for colorization, experimenting
with output spaces and loss functions.  We design the network output to serve
as an intermediate representation, appropriate for direct or biased sampling.
We introduce an energy minimization procedure for optionally biasing sampling
towards a reference image.  Second, we develop a novel and efficient
computational strategy for network training that is widely applicable to
hypercolumn architectures.

\subsection{Color Spaces}
\label{sec:method_color}

We generate training data by converting color images to grayscale according to
$L = \frac{R+G+B}{3}$.  This is only one of many desaturation options and
chosen primarily to facilitate comparison with Deshpande~\etal~\cite{
deshpande2015learning}.  For the representation of color predictions, using
RGB is overdetermined, as lightness $L$ is already known.  We instead consider
output color spaces with $L$ (or a closely related quantity) conveniently
appearing as a separate pass-through channel:
\begin{itemize}
   \item{
      \textbf{Hue/chroma}.
      Hue-based spaces, such as HSL, can be thought of as a color cylinder,
      with angular coordinate $H$ (hue), radial distance $S$ (saturation),
      and height $L$ (lightness).  The values of $S$ and $H$ are unstable
      at the bottom (black) and top (white) of the cylinder.  HSV
      describes a similar color cylinder which is only unstable at the bottom.
      However, $L$ is no longer one of the channels.  We wish to avoid both
      instabilities and still retain $L$ as a channel.  The solution is a
      color bicone, where chroma ($C$) takes the place of saturation. 
      Conversion to HSV is given by
         $V\,=\,L\,+\,\frac{C}{2},\; S\,=\,\frac{C}{V}$.
   }
   \item{
      \textbf{Lab} and $\bfgreek{alpha}\bfgreek{beta}$.
      Lab (or L*a*b) is designed to be perceptually linear.  The color vector
      $(a,b)$ defines a Euclidean space where the distance to the origin
      determines chroma.  Deshpande~\etal~\cite{deshpande2015learning} use a
      color space somewhat similar to Lab, denoted ``ab''. To differentiate,
      we call their color space $\alpha\beta$.
   }
\end{itemize}

\subsection{Loss}
\label{sec:method_loss}

For any output color representation, we require a loss function for measuring
prediction errors.  A first consideration, also used in~\cite{cheng2015deep},
is $L_2$ regression in the ab components of Lab:
\begin{equation}
   \label{eq:loss-l2}
    L_\mathrm{reg}(\vx, \vy) = \| f(\vx) - \vy \|^2
\end{equation}
where $\mathcal{Y} = \mathbb{R}^2$ describes the $(a, b)$ vector space.
However, regression targets do not handle multimodal color distributions well.
To address this, we instead predict distributions over a set of color bins,
a technique also used in~\cite{charpiat2008automatic}:
\begin{equation}
   \label{eq:loss-hist}
   L_\mathrm{hist}(\vx, \vy) = D_\mathrm{KL}( \vy \|f(\vx))
\end{equation}
where $\mathcal{Y} = [0, 1]^K$ describes a histogram over $K$ bins, and
$D_\mathrm{KL}$ is the KL divergence.  The ground-truth histogram $\vy$ is set
as the empirical distribution in a rectangular region of size $R$ around the
center pixel.  We use $R=7$, but experiments show that $R=1$ works well too; in
the latter case the loss becomes the log loss. For histogram predictions, the last
layer of neural network $f$ is always a softmax.

There are several choices of how to bin the color space.  We bin the Lab axes by
evenly spaced Gaussian quantiles ($\mu = 0, \sigma = 25$).  They can be
encoded separately for $a$ and $b$ (as marginal distributions), in which case
our loss becomes the sum of two separate terms defined by \eqref{eq:loss-hist}.
They can also be encoded as a joint distribution over $a$ and $b$, in which
case we let the quantiles form a 2D grid of bins.  In our experiments, we set
$K = 32$ for marginal distributions and $K = 16 \times 16$ for joint.  We
determined these numbers, along with $\sigma$, to offer a good compromise
of output fidelity and output complexity.

For hue/chroma, we only consider marginal distributions and bin axes uniformly
in $[0,1]$.  Since hue becomes unstable as chroma approaches zero, we add a
sample weight to the hue based on the chroma:
\begin{equation} \label{eq:loss-hc}
   L_\mathrm{hue/chroma}(\vx, \vy) =
      D_\mathrm{KL}( \vy_\mathrm{C}\|f_\mathrm{C}(\vx) ) +
      \lambda_H y_\mathrm{C} D_\mathrm{KL}(\vy_\mathrm{H}\|f_\mathrm{H}(\vx))
\end{equation}
where $\mathcal{Y} = [0,1]^{2 \times K}$ and $y_C \in [0,1]$ is the sample
pixel's chroma.  We set $\lambda_H = 5$, roughly the inverse expectation of
$y_\mathrm{C}$, thus equally weighting hue and chroma.

\subsection{Inference}
\label{sec:method_inference}

Given network $f$ trained according to a loss function in the previous section,
we evaluate it at every pixel $n$ in a test image: $\hat{\vy}_n = f(\vx_n)$.
For the $L_2$ loss, all that remains is to combine each $\hat{\vy}_n$ with the
respective lightness and convert to RGB.  With histogram predictions, we
consider options for inferring a final color:
\begin{itemize}
   \item{
      \textbf{Sample}
      Draw a sample from the histogram.  If done per pixel, this may create
      high-frequency noise in areas of high-entropy histograms.
   }
   \item{
      \textbf{Mode}
      Take the $\arg\max_k \hat{y}_{n,k}$ as the color.  This can create
      jarring transitions between colors, and is prone to vote
      splitting for proximal centroids.
   }
   \item{
      \textbf{Median}
      Compute cumulative sum of $\hat{\vy}_n$ and use linear interpolation to
      find the value at the middle bin.  Undefined for circular
      histograms, such as hue.
   }
   \item{
      \textbf{Expectation}
      Sum over the centroids of the color bins weighted by the histogram.
   }
\end{itemize}
In the Lab color space, we achieve the best qualitative and quantitative results
using expectations.  For hue/chroma, the best results are achieved by taking
the median of the chroma.  Many objects can appear both with and without
chroma, which means $C = 0$ is a particularly common bin.  This mode draws the
expectation closer to zero, producing less saturated images.  As for hue, since
it is circular, we first compute the complex expectation:
\begin{equation}
   z =
      \mathbb{E}_{H \sim f_h(\vx)}[H]
      \triangleq \frac 1K \sum_k [f_h(x)]_k \mathrm{e}^{i\theta_k}, \quad
      \theta_k = 2\pi \frac{k + 0.5}{K}
\end{equation}
We then set hue to the argument of $z$ remapped to lie in $[0,1)$.

\begin{figure}
    \begin{center}
        \includegraphics[width=0.4\linewidth]{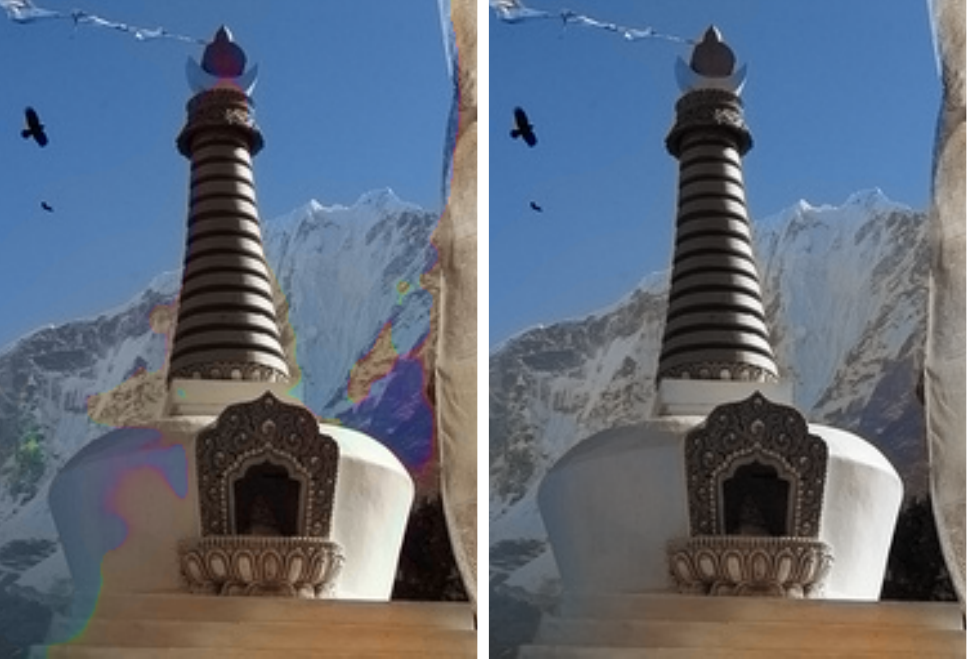}
    \end{center}
    \caption[Artifacts without chromatic fading]{Artifacts without chromatic fading ({\em left}).}
    \label{fig:chromatic-fading}
\end{figure}

In cases where the estimate of the chroma is high and $z$ is close to zero,
the instability of the hue can create artifacts
(\cref{fig:chromatic-fading}).
A simple, yet effective, fix is chromatic fading: downweight the chroma if the
absolute value of $z$ is too small.  We thus re-define the predicted chroma as:
\begin{align}
C_\mathrm{faded} &= \max(\eta^{-1}|z|, 1) \cdot C
\end{align}
In our experiments, we set $\eta = 0.03$ (obtained via cross-validation).

\subsection{Ground-Truth Histogram Transfer}
\label{sec:method-transfer}

So far, we have only considered the fully automatic color inference task.
Deshpande~\etal~\cite{deshpande2015learning}, test a separate task where
the ground-truth histogram in the two non-lightness color channels of the
original color image is made available.\footnote{
   Note that if the histogram of the $L$ channel were available, it would be
   possible to match lightness to lightness exactly and thus greatly narrow
   down color placement.
}
In order to compare, we propose two histogram transfer methods.  We refer to
the predicted image as the \emph{source} and the ground-truth image as the
\emph{target}.

\textbf{Lightness-normalized quantile matching}.
\label{sec:energy}
Divide the RGB representation of both source and target by their respective
lightness.  Then, compute marginal histograms over the resulting three color
channels.  Alter each source histogram to fit the corresponding target
histogram by quantile matching, and multiply by lightness.  Though it does not
exploit our richer color distribution predictions, quantile matching beats the
cluster correspondence method of~\cite{deshpande2015learning}
(see \cref{tab:sun6}).

\textbf{Energy minimization}.
We phrase histogram matching as minimizing energy:
\begin{equation}\label{eq:energy-min}
   E =
      \frac 1N \sum_n D_\mathrm{KL}(\hat{\vy}^*_n \| \hat{\vy}_n) +
      \lambda D_{\chi^2}(\langle \hat{\vy^*} \rangle, \vt)
\end{equation}
where $N$ is the number of pixels,
$\hat{\vy}, \hat{\vy}^* \in [0, 1]^{N \times K}$ are the predicted and
posterior distributions, respectively.  The target histogram is denoted by
$\vt \in [0, 1]^K$.  The first term contains unary potentials that anchor
the posteriors to the predictions.  The second term is a symmetric $\chi^2$
distance to promote proximity between source and target histograms.  Weight
$\lambda$ defines relative importance of histogram matching.  We estimate the
source histogram as
   $\langle \hat{\vy}^* \rangle = \frac 1N \sum_n \hat{\vy}^*_n$.
We parameterize the posterior for all pixels $n$ as:
   $\hat{\vy}^*_n = \mathrm{softmax}(\log \hat{\vy}_n + \vb)$,
where the vector $\vb \in \mathbb{R}^K$ can be seen as a global bias for each
bin.  It is also possible to solve for the posteriors directly; this does not
perform  better quantitatively and is more prone to introducing artifacts.
We solve for $\vb$ using gradient descent on $E$ and use the resulting
posteriors in place of the predictions.  In the case of marginal
histograms, the optimization is run twice, once for each color channel.

\subsection{Neural Network Architecture}

Our base network is a fully convolutional version of VGG-16~\cite{vgg16} with
two changes: (1)~the classification layer ($\texttt{fc8}$) is discarded, and
(2)~the first filter layer (\texttt{conv1\_1}) operates on a single intensity
channel instead of mean-subtracted RGB.  We extract a hypercolumn descriptor
for a pixel by concatenating the features at its spatial location in all
layers, from \texttt{data} to \texttt{conv7} (\texttt{fc7}), resulting in a
12,417 channel descriptor.  We feed this hypercolumn into a fully connected
layer with $1024$ channels (\texttt{h\_fc1} in \cref{fig:schematic}), to
which we connect output predictors.

\begin{figure}[!th]
    \begin{center}
    \begin{minipage}[b]{0.156\linewidth}
        \vspace{0pt}
        \begin{center}
            \includegraphics[width=\textwidth]{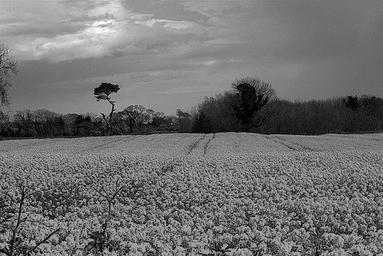}\\
            \includegraphics[width=\textwidth]{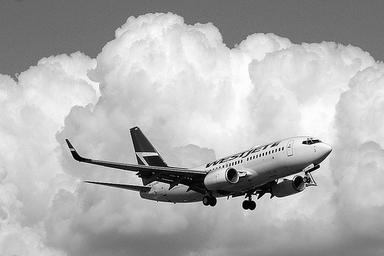}\\
            \includegraphics[width=\textwidth]{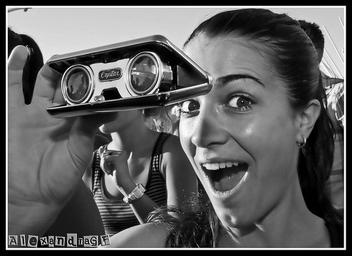}\\
            \includegraphics[width=\textwidth]{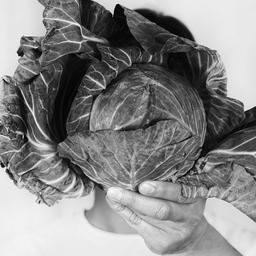}\\
            \includegraphics[width=\textwidth]{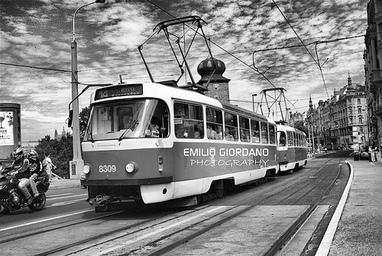}\\
            \includegraphics[width=\textwidth]{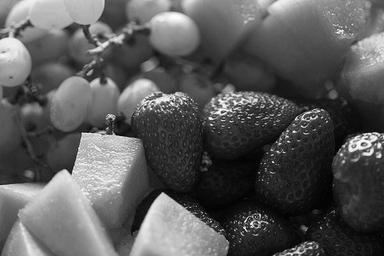}\\
            \scriptsize{\textbf{\textsf{Input}}}
        \end{center}
    \end{minipage}
    \begin{minipage}[b]{0.156\linewidth}
        \vspace{0pt}
        \begin{center}
            \includegraphics[width=\textwidth]{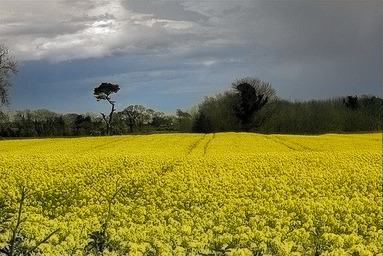}\\
            \includegraphics[width=\textwidth]{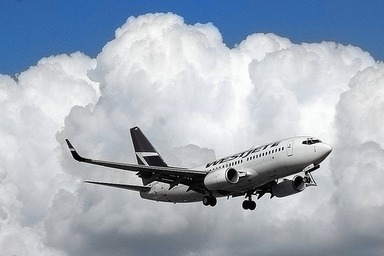}\\
            \includegraphics[width=\textwidth]{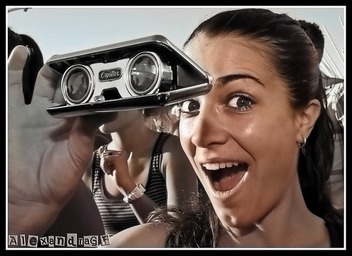}\\
            \includegraphics[width=\textwidth]{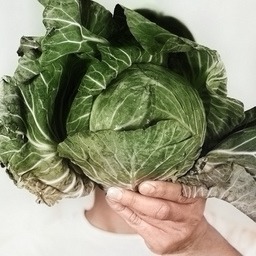}\\
            \includegraphics[width=\textwidth]{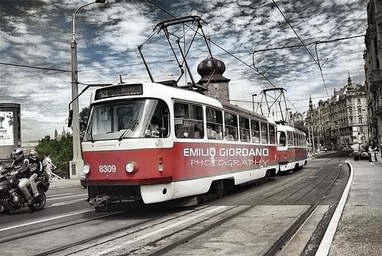}\\
            \includegraphics[width=\textwidth]{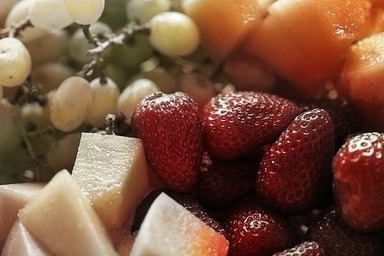}\\
            \scriptsize{\textbf{\textsf{Our Method}}}
        \end{center}
    \end{minipage}
    \begin{minipage}[b]{0.156\linewidth}
        \vspace{0pt}
        \begin{center}
            \includegraphics[width=\textwidth]{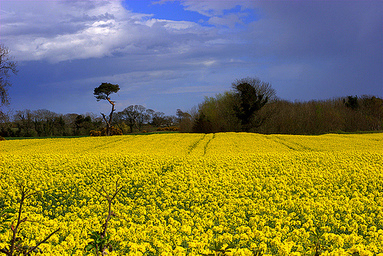}\\
            \includegraphics[width=\textwidth]{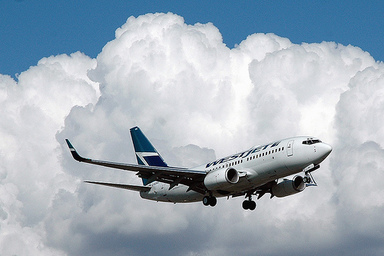}\\
            \includegraphics[width=\textwidth]{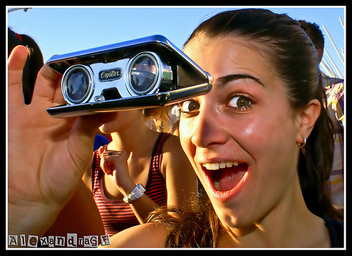}\\
            \includegraphics[width=\textwidth]{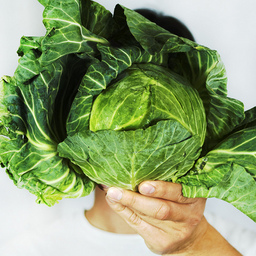}\\
            \includegraphics[width=\textwidth]{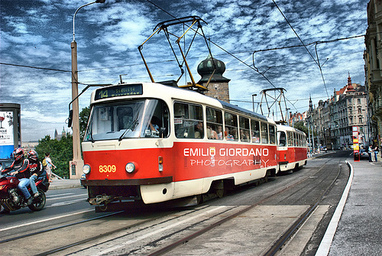}\\
            \includegraphics[width=\textwidth]{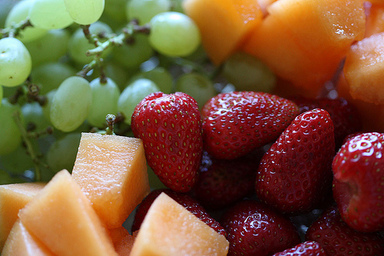}\\
            \scriptsize{\textbf{\textsf{Ground-truth}}}
        \end{center}
    \end{minipage}
    \hfill
    \begin{minipage}[b]{0.156\linewidth}
        \vspace{0pt}
        \begin{center}
            \includegraphics[width=\textwidth]{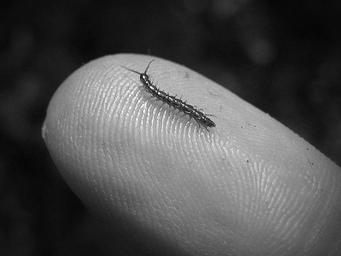}\\
            \vspace{0.0075\linewidth}
            \includegraphics[width=\textwidth]{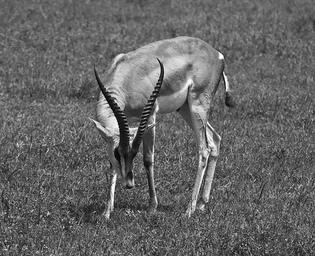}\\
            \vspace{0.0075\linewidth}
            \includegraphics[width=\textwidth]{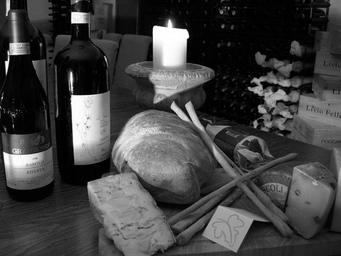}\\
            \vspace{0.0075\linewidth}
            \includegraphics[width=\textwidth]{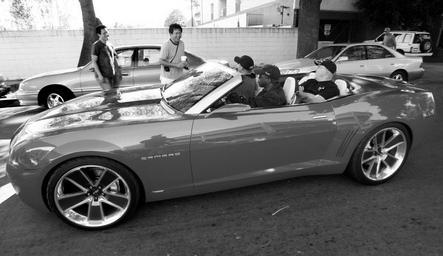}\\
            \vspace{0.0075\linewidth}
            \includegraphics[width=\textwidth]{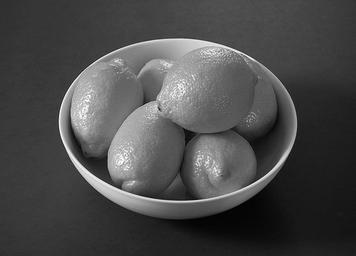}\\
            \vspace{0.0075\linewidth}
            \includegraphics[width=\textwidth]{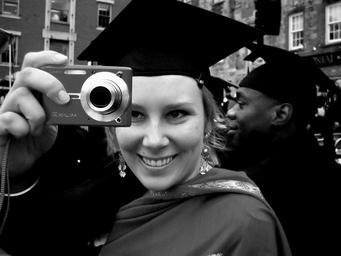}\\
            \scriptsize{\textbf{\textsf{Input}}}
        \end{center}
    \end{minipage}
    \begin{minipage}[b]{0.156\linewidth}
        \vspace{0pt}
        \begin{center}
            \includegraphics[width=\textwidth]{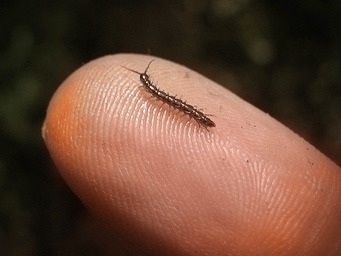}\\
            \vspace{0.0075\linewidth}
            \includegraphics[width=\textwidth]{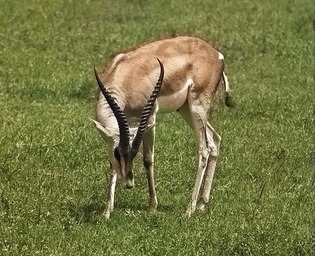}\\
            \vspace{0.0075\linewidth}
            \includegraphics[width=\textwidth]{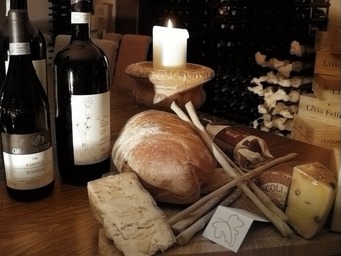}\\
            \vspace{0.0075\linewidth}
            \includegraphics[width=\textwidth]{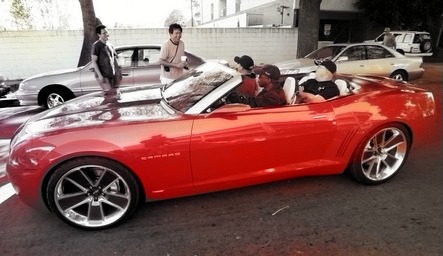}\\
            \vspace{0.0075\linewidth}
            \includegraphics[width=\textwidth]{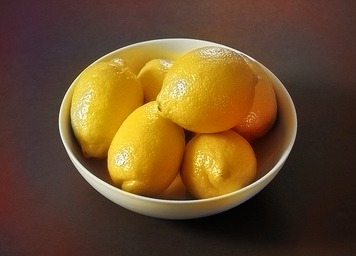}\\
            \vspace{0.0075\linewidth}
            \includegraphics[width=\textwidth]{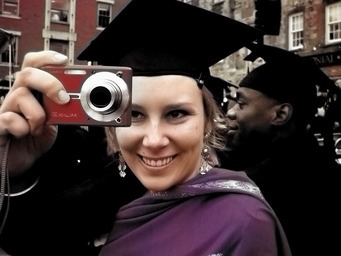}\\
            \scriptsize{\textbf{\textsf{Our Method}}}
        \end{center}
    \end{minipage}
    \begin{minipage}[b]{0.156\linewidth}
        \vspace{0pt}
        \begin{center}
            \includegraphics[width=\textwidth]{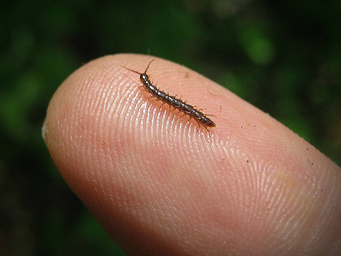}\\
            \vspace{0.0075\linewidth}
            \includegraphics[width=\textwidth]{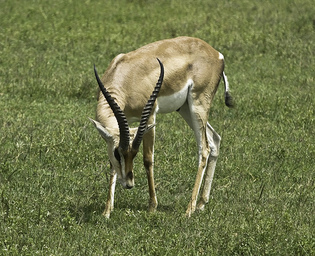}\\
            \vspace{0.0075\linewidth}
            \includegraphics[width=\textwidth]{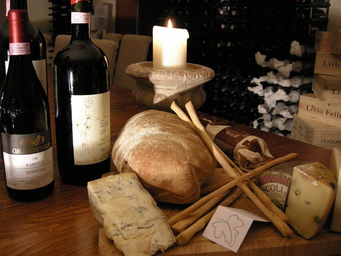}\\
            \vspace{0.0075\linewidth}
            \includegraphics[width=\textwidth]{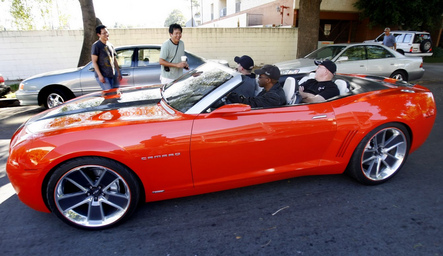}\\
            \vspace{0.0075\linewidth}
            \includegraphics[width=\textwidth]{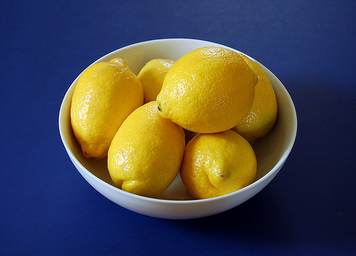}\\
            \vspace{0.0075\linewidth}
            \includegraphics[width=\textwidth]{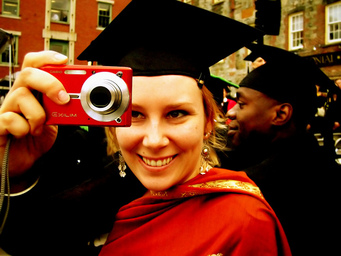}\\
            \scriptsize{\textbf{\textsf{Ground-truth}}}
        \end{center}
    \end{minipage}
    \end{center}
    \caption[Fully automatic colorization results on ImageNet/ctest10k]{
        \textbf{Fully automatic colorization results on ImageNet/ctest10k.}
        Our system reproduces known object color properties (\eg, faces, sky,
        grass, fruit, wood), and coherently picks colors for objects without
        such properties (\eg, clothing).
    }
    \label{fig:ctest10k-examples}
\end{figure}

We initialize with a version of VGG-16 pretrained on ImageNet, adapting it to
grayscale by averaging over color channels in the first layer and rescaling
appropriately.  Prior to training for colorization, we further fine-tune the
network for one epoch on the ImageNet classification task with grayscale
input.  As the original VGG-16 was trained without batch normalization~\cite{
ioffe2015batch}, scale of responses in internal layers can vary dramatically,
presenting a problem for learning atop their hypercolumn concatenation.
Liu~\etal~\cite{liu2015parsenet} compensate for such variability by applying
layer-wise $L_2$ normalization.  We use an alternative that we call
re-balancing the network.

\subsection{Network Re-Balancing}
\label{sec:rebalancing}

Before colorization training begins, we take the ImageNet pretrained network and shift
weights around such that each layer's activation has roughly unit second moment
($\mathbb{E}[X^2] \approx 1$). This is made possible thanks to ReLU
activations, which are piecewise linear around zero. To adjust the scale of the
activations of layer $l$ by factor $m$, without changing any other layer's
activation, the weights $\vW$ and the biases $\vb$ are updated according to:
\begin{equation}
    \vW_l \leftarrow m \vW_l, \quad\,\,
    \vb_l \leftarrow m \vb_l, \quad\,\,
    \vW_{l+1} \leftarrow \frac 1m \vW_{l+1}
\end{equation}
It is easy to show that the activations of layer $l+1$ are invariant to $m$,
while the activations of layer $l$ change by a factor of $m$. We
set $m = \frac{1}{\sqrt{\hat{\mathbb{E}}[X^2]}}$, estimated for each layer
separately.

\begin{figure}
    \centering
    \input{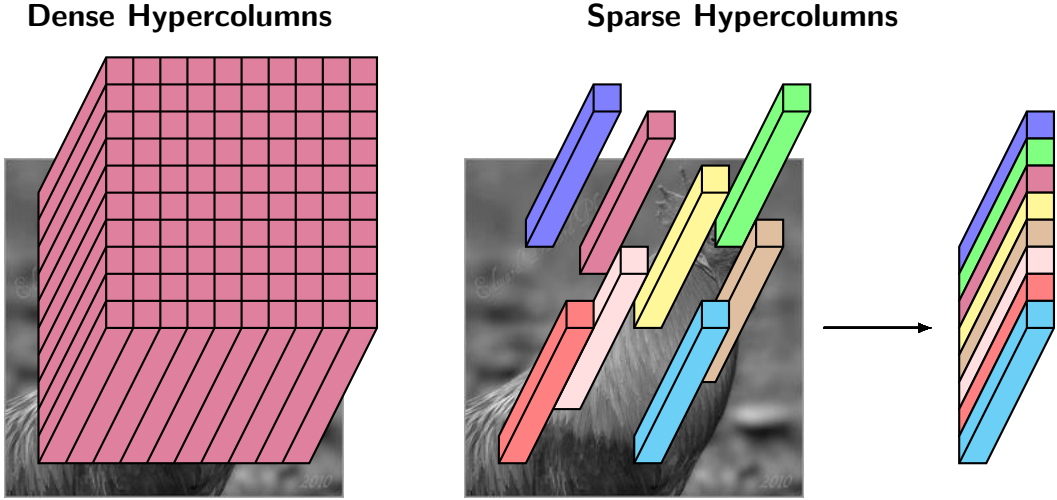}
    \caption[Dense vs.\ sparse hypercolumns]{
        \textbf{Dense vs.\ sparse hypercolumns.}
        To compute a hypercolumn for each pixel in an image, all intermediate layers are resampled and concatenated (left). This has a large memory footprint, but is necessary if we want to colorize all pixels in an image. However, during training, we have found that a sparse sample of hypercolumns that are computed with direct bilinear sampling works equally well and reduces memory usage (right). See \cref{tab:sparse} for experimental results.
    }
    \label{fig:sparse}
\end{figure}


\subsection{Dense vs.\ Sparse Sampling}
\label{sec:sparse-hypercolumns}

Computing the hypercolumn for all pixels in an image can be done by
re-sampling all layers to a fixed size and concatenating along the feature
axis. The fixed size does not have to be the size of the original input
and can instead be smaller. For instance, to make the colorizer fast at test time, we
recommend setting the target size to a factor 4 smaller than the input size.
The image containing the two color channels are finally upsampled and combined
with the original-sized lightness channel. Color channels contain far less
high-frequency signal than the intensity channel, so this incurs a minimum
degradation to the results.

Producing a densely sampled map of hypercolumns is quite fast. At least if we
use bilinear resampling and execute it on the GPU with fast convolutional
kernels. However, the memory footprint is high and a single hypercolumn
together with subsequent layer \texttt{h\_fc1} (with gradients at single precision) takes about
$100$~KiB. A dense sampling of a $256\times256$ image would take $6.7$~GiB.
Reducing the target image to $64\times64$ reduces it to $430$~MiB. However, this
is per image, so even a batch size of 10 would fill a third of our memory on a
$12$~GiB GPU. Even worse, depending on the implementation and the software used,
intermediate values prior to concatenation often require space too.

\begin{table}
    \centering
    \begin{minipage}{0.46\linewidth}
        \vspace{0pt}
        \begin{tabular*}{\columnwidth}{lr}
            \toprule
            Samples/image & Segmentation (\%mIU) \\ \midrule
            1                 & $54.5 \pm 0.4$       \\
            4                 & $63.6 \pm 0.4$       \\
            16                & $67.6 \pm 0.4$       \\
            64                & $68.5 \pm 0.2$       \\
            256               & $69.2 \pm 0.3$       \\
            1024              & $68.9 \pm 0.5$       \\
            \bottomrule
        \end{tabular*}
    \end{minipage}
    \quad
    \begin{minipage}{0.46\linewidth}
        \vspace{0pt}
        \caption[Hypercolumn sparsity]{
            \textbf{Hypercolumn sparsity.} We evaluate on VOC 2012 semantic segmentation
            (val) the effectiveness of sparse samples. We train on large inputs
            ($448 \times 448$) made possible by the memory savings.  Measurable
            benefits stop around 256 samples. This is only 0.1\% of the pixels in
            the image and still only 2\% if we downsample our dense map by a factor 4.
        }
        \label{tab:sparse}
    \end{minipage}
\end{table}

Instead, we have discovered that extracting hypercolumns at only a sparse set of
locations is good enough during training.  The values of each hypercolumn can be
computed directly using bilinear interpolation and organized as if it were
non-spatial data (see \cref{fig:sparse}).
A custom layer performs the
sparse extraction, which we have made available for both Caffe~\cite{jia2014caffe} and TensorFlow~\cite{tensorflow2015-whitepaper-full}.\footnote{
   \texttt{https://github.com/gustavla/autocolorize}
}
In the backward pass, gradients are propagated to the four closest spatial
cells involved in the bilinear interpolation. Locks ensure atomicity of
gradient updates, without incurring any measurable performance penalty.

We investigate in \cref{tab:sparse} how many samples suffice when applying
sparse samples on the VOC 2012 Semantic Segmentation~\cite{pascal-voc-2012} benchmark. We train on
large inputs ($448\times448$), afforded by the sparse sampling. After around
256 samples, there is no gain to optimization by increasing sample locations further.
This is likely due to the high correlation between the hypercolumns of a single image.
The top convolutional layers (\texttt{fc6}/\texttt{conv6} and
\texttt{fc7}/\texttt{conv7}) arguably need the most samples to converge properly.
However, even at our input size, these layers are reduced to $14\times14$
spatial dimensions, containing only 192 distinct elements. At 256 samples,
we thus get a complete sample of the high level features, at the cost of only
$26$~MiB per image.

Sparse sampling also makes it easy and cheap to extract pairs or triplets of
hypercolumns. In \cref{sec:proxy-tasks} we use this to predict
relative optical flow between pairs of locations.

\begin{figure}
    \begin{center}
    \begin{minipage}[b]{0.157\linewidth}
        \begin{center}
            \includegraphics[width=\textwidth]{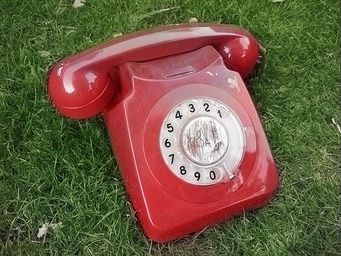}\\
            \includegraphics[width=\textwidth]{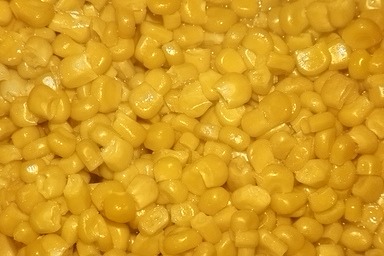}\\
            \includegraphics[width=\textwidth]{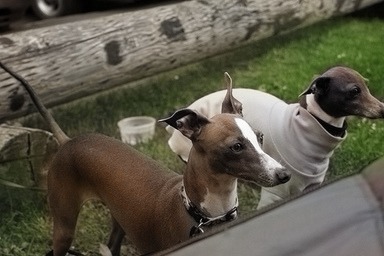}\\
            \includegraphics[width=\textwidth]{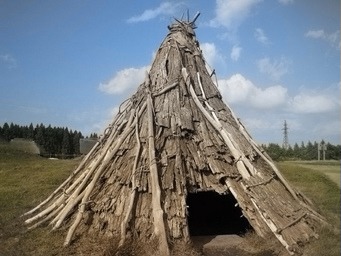}\\
            \includegraphics[width=\textwidth]{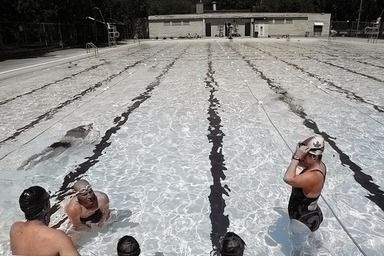}\\
            \includegraphics[width=\textwidth]{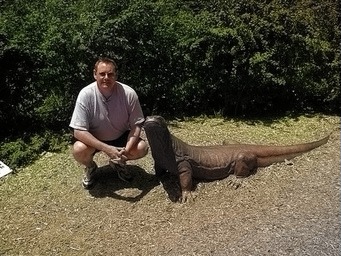}\\
        \end{center}
    \end{minipage}
    \hfill
    \begin{minipage}[b]{0.1585\linewidth}
        \begin{center}
            \includegraphics[width=\textwidth]{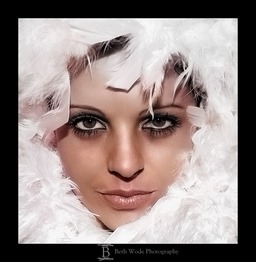}\\
            \includegraphics[width=\textwidth]{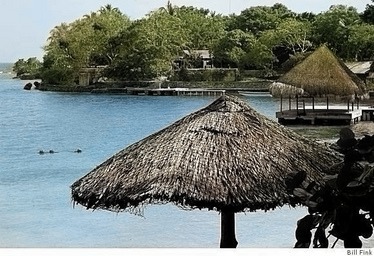}\\
            \includegraphics[width=\textwidth]{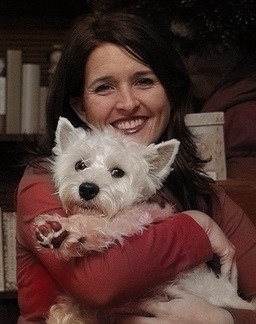}\\
            \includegraphics[width=\textwidth]{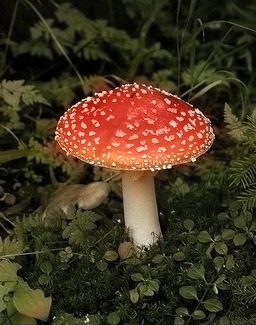}\\

        \end{center}
    \end{minipage}
    \hfill
    \begin{minipage}[b]{0.1575\linewidth}
        \begin{center}
            \includegraphics[width=\textwidth]{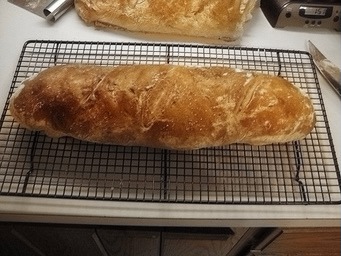}\\
            \includegraphics[width=\textwidth]{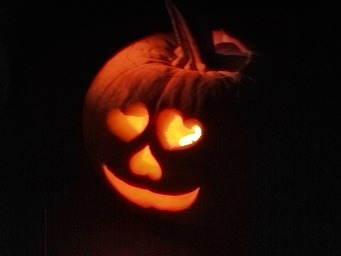}\\
            \includegraphics[width=\textwidth]{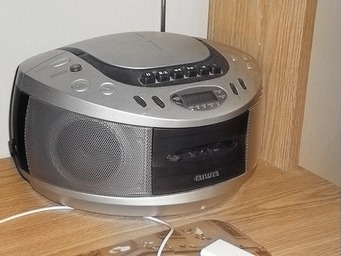}\\
            \includegraphics[width=\textwidth]{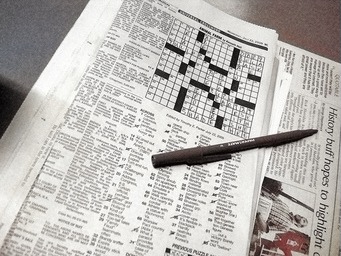}\\
            \includegraphics[width=\textwidth]{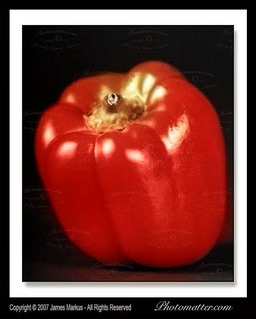}\\
        \end{center}
    \end{minipage}
    \hfill
    \begin{minipage}[b]{0.1575\linewidth}
        \begin{center}
            \includegraphics[width=\textwidth]{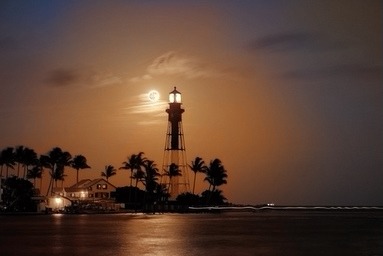}\\
            \includegraphics[width=\textwidth]{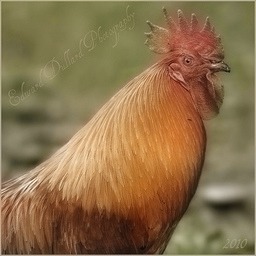}\\
            \includegraphics[width=\textwidth]{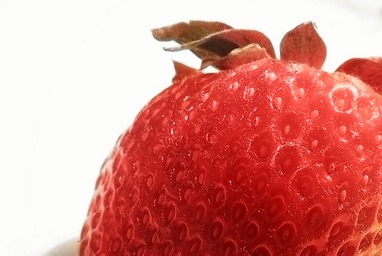}\\
            \includegraphics[width=\textwidth]{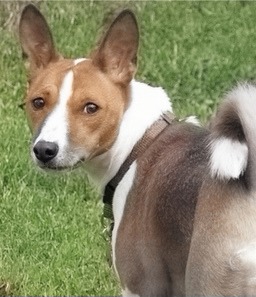}\\
            \includegraphics[width=\textwidth]{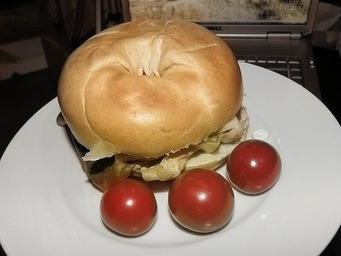}\\
        \end{center}
    \end{minipage}
    \hfill
    \begin{minipage}[b]{0.162\linewidth}
        \begin{center}
            \includegraphics[width=\textwidth]{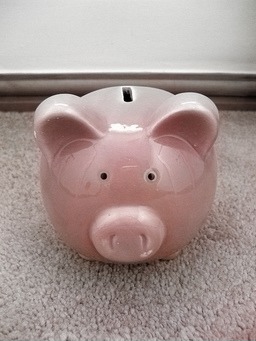}\\
            \includegraphics[width=\textwidth]{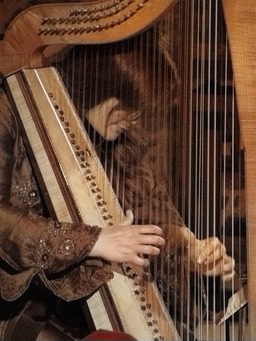}\\
            \includegraphics[width=\textwidth]{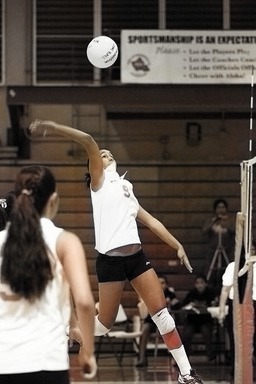}\\
        \end{center}
    \end{minipage}
    \hfill
    \begin{minipage}[b]{0.162\linewidth}
        \begin{center}
            \includegraphics[width=\textwidth]{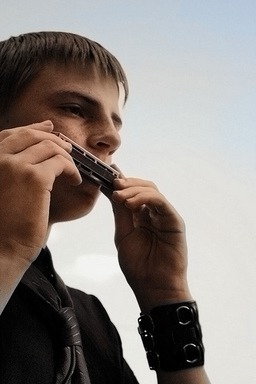}\\
            \includegraphics[width=\textwidth]{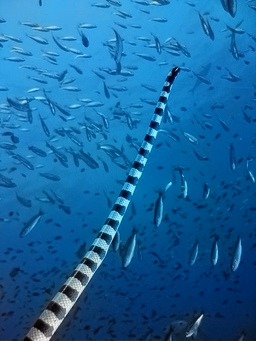}\\
            \includegraphics[width=\textwidth]{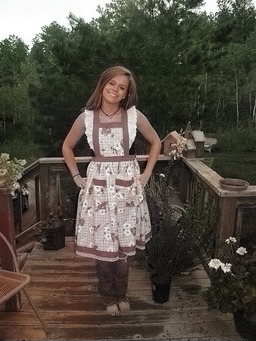}\\
        \end{center}
    \end{minipage}
    \end{center}
    \vspace{-0.05\linewidth}
    \rule{1.0\linewidth}{0.5pt}
    \vspace{-0.076\linewidth}
    \begin{center}
    \begin{minipage}[t]{0.3425\linewidth}
        \vspace{0pt}
        \begin{center}
            \includegraphics[width=.485\textwidth]{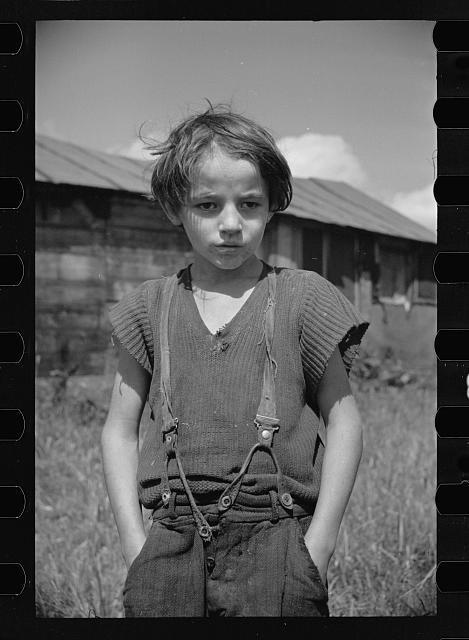}
            \includegraphics[width=.485\textwidth]{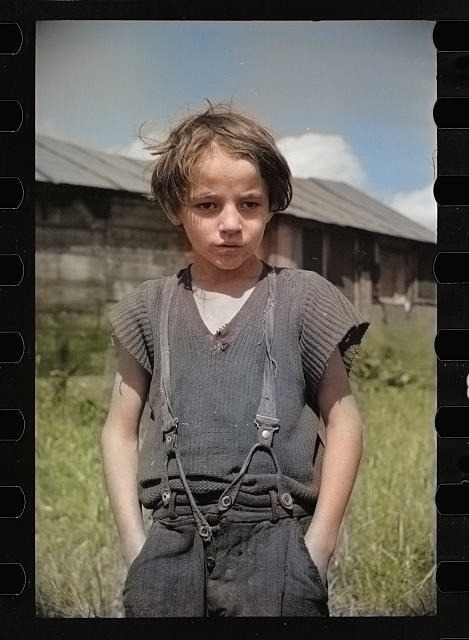}
        \end{center}
    \end{minipage}
    \hfill
    \begin{minipage}[t]{0.32\linewidth}
        \vspace{0pt}
        \begin{center}
            \includegraphics[width=.48\textwidth]{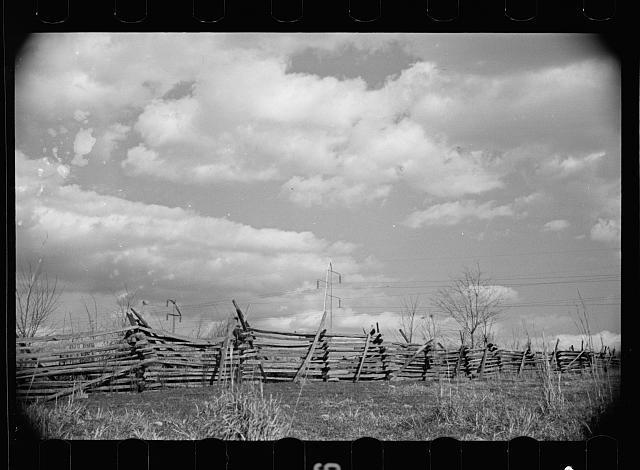}
            \includegraphics[width=.48\textwidth]{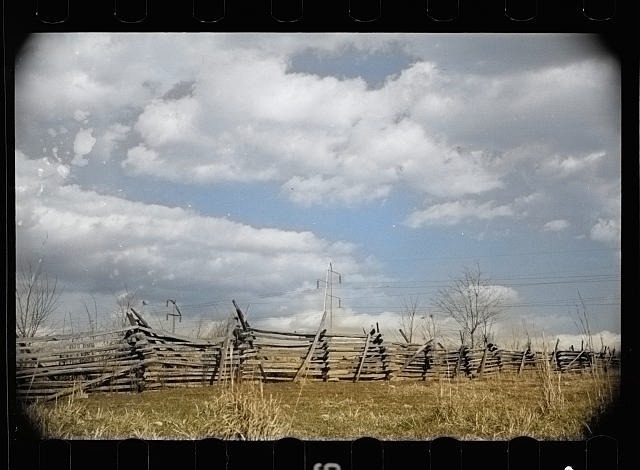}\\
            \includegraphics[width=.48\textwidth]{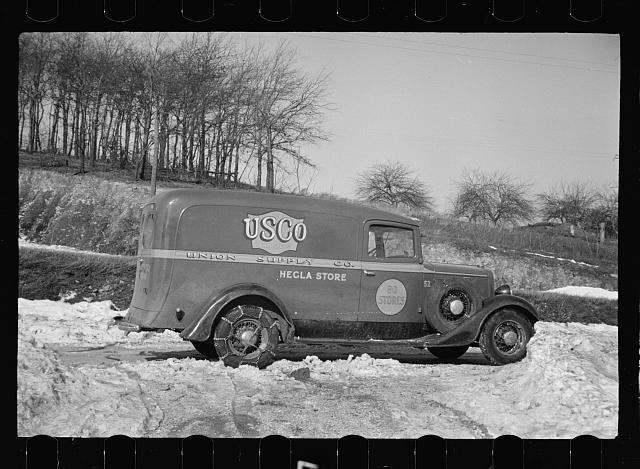}
            \includegraphics[width=.48\textwidth]{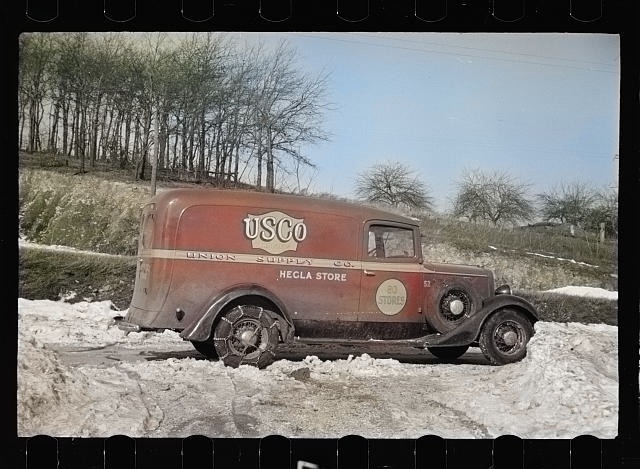}
        \end{center}
    \end{minipage}
    \hfill
    \begin{minipage}[t]{0.32\linewidth}
        \vspace{0pt}
        \begin{center}
            \includegraphics[width=.48\textwidth]{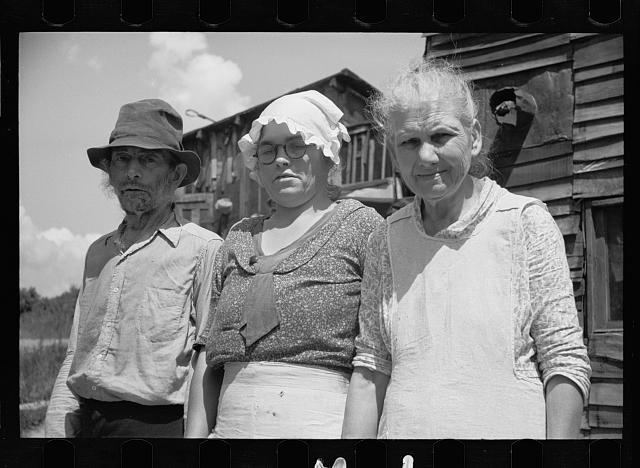}
            \includegraphics[width=.48\textwidth]{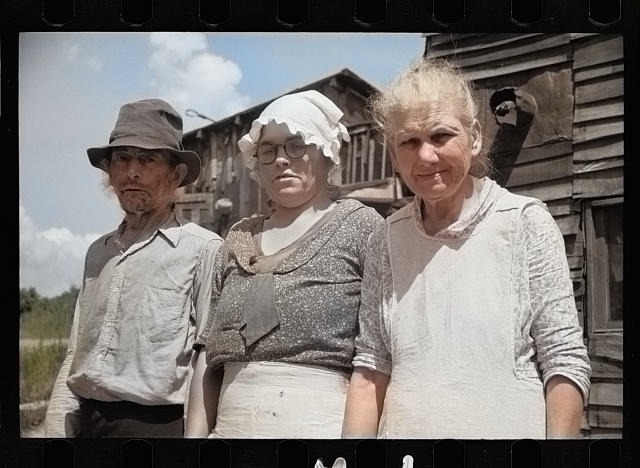}\\
            \includegraphics[width=.48\textwidth]{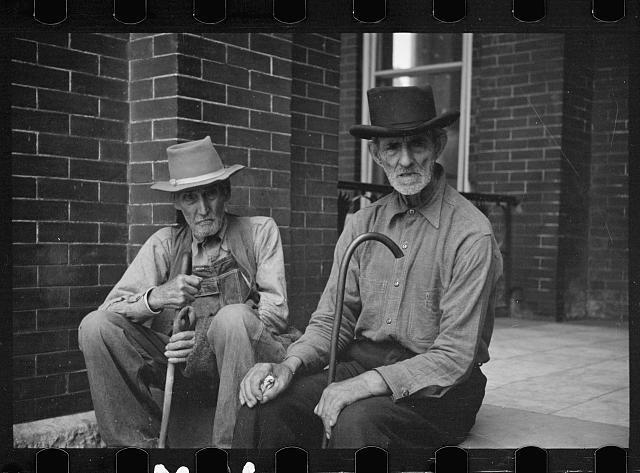}
            \includegraphics[width=.48\textwidth]{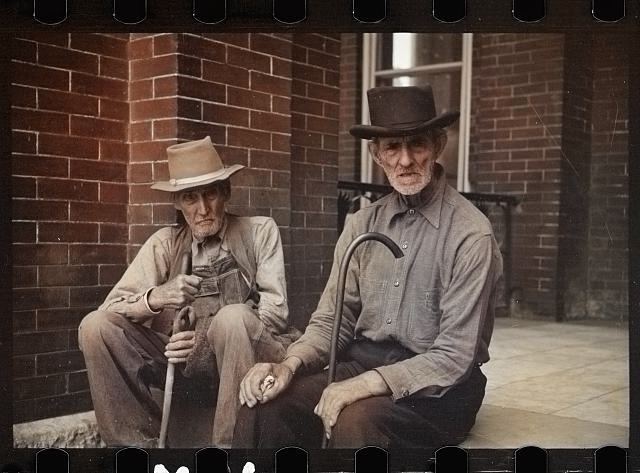}
        \end{center}
    \end{minipage}
    \end{center}
    \caption[Additional colorization results]{
        \textbf{Additional colorization results.}
        \emph{Top:}
            Our automatic colorizations of these ImageNet examples
            are difficult to distinguish from real color images.
        \emph{Bottom:} B\&W photographs.
    }
    \label{fig:ctest10k-more-examples}
\end{figure}

\begin{figure}
    \setlength\fboxsep{0pt}
    \begin{center}
        \begin{minipage}[t]{0.1625\linewidth}
        \begin{center}
            \begin{minipage}[t]{0.975\linewidth}
            \begin{center}
                \includegraphics[width=\textwidth]{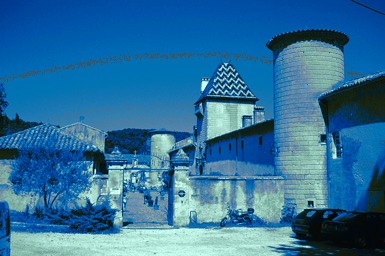}\\
                \includegraphics[width=\textwidth]{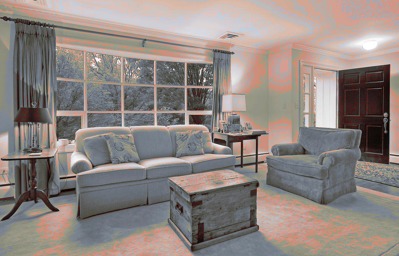}\\
                \includegraphics[width=\textwidth]{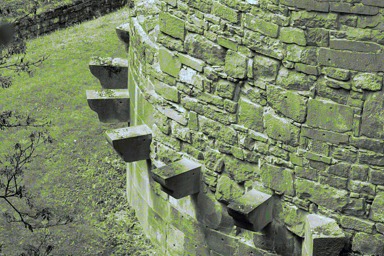}\\
                \scriptsize{\textbf{\textsf{Grayscale only}}}
            \end{center}
            \end{minipage}\\
            \rule{0.95\linewidth}{0.5pt}\\
            \scriptsize{\textbf{\textsf{Welsh~\etal~\cite{welsh2002transferring}}}}
        \end{center}
        \end{minipage}
        \hfill
        \begin{minipage}[t]{0.325\linewidth}
        \begin{center}
            \begin{minipage}[t]{0.4875\linewidth}
            \begin{center}
                \includegraphics[width=\textwidth]{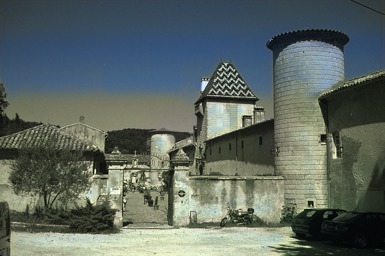}\\
                \includegraphics[width=\textwidth]{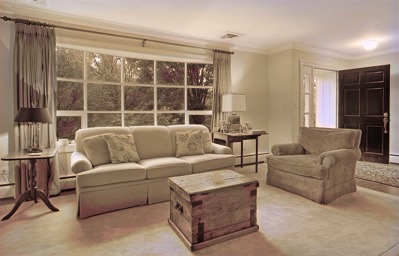}\\
                \includegraphics[width=\textwidth]{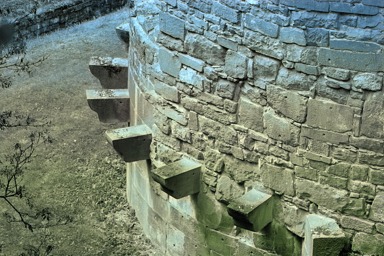}\\
                \scriptsize{\textbf{\textsf{\textcolor{white}{y}GT Scene\textcolor{white}{y}}}}
            \end{center}
            \end{minipage}
            \begin{minipage}[t]{0.4875\linewidth}
            \begin{center}
                \includegraphics[width=\textwidth]{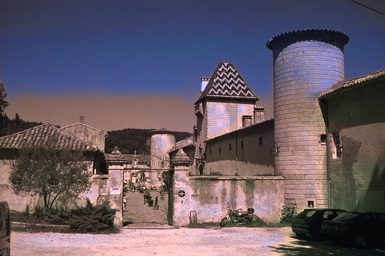}\\
                \includegraphics[width=\textwidth]{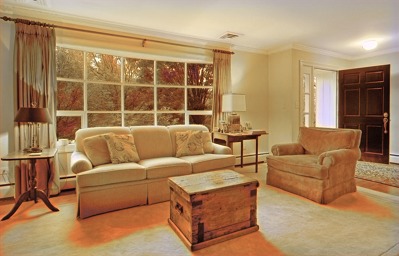}\\
                \includegraphics[width=\textwidth]{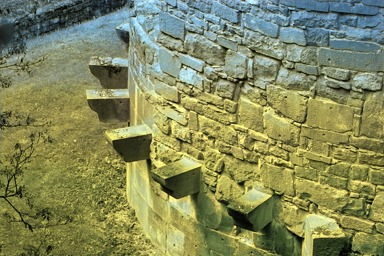}\\
                \scriptsize{\textbf{\textsf{GT Scene \& Hist}}}
            \end{center}
            \end{minipage}\\
            \rule{0.975\linewidth}{0.5pt}\\
            \scriptsize{\textbf{\textsf{Deshpande~\etal~\cite{deshpande2015learning}}}}
        \end{center}
        \end{minipage}
        \hfill
        \begin{minipage}[t]{0.325\linewidth}
        \begin{center}
            \begin{minipage}[t]{0.4875\linewidth}
            \begin{center}
                \includegraphics[width=\textwidth]{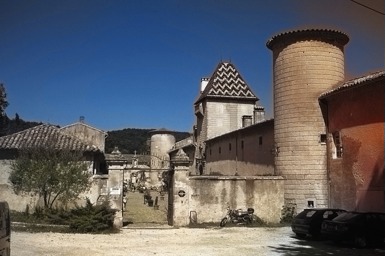}\\
                \includegraphics[width=\textwidth]{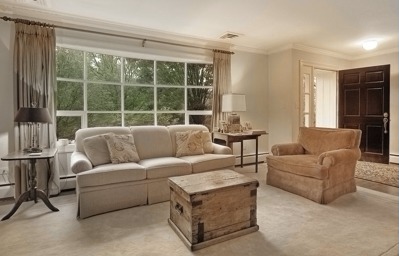}\\
                \includegraphics[width=\textwidth]{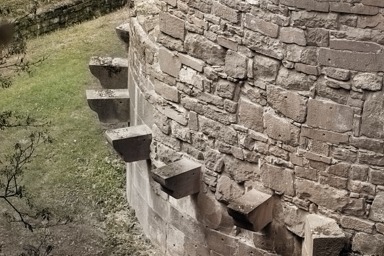}\\
                \scriptsize{\textbf{\textsf{Grayscale only}}}
            \end{center}
            \end{minipage}
            \begin{minipage}[t]{0.4875\linewidth}
            \begin{center}
                \includegraphics[width=\textwidth]{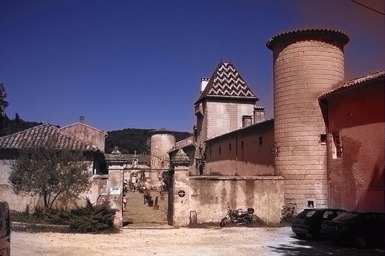}\\
                \includegraphics[width=\textwidth]{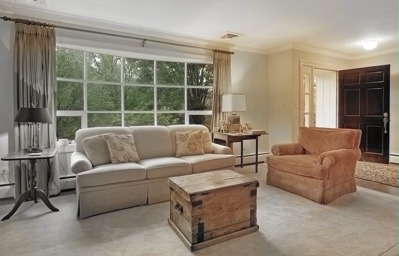}\\
                \includegraphics[width=\textwidth]{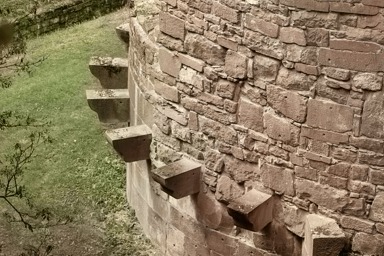}\\
                \scriptsize{\textbf{\textsf{GT Histogram}}}
            \end{center}
            \end{minipage}\\
            \rule{0.975\linewidth}{0.5pt}\\
            \scriptsize{\textbf{\textsf{Our Method}}}
        \end{center}
        \end{minipage}
        \hfill
        \begin{minipage}[t]{0.1625\linewidth}
        \begin{center}
            \begin{minipage}[t]{0.975\linewidth}
            \begin{center}
                \includegraphics[width=\textwidth]{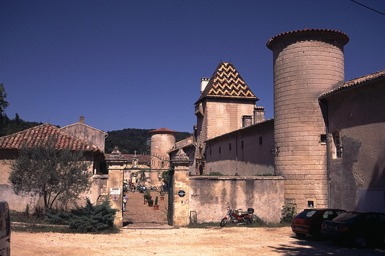}\\
                \includegraphics[width=\textwidth]{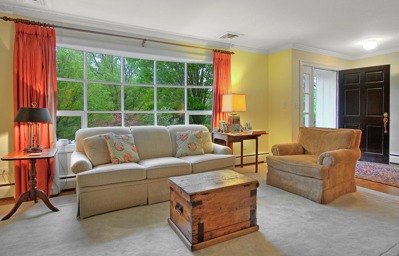}\\
                \includegraphics[width=\textwidth]{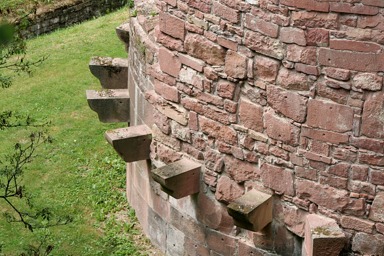}\\
                \scriptsize{\textbf{\textsf{Ground-truth}}}
            \end{center}
            \end{minipage}
        \end{center}
        \end{minipage}
    \end{center}
    \caption[Examples on SUN-6]{
        \textbf{Examples on SUN-6.}
        GT Scene: test image scene class is available.
        GT Hist: test image color histogram is available.
        We obtain colorizations with visual quality better than those from
        prior work, even though we do not exploit reference images or known
        scene class.  Our energy minimization method
        (Section~\ref{sec:method-transfer}) for GT Hist further
        improves results.  In either mode, our method appears less dependent
        on spatial priors: note splitting of the sky in the first row and
        correlation of green with actual grass in the last row.
    }
    \label{fig:sun6-examples}
\end{figure}

\section{Experiments}
\label{sec:color-experiments}

Starting from pretrained VGG-16-Gray, described in the previous
section, we attach \texttt{h\_fc1} and output prediction layers with
Xavier initialization~\cite{glorot2010understanding}, and fine-tune the
entire system for colorization.  We consider multiple prediction layer
variants: Lab output with $L_2$ loss, and both Lab and hue/chroma marginal or
joint histogram output with losses according to
\cref{eq:loss-hist,eq:loss-hc}.  We train each system
variant end-to-end for one epoch on the 1.3M images of the ImageNet
training set, each resized to at most $256$ pixels in smaller dimension.
A single epoch takes approximately 17 hours on a GTX Titan X GPU. At test
time, colorizing a single $512 \times 512$ pixel image takes $0.5$ seconds.

\begin{table}[t!]
\begin{minipage}[t]{.52\textwidth}
\centering
    \begin{tabular*}{\textwidth}{l @{\extracolsep{\fill}} rr}
        \toprule
        Model$\backslash$Metric  & RMSE        & PSNR        \\
        \midrule
        No colorization          & 0.343       & 22.98       \\ 
        Lab, $L_2$               & 0.318       & 24.25       \\ 
        Lab, $K = 32$            & 0.321       & 24.33       \\ 
        Lab, $K = 16 \times 16$  & 0.328       & 24.30       \\ 
        Hue/chroma, $K = 32$     & 0.342       & 23.77       \\ 
        \quad + chromatic fading & {\bf 0.299} & {\bf 24.45} \\
        \bottomrule
    \end{tabular*}
    \caption[Validation on ImageNet/cval1k]{
        \textbf{ImageNet/cval1k.}
        Validation performance of system variants.
        Hue/chroma is best, but only with chromatic fading.
    }
    \label{tab:cval1k}
\end{minipage}
\hfill
\begin{minipage}[t]{.45\textwidth}
\centering
    \begin{tabular*}{\textwidth}{l @{\extracolsep{\fill}} rr}
        \toprule
        Model$\backslash$Metric & RMSE        & PSNR        \\
        \midrule
        \texttt{data..fc7}      & {\bf 0.299} & {\bf 24.45} \\ 
        \texttt{data..conv5\_3} & 0.306       & 24.13       \\ 
        \texttt{conv4\_1..fc7}  & 0.302       & {\bf 24.45} \\ 
        \texttt{conv5\_1..fc7}  & 0.307       & 24.38       \\ 
        \texttt{fc6..fc7}       & 0.323       & 24.22       \\ 
        \texttt{fc7}            & 0.324       & 24.19       \\ 
        \bottomrule
    \end{tabular*}
    \caption[Ablation study of hypercolumn components]{
        \textbf{ImageNet/cval1k.}
        Ablation study of hypercolumn components.
    }
    \label{tab:ablation}
\end{minipage}
\end{table}

We create two disjoint subsets of the ImageNet validation data for our own use:
1,000 validation images (\textbf{cval1k}) and 10,000 test images
(\textbf{ctest10k}).  Each set has a balanced representation for ImageNet
categories, and excludes any images encoded as grayscale, but may include
images that are naturally grayscale (\eg, closeup of nuts and bolts), where
an algorithm should know not to add color.  Category labels are
discarded; only images are available at test time.  We propose
\textbf{ctest10k} as a standard benchmark with the following metrics:
\begin{itemize}
    \item{
        \textbf{RMSE}:
        root mean square error in $\alpha\beta$ averaged over all
        pixels~\cite{deshpande2015learning}.
    }
    \item{
        \textbf{PSNR}:
        peak signal-to-noise ratio in RGB calculated per
        image~\cite{cheng2015deep}.  We use the arithmetic mean of PSNR over
        images, instead of the geometric mean as in
        Cheng~\etal~\cite{cheng2015deep};
        geometric mean is overly sensitive to outliers.
    }
\end{itemize}
By virtue of comparing to ground-truth color images, quantitative
colorization metrics can penalize reasonable, but incorrect, color guesses for
many objects (\eg, red car instead of blue car) more than jarring artifacts.
This makes qualitative results for colorization as important as quantitative;
we report both.

\Cref{fig:ctest10k-examples,fig:ctest10k-more-examples}
show example test results of our best system variant, selected according to
performance on the validation set and trained for a total of $10$ epochs.  This
variant predicts hue and chroma and
uses chromatic fading during image generation.

\Cref{tab:cval1k} provides
complete validation benchmarks for all system variants, including the trivial
baseline of no colorization (returning the grayscale input).  On ImageNet test
(\textbf{ctest10k}), our selected model obtains $0.293$ (RMSE, $\alpha\beta$,
avg/px) and $24.94$~dB (PSNR, RGB, avg/im), compared to $0.333$ and $23.27$~dB for
the no colorization baseline.

Table~\ref{tab:ablation} examines the importance of different neural network
layers to colorization; it reports validation performance of ablated systems
that include only the specified subsets of layers in the hypercolumn used to
predict hue and chroma.  Some lower layers may be discarded without much
performance loss, yet higher layers alone (\texttt{fc6..fc7}) are insufficient
for good colorization.

Our ImageNet colorization benchmark is new to a field lacking an
established evaluation protocol.  We therefore focus on comparisons with
two recent papers~\cite{deshpande2015learning,cheng2015deep}, using their
self-defined evaluation criteria.  To do so, we run our ImageNet-trained
hue and chroma model on two additional datasets:
\begin{itemize}
    \item{
        \textbf{\bf SUN-A}~\cite{patterson2014sun} is a subset of the SUN
        dataset~\cite{xiao2010sun} containing $47$ object categories.
        Cheng~\etal~\cite{cheng2015deep} train a colorization system on
        $2688$ images and report results on $1344$ test images.  We were
        unable to obtain the list of test images, and therefore report results
        averaged over five random subsets of $1344$ SUN-A images.  We do
        not use any SUN-A images for training.
    }
    \item{
        \textbf{SUN-6}, another SUN subset, used by
        Deshpande~\etal~\cite{deshpande2015learning}, includes images
        from six scene categories (beach, castle, outdoor, kitchen,
        living room, bedroom).  We compare our results on $240$ test images
        to those reported in~\cite{deshpande2015learning} for their method
        as well as for Welsh~\etal~\cite{welsh2002transferring} with
        automatically matched reference images as
        in~\cite{morimoto2009automatic}.
        Following~\cite{deshpande2015learning}, we consider another evaluation
        regime in which ground-truth target color histograms are available.
    }
\end{itemize}
\begin{table}[!t]
\begin{minipage}[t]{.40\textwidth}
    \begin{tabular*}{\textwidth}[t]{l @{\extracolsep{\fill}} r}
        \toprule
        Method                                       & RMSE           \\
        \midrule
        Grayscale (no colorization)                  & 0.285          \\
        Welsh~\etal~\cite{welsh2002transferring}     & 0.353          \\
        Deshpande~\etal~\cite{deshpande2015learning} & 0.262          \\
        \quad + GT Scene                             & 0.254          \\
        Our Method                                   & \textbf{0.211} \\
        \bottomrule
    \end{tabular*}
    \caption[Results on SUN-6]{
        \textbf{SUN-6.}
        Comparison with competing methods.
    }
    \label{tab:sun6}
\end{minipage}
\hfill
\begin{minipage}[t]{.56\textwidth}
    \begin{tabular*}{\textwidth}[t]{l @{\extracolsep{\fill}} r}
        \toprule
        Method                                           & RMSE \\
        \midrule
        Deshpande~\etal~(C)~\cite{deshpande2015learning} & 0.236 \\
        Deshpande~\etal~(Q)                              & 0.211 \\
        Our Method (Q)                                   & 0.178 \\
        Our Method (E)                                   & \textbf{0.165} \\
        \bottomrule
    \end{tabular*}
    \caption[Results on SUN-6 (with ground truth histograms)]{
        \textbf{SUN-6 (GT Hist).}
        Comparison using ground-truth histograms.
        Results for Deshpande~\etal~\cite{deshpande2015learning} use GT Scene.
    }
    \label{tab:sun6-gth}
\end{minipage}
\end{table}
\begin{figure}[!t]
    \begin{minipage}[t]{.463\linewidth}
        \vspace{0pt}
        \centering
        \includegraphics[width=.97\linewidth]{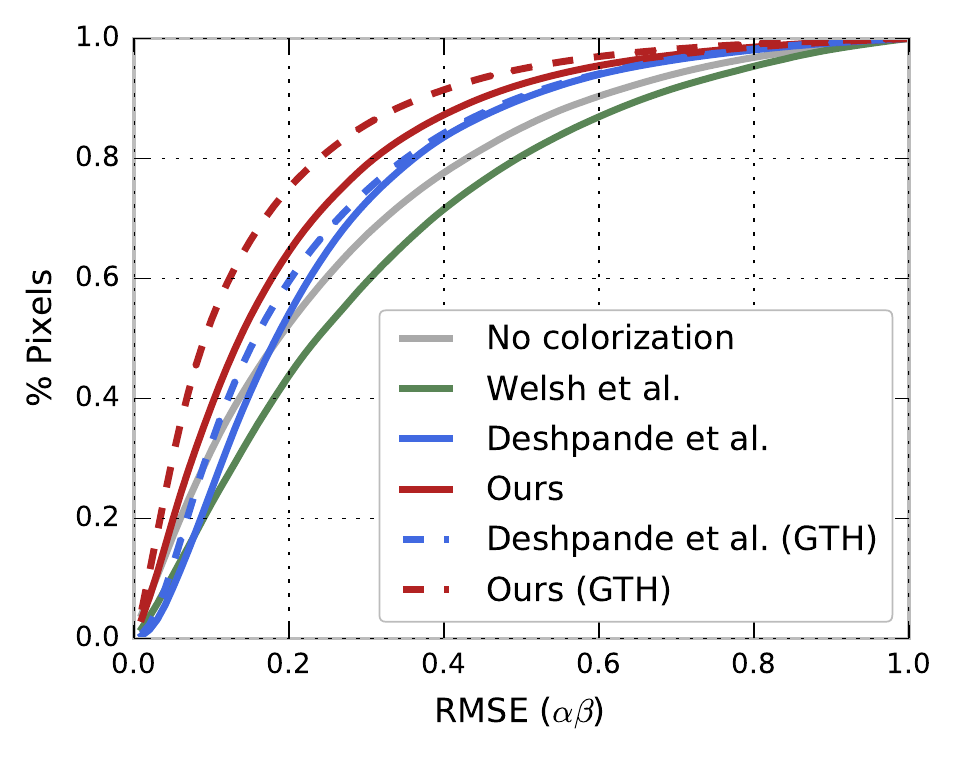}\\
        \caption[Cumulative histogram of per-pixel error on SUN-6]{
            \textbf{SUN-6.}
            Cumulative histogram of per-pixel error
            (higher means more pixels with lower error).
            Results for Deshpande~\etal~\cite{deshpande2015learning}
            use GT Scene.
        }
        \label{fig:sun6-cum-hist}
    \end{minipage}\hspace{1em}%
    \begin{minipage}[t]{.513\linewidth}
        \vspace{0pt}
        \centering
        \includegraphics[width=\linewidth]{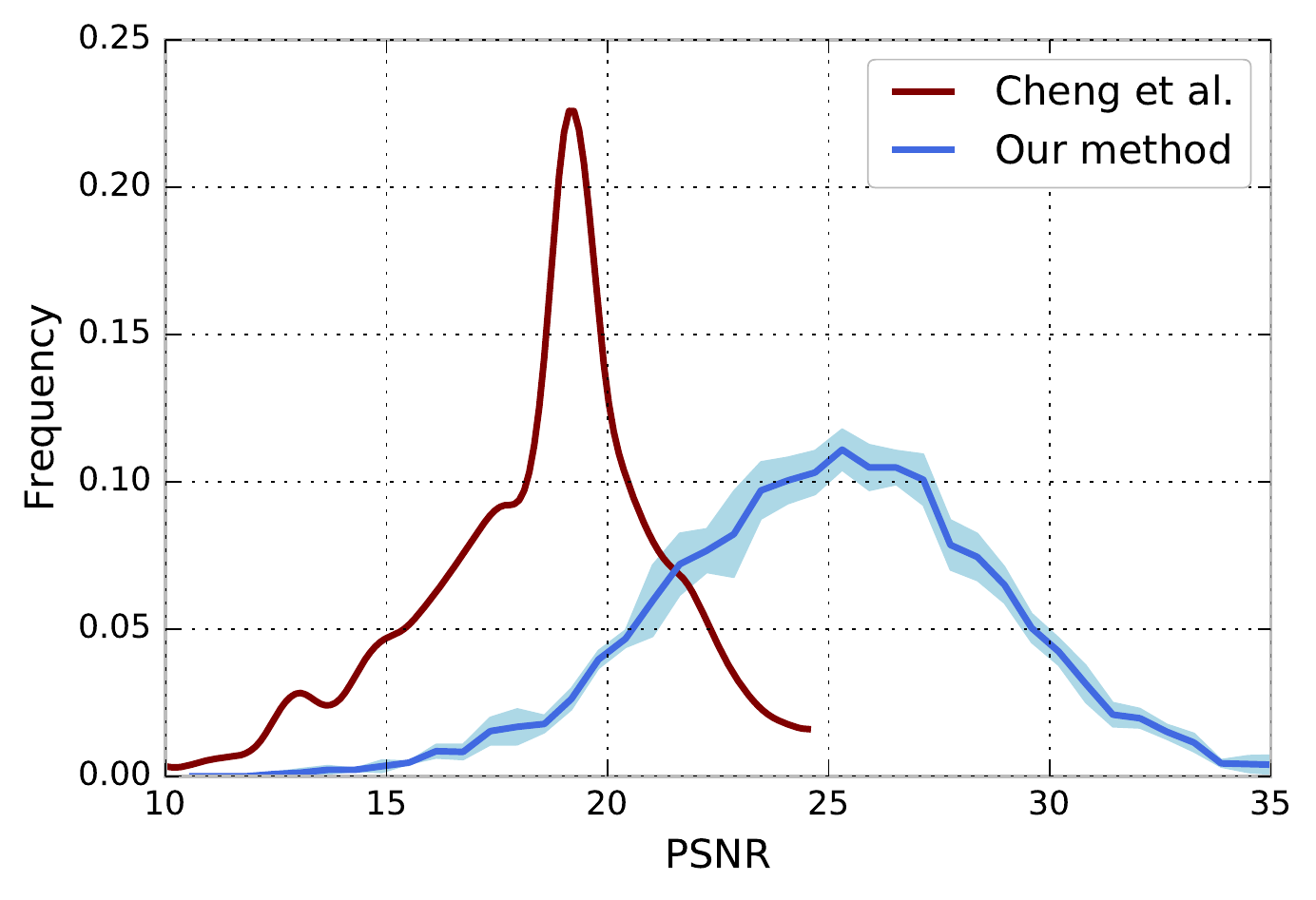}
        \caption[Histogram of per-image PSNR on SUN-A]{
            \textbf{SUN-A.}
            Histogram of per-image PSNR for~\cite{cheng2015deep} and our
            method. The highest geometric mean PSNR reported for experiments
            in~\cite{cheng2015deep} is 24.2, vs.\ our {$\mathbf{32.7}\pm2.0$}.
        }
        \label{fig:sun-deep-histogram}
\end{minipage}
\end{figure}
\Cref{fig:sun6-examples} shows a comparison of results on SUN-6.
Forgoing usage of ground-truth global histograms, our fully automatic
system produces output qualitatively superior to methods relying on
such side information.  \Cref{tab:sun6,tab:sun6-gth}
report quantitative performance corroborating this view.  The
partially automatic systems in \cref{tab:sun6-gth} adapt output
to fit global histograms using either: (C) cluster
correspondences~\cite{deshpande2015learning}, (Q) quantile matching, or
(E) our energy minimization described in \cref{sec:method-transfer}.
Our quantile matching results are superior to those
of~\cite{deshpande2015learning} and our new energy minimization procedure
offers further improvement.

Figures~\ref{fig:sun6-cum-hist} and~\ref{fig:sun-deep-histogram} compare
error distributions on SUN-6 and SUN-A.  As in Table~\ref{tab:sun6}, our
fully automatic approach dominates all competing methods, even those
which use auxiliary information.  It is only outperformed by the version of
itself augmented with ground-truth global histograms.  On SUN-A,
Figure~\ref{fig:sun-deep-histogram} shows clear separation between our
method and~\cite{cheng2015deep} on per-image PSNR.

With regard to concurrent work, Zhang~\etal~\cite{zhang2016colorful} include a
comparison of our results to their own.  The two systems are competitive in
terms of quantitative measures of colorization accuracy.  Their system, set to
produce more vibrant colors, has an advantage in terms of human-measured
preferences.  In contrast, an off-the-shelf VGG-16 network for image
classification, consuming our system's color output, more often produces
correct labels, suggesting a realism advantage.  We refer interested readers
to~\cite{zhang2016colorful} for the full details of this comparison.

\begin{figure}[!t]
    \begin{center}
    \begin{minipage}[b]{0.1880\linewidth}
        \vspace{0pt}
        \begin{center}
            \includegraphics[width=\textwidth]{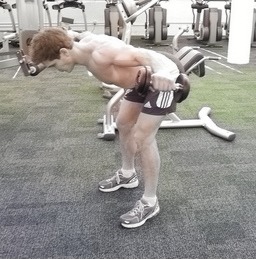}\\
            \includegraphics[width=\textwidth]{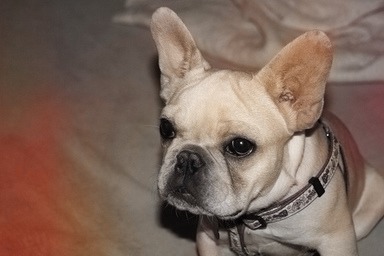}
        \end{center}
    \end{minipage}
    \hfill
    \begin{minipage}[b]{0.2045\linewidth}
        \vspace{0pt}
        \begin{center}
            \includegraphics[width=\textwidth]{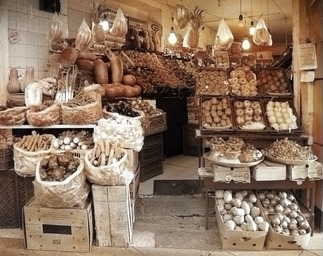}\\
            \includegraphics[width=\textwidth]{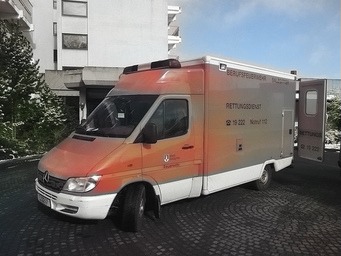}
        \end{center}
    \end{minipage}
    \hfill
    \begin{minipage}[b]{0.1802\linewidth}
        \vspace{0pt}
        \begin{center}
            \includegraphics[width=\textwidth]{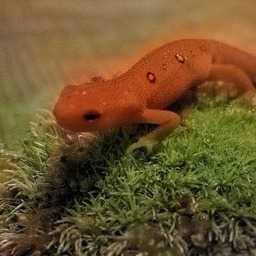}\\
            \includegraphics[width=\textwidth]{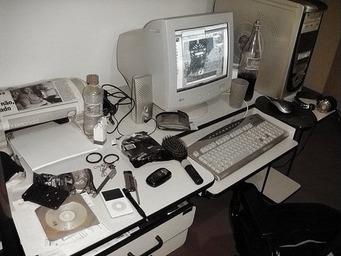}
        \end{center}
    \end{minipage}
    \hfill
    \begin{minipage}[b]{0.1895\linewidth}
        \vspace{0pt}
        \begin{center}
            \includegraphics[width=\textwidth]{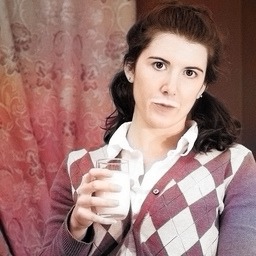}\\
            \includegraphics[width=\textwidth]{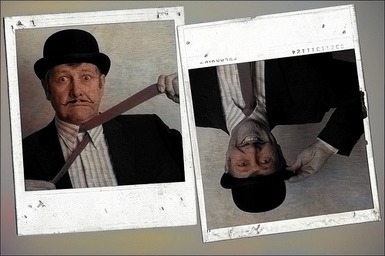}
        \end{center}
    \end{minipage}
    \hfill
    \begin{minipage}[b]{0.1860\linewidth}
        \vspace{0pt}
        \begin{center}
            \includegraphics[width=\textwidth]{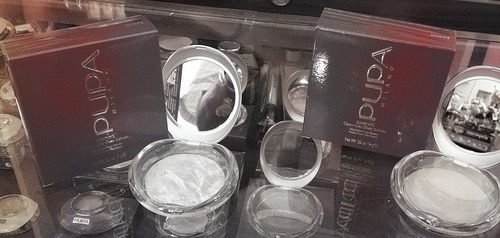}\\
            \includegraphics[width=\textwidth]{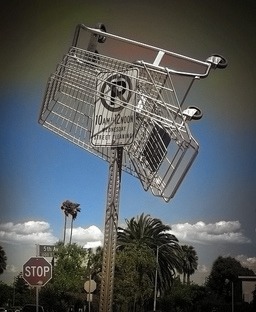}
        \end{center}
    \end{minipage}
    \vspace{-0.25cm}
    \end{center}
    \caption[Failure modes]{
        \textbf{Failure modes.}
        \emph{Top row, left-to-right:}
            texture confusion,
            too homogeneous,
            color bleeding,
            unnatural color shifts ($\times 2$).
        \emph{Bottom row:}
            inconsistent background,
            inconsistent chromaticity,
            not enough color,
            object not recognized (upside down face partly gray),
            context confusion (sky).
    }
    \label{fig:failure-modes}
\end{figure}

Though we achieve significant improvements over prior state-of-the-art, our
results are not perfect.  Figure~\ref{fig:failure-modes} shows examples of
significant failures.  Minor imperfections are also present in some of the
results in Figures~\ref{fig:ctest10k-examples}
and~\ref{fig:ctest10k-more-examples}.  We believe a common failure mode
correlates with gaps in semantic interpretation: incorrectly identified or
unfamiliar objects and incorrect segmentation.  In addition, there are
``mistakes'' due to natural uncertainty of color -- \eg~the graduation robe
at the bottom right of Figure~\ref{fig:ctest10k-examples} is red, but could
as well be purple.

Since our method produces histograms, we can provide interactive means of
biasing colorizations according to user preferences.  Rather than output a
single color per pixel, we can sample color for image regions and evaluate
color uncertainty.  Specifically, solving our energy minimization formulation
(Equation~\ref{eq:energy-min}) with global biases~$\mathbf{b}$ that are not
optimized based on a reference image, but simply ``rotated'' through color
space, induces changed color preferences throughout the image.  The uncertainty
in the predicted histogram modulates this effect.

\begin{figure}[!t]
    \centering
    \begin{tabular}{lllll}
        \includegraphics[width=.175\textwidth]{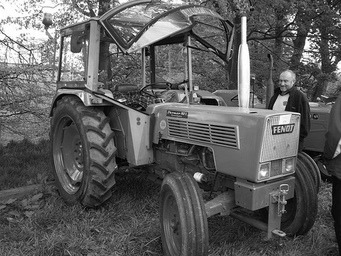}&
        \includegraphics[width=.175\textwidth]{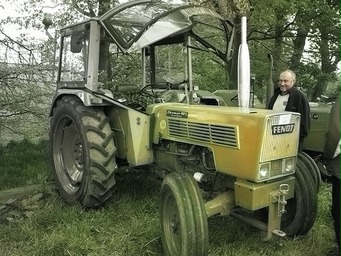}&
        \includegraphics[width=.175\textwidth]{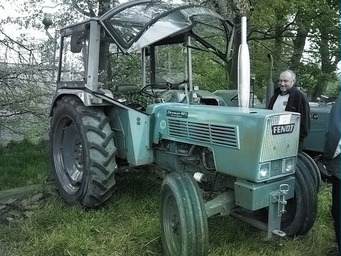}&
        \includegraphics[width=.175\textwidth]{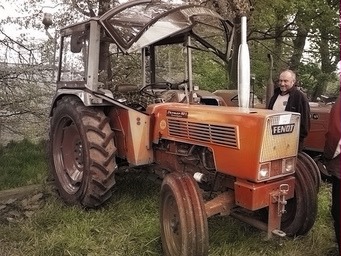}&
        \includegraphics[width=.175\textwidth]{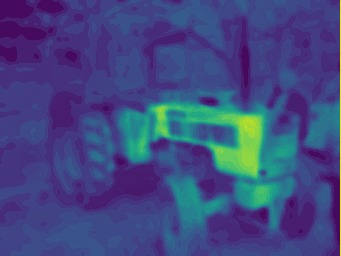}
    \end{tabular}
    \caption[Sampling multiple colorizations]{
        \textbf{Sampling multiple colorizations.}
        \emph{Left:} Image and 3 samples;
        \emph{Right:} Uncertainty map.
    }
    \label{fig:warhol}
\end{figure}

Figure~\ref{fig:warhol} shows multiple sampled colorizations, together with a
visualization of uncertainty.  Here, uncertainty is the entropy of the
predicted hue multiplied by the chroma.  Our distributional output and energy
minimization framework open the path for future investigation of
human-in-the-loop colorization tools.

\section{Conclusion}
\label{sec:final}

We present a system that demonstrates state-of-the-art ability to automatically
colorize grayscale images.  Two novel contributions enable this progress: a
deep neural architecture that is trained end-to-end to incorporate semantically
meaningful features of varying complexity into colorization, and a color
histogram prediction framework that handles uncertainty and ambiguities
inherent in colorization while preventing jarring artifacts.  Our fully
automatic colorizer produces strong results, improving upon previously leading
methods by large margins on all datasets tested; we also propose a new
large-scale benchmark for automatic image colorization, and establish a strong
baseline with our method to facilitate future comparisons.  Our colorization
results are visually appealing even on complex scenes, and allow for effective
post-processing with creative control via color histogram transfer and
intelligent, uncertainty-driven color sampling.

In \cref{chp:selfsup}, we reveal that colorization has utility beyond its
immediate application as a promising avenue for self-supervised visual
learning.

\setcounter{chapter}{2}
\chapter{Self-Supervised Representation Learning}
\label{chp:selfsup}

Returning to representation learning, we demonstrate in this chapter how
colorization can be used for self-supervised representation learning and how it
relates and compares to supervised pretraining.

\section{Introduction}

The success of deep feed-forward networks is rooted in their ability to scale
up with more training data. The availability of more data can
generally afford an increase in model complexity. However, this
need for expensive, tedious and error-prone human annotation is
severely limiting, reducing our ability to build models for new
domains, and for domains in which annotations are particularly
expensive (e.g., image segmentation). At the same time, we have access
to enormous amounts of unlabeled visual data, which is essentially
free. This work is an attempt to improve means of leveraging
this abundance. We manage to bring it one step closer to the results
of using labeled data, but the eventual long term goal of self-supervision
may be to supplant supervised pretraining completely.

Alternatives to supervised training that do not need labeled data have seen
limited success, as demonstrated in \cref{chp:unsupervised}. Unsupervised
learning methods, such as compressed embeddings
trained by minimizing reconstruction error, have seen more success in image
synthesis~\cite{kingma2014auto}\cmt{ and compression}, than for representation
learning. Semi-supervised learning, jointly training a supervised and an
unsupervised loss, offers a middle
ground~\cite{grandvalet2004semi,sajjadi2016semi}. However, recent works tend to
prefer a sequential combination instead (unsupervised pretraining, supervised
fine-tuning)~\cite{doersch2015unsupervised,donahue2016adversarial}, possibly
because it prevents the unsupervised loss from being disruptive in the
late stages of training. Perhaps more importantly, it reduces the engineering and computational burden
if a pretrained network is widely disseminated. A related endeavor
to unsupervised learning is developing models that work with weaker forms of
supervision~\cite{bearman2016point,xu2015weak}. This reduces the human burden
only somewhat and pays a price in model performance.

\begin{figure}[t]
    \centering
    \begin{minipage}{0.65\linewidth}
        \vspace{0pt}
        \includegraphics[width=0.99\linewidth]{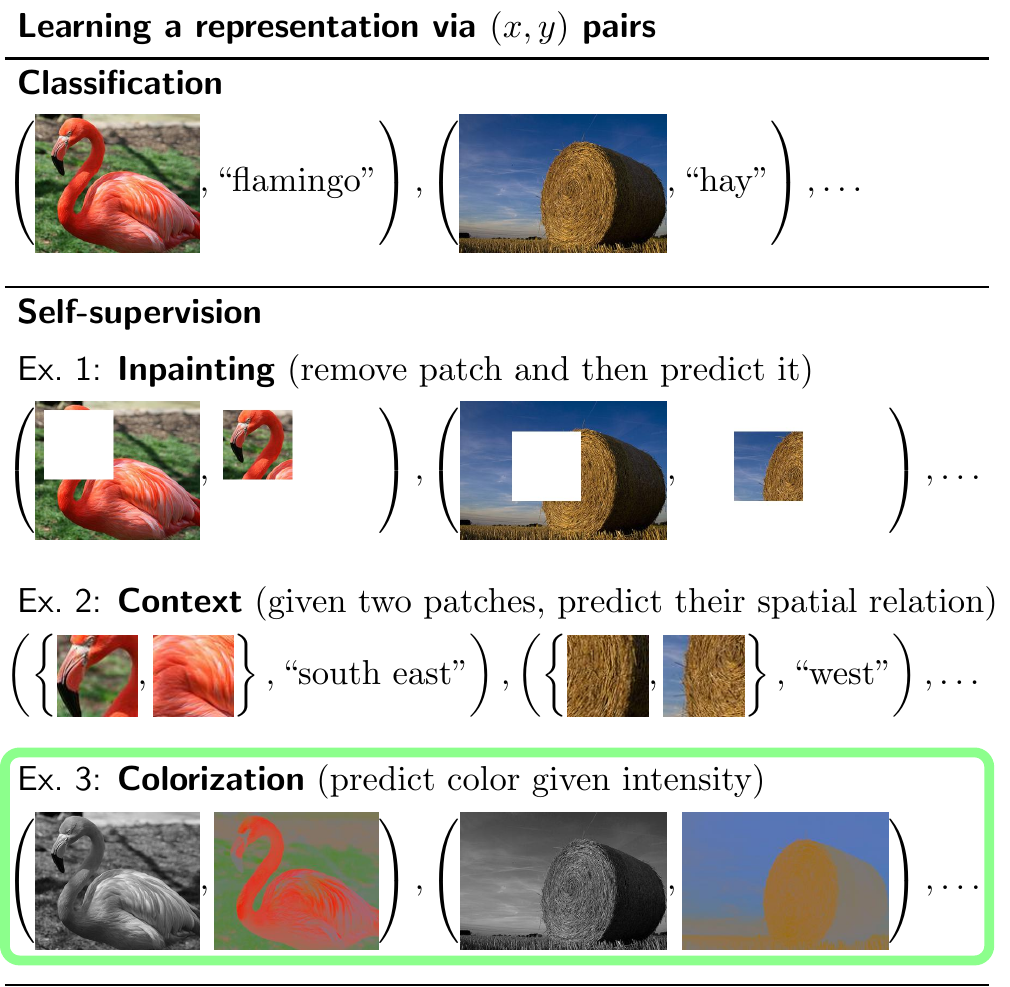}

    \end{minipage}
    \begin{minipage}{0.34\linewidth}
        \vspace{0pt}
        \caption[Self-supervision overview]{
            \textbf{Self-supervision.} Using a representation that was originally trained for classification on $(x, y)$
        pairs to initialize a network has become standard practice
        in computer vision. Self-supervision is a family of alternative
        pretraining methods that do not require any labeled data, since labels are
        ``manufactured" through unlabeled data. Examples include inpainting~\cite{contextencoders}, context prediction~\cite{contextpred}, and, the main focus of this chapter, automatic colorization~\cite{larsson2016learning,zhang2016colorful,larsson2017colorproxy}.
    }
    \end{minipage}
    \label{fig:overview}
\end{figure}

Recently, {\em self-supervision} has emerged as a new flavor of unsupervised
learning~\cite{doersch2015unsupervised,wang2015unsupervised,pathak2016context}. The key
observation is that perhaps part of the benefit of labeled data is that it
leads to using a discriminative loss.  This type of loss may be better
suited for representation learning than, for instance, a reconstruction or
likelihood-based loss. Self-supervision is a way to use a discriminative loss
on unlabeled data by partitioning each input sample in two, predicting the
parts' association. We focus on self-supervised colorization, where each image
is split into its intensity and its color, as described in
\cref{chp:colorization}.

Our main contributions to self-supervision are:

\begin{itemize}
    \item The idea of using colorization for self-supervision with
        state-of-the-art results on VOC 2007 Classification and VOC 2012
        Segmentation, among methods that do not use ImageNet labels.
    \item The first in-depth analysis of self-supervision via colorization. We
        study the impact of loss, network architecture and training details,
        showing that there are many important aspects that influence results.
    \item An empirical study of various formulations of ImageNet pretraining
        and how they compare to self-supervision.
\end{itemize}

\section{Related Work} \label{sec:selfsup-related}

In our work on replacing classification-based pretraining for downstream supervised tasks, the first thing to consider is clever network initializations.
Networks that are initialized to promote uniform scale of activations across
layers, converge more easily and
faster~\cite{glorot2010understanding,he2015msra}. The uniform scale however is
only statistically predicted given broad data assumptions, so this idea can be
taken one step further by looking at the activations of actual data and
normalizing~\cite{mishkin2015all}. Using some training data to initialize
weights blurs the line between initialization and unsupervised
pretraining. For instance, using layer-wise $k$-means
clustering~\cite{coates2010analysis,krahenbuhl2016datadriven} should be
considered unsupervised pretraining (see \cref{chp:unsupervised}), even though
it may be a particularly fast one.

Unsupervised pretraining can be used to facilitate optimization or to expose
the network to orders of magnitude larger unlabeled data. The former was once a
popular motivation, but fell out of favor as it was made unnecessary by
improved training techniques~(\eg, introduction of non-saturating
activations~\cite{nair2010rectified}, better
initialization~\cite{glorot2010understanding} and training
algorithms~\cite{qian1999momentum,kingma2015adam}). The second motivation of
leveraging more data, which can also be realized as semi-supervised training,
is an open problem with current best methods rarely used in competitive vision
systems.
Recent methods of self-supervised feature learning have seen several
incarnations, broadly divided into methods that exploit temporal
or spatial structure in natural visual data:

\textbf{Temporal.}
There have been a wide variety of methods that use the correlation between
adjacent video frames as a learning signal. One way
is to try to predict future frames, which is an analogous task to language
modeling and often uses similar techniques based on RNNs and
LSTMs~\cite{srivastava2015video,ranzato2014video}. It is also possible to train
an embedding where temporally close frames are considered similar (using either
pairs~\cite{mobahi2009deep,isola2015learning,jayaraman2015slow} or
triplets~\cite{wang2015unsupervised}). Another method that uses a triplet loss
presents three frames and tries to predict if they are correctly
ordered~\cite{misra2016unsupervised}. Pathak~\etal~\cite{pathak2016move} learn
general-purpose representation by predicting saliency based on optical flow.
Owens~\etal~\cite{owens2016ambient},
somewhat breaking from the temporal category, operate on a single video frame
to predict a statistical summary of the audio from the entire clip. The first
video-based self-supervision methods were based on Independent Component
Analysis (ICA)~\cite{van1998independent,hurri2003simple}. Recent
follow-up work generalizes this to a nonlinear
setting~\cite{hyvarinen2016unsupervised}.

\textbf{Spatial.} Methods that operate on single-frame input typically use the
spatial dimensions to divide samples for self-supervision. Given a pair of
patches from an image, Doersch~\etal~\cite{doersch2015unsupervised} train
representations by predicting which of eight possible spatial compositions the
two patches have. Noroozi \& Favaro~\cite{noroozi2016jigsaw} take this further and learns a representation by solving a 3-by-3 jigsaw
puzzle. The task of inpainting (remove some pixels, then predict
them) is utilized for representation learning by
Pathak~\etal~\cite{pathak2016context}.  There has also been work on adding encoders
to Generative Adversarial Networks that can be used to learn representations~\cite{donahue2016adversarial,dumoulin2016adversarially}. This is not what we typically
regard as self-supervision, but it does similarly pose a supervised learning
task (\emph{real} vs.\ \emph{synthetic}) on unlabeled data to drive
representation learning.


\textbf{Colorization.} Lastly there is
colorization~\cite{larsson2016learning,larsson2017colorproxy,zhang2016colorful,zhang2017split}.
Broadly speaking, the two previous categories split input samples along a
spatio-temporal line, either predicting one given the other or predicting the
line itself. Automatic colorization departs from this as it asks to predict
color over the same pixel as its center of input, without discarding any
spatial information. We speculate that this may make it more suitable to tasks
of similar nature, such as semantic segmentation; we demonstrate strong results
on this benchmark.


\begin{figure}[t]
    \qquad
    \begin{minipage}{0.34\linewidth}
        \vspace{0pt}
        \caption[Feature re-use/re-purpose]{
            \textbf{Feature re-use/re-purpose.} The left column visualizes top
            activations from the colorization network (same as in \cref{fig:viz}).
            The right column visualizes the corresponding feature after the network
            has been fine-tuned for semantic segmentation. Features are either
            re-used as is (top), specialized (middle), or scrapped and replaced (bottom).
            See \cref{fig:corr-after-finetune} for a quantitative study.
        }
    \end{minipage}
    \;\;\;\;\;
    \begin{minipage}{0.6\linewidth}
        \vspace{0pt}
        \includegraphics{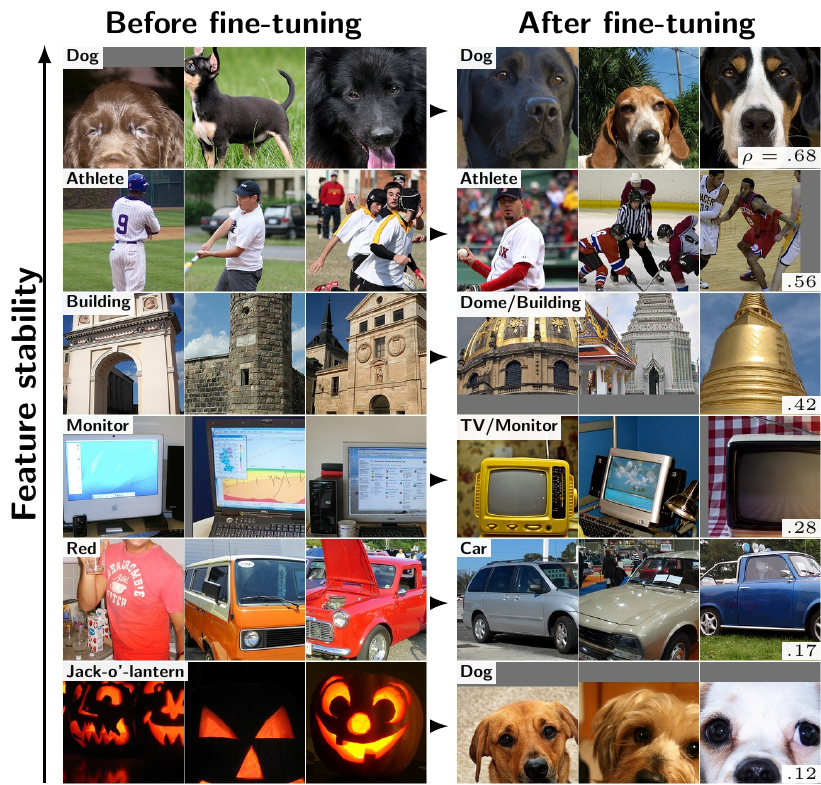}
    \end{minipage}
    \label{fig:viz-after-finetune}
\end{figure}

\begin{figure}
    \centering
    \includegraphics[width=0.85\linewidth]{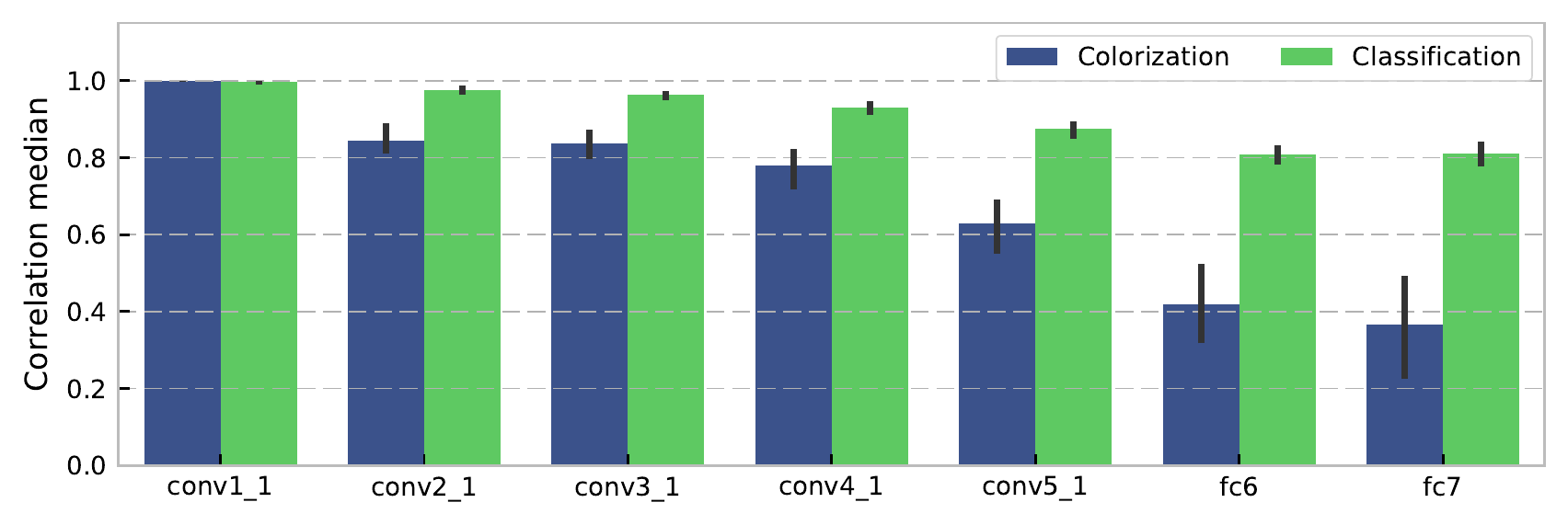}
    \caption[Feature shift]{
        \textbf{Feature shift.} The correlation between feature activations for
        layers of VGG-16 before and after fine-tuning for semantic segmentation. The
        bar heights indicate median correlation and error bars indicate
        interquartile range. See \cref{fig:viz-after-finetune} for qualitative
        examples.
    }
    \label{fig:corr-after-finetune}
\end{figure}

\section{Method}

We discuss training separately in two parts, pretraining (when colorization is
the target task) and fine-tuning (when colorization is a proxy task).

\subsection{Colorization as the Target Task}

In this section we revisit training an automatic colorizer from
\cref{chp:colorization}. However, we re-evaluate some of the design decisions
that were made with the goal of producing aesthetic color images and instead
consider their impact on learning representations.


\textbf{Loss.} We again consider both a regression loss for
Lab color
values~\cite{larsson2016learning,zhang2016colorful,iizuka2016color}, as well as
a KL divergence loss for hue/chroma histograms~\cite{larsson2016learning}. We evaluate
their ability to learn representations, disregarding their ability to do
colorization. In our comparison, we make sure that the losses are scaled
similarly, so that their effective learning rates are as close as possible
(even better would be to use NormSGD described in \cref{sec:norm-sgd}).

\textbf{Hypercolumn.} In \cref{chp:colorization}, hypercolumns were used with
sparse samples to train our colorization network.
Note that hypercolumns can be used for colorization pretraining, as well as for
segmentation as a downstream task. We know it is beneficial for downstream
tasks~\cite{mostajabi2015feedforward} and evaluate its use for pretraining.  Since we have reasons to believe
that hypercolumn training may disrupt residual training, we do not train our
ResNet colorizer from scratch with hypercolumns.

\textbf{Dataset.} We train on 3.7M \emph{unlabeled} images by combining 1.3M from
ImageNet~\cite{imagenet} and 2.4M from Places205~\cite{zhou2014learning}. The
dataset contains some grayscale images and we do not make an effort to sort
them out, since there is no way to tell a legitimately achromatic image from a
desaturated one.

\textbf{Training.} All training is done with standard Stochastic Gradient
Descent with momentum set to 0.9. The colorization network is initialized with
Xavier initialization~\cite{glorot2010understanding} and trained with batch
normalization without re-biasing or re-scaling
parameters~\cite{ioffe2015batch}. Each time an image is processed, it is
randomly mirrored and the image is randomly scaled such that the shortest side
is between 352 and 600. Finally, a 352-by-352 patch is extracted and
desaturated and then fed through the network. In our comparative studies, we
train using a colorization loss for 3 epochs (spending 2 epochs on the initial
learning rate). In our longer running experiments, we train for about 10
epochs. For our best ResNet model, we train significantly longer (35 epochs),
although on smaller inputs (224-by-224); we found large input sizes to be
more important during downstream training.
Code and trained models are made publicly
available.\footnote{\url{https://github.com/gustavla/self-supervision}}

\subsection{Colorization as a Proxy Task}

The downstream task is trained by initializing weights from the
colorization-from-scratch network. Some key training considerations follow:

\textbf{Early stopping.} Training on a small sample size is prone to
overfitting. We find that the most effective method of preventing this is
careful cross validation of the learning rate schedule. Models that initialize
differently (random, colorization, classification), need very different early
stopping schedules. Finding a method that works well in all these settings was
key to a fair and informative study. We split
the training data 90/10 and only train on the 90\%; the rest is used to monitor
overfitting. Each time the 10\% validation score (not surrogate loss) stops
improving, the learning rate is dropped. After this is done twice, the training
is concluded. For our most competitive experiments (\cref{tab:comparison}), we
then re-train using 100\% of the data with the cross-validated learning rate
schedule fixed.

\textbf{Receptive field.} Previous work on semantic segmentation has shown the
importance of large receptive
fields~\cite{mostajabi2015feedforward,yu2015multiscale}. One way to
accomplish this is by using dilated
convolutions~\cite{yu2015multiscale,wu2016resnet}. However, this redefines the
interpretation of filters and thus requires re-training.  Instead, we add two
additional blocks (2-by-2 max pooling of stride 2, 3-by-3 convolution with 1024 features)
at the top of the network, each expanding the receptive field with 160 pixels
per block (320 in total). In tables, we denote this change as ``+FoV.'' We
train on large input images (448-by-448) in order
to appreciate the enlarged receptive field. Using sparsely sampled targets
(\cref{sec:sparse-hypercolumns}) is particularly important for such large input size.

\textbf{Hypercolumn.} Note that using a hypercolumn when the downstream task is
semantic segmentation is a separate design choice that does not need to be
coupled with the use of hypercolumns during colorization pretraining. In either
case, the post-hypercolumn parameter weights are never re-used. For ResNet, we
use a subset of the full hypercolumn.\footnote{ResNet-152 hypercolumn:
\texttt{conv1}, \texttt{res2\{a,b,c\}}, \texttt{res3b\{1,4,7\}},
\texttt{res4b\{5,10,15,20,25,30,35\}}, \texttt{res5c}}

\textbf{Batch normalization.} The models trained from scratch use
parameter-free batch normalization. However, for downstream training, we absorb
the mean and variance into the weights and biases and train without batch
normalization (with the exception of ResNet, where in our experience it is still necessary).
For networks that were not trained with batch normalization and are not
well-balanced in scale across layers (\eg, ImageNet-pretrained VGG-16), we
re-balance the network so that each layer's activation has unit
variance (see \cref{sec:rebalancing}).

\begin{table}
\begin{center}
    \begin{tabular*}{\linewidth}{llr|rr}
        \toprule
        Initialization                                                      & Architecture        &                                  & Classification                  & Segmentation                            \\
                                                                            &                     &                                  & {\footnotesize \%mAP}   & {\footnotesize \%mIU}           \\ \midrule
        ImageNet                                                            & VGG-16              & {\scriptsize (+FoV)} \scriptsize & 86.9\cmt{a230}          & 69.5                            \\ \toprule
        Random (ours)                                                       & AlexNet             &                                  & 46.2\cmt{a231}          & 23.5\cmt{a228}                  \\
        Shortcut Autoencoder (ours)                                         & AlexNet             &                                  & 53.3\cmt{mu104}         & 28.7\cmt{a255}                  \\ \midrule
        Random~\cite{pathak2016context}                                     & AlexNet             &                                  & 53.3                    & 19.8\cmt{H:a116}                \\
        $k$-means~\cite{krahenbuhl2016datadriven,donahue2016adversarial}    & AlexNet             &                                  & 56.6                    & 32.6                            \\
        $k$-means~\cite{krahenbuhl2016datadriven}                           & VGG-16              &                                  & 56.5                    & \dash                           \\
        $k$-means~\cite{krahenbuhl2016datadriven}                           & GoogLeNet           &                                  & 55.0                    & \dash                           \\ \midrule
        Pathak~\etal~\cite{pathak2016context}                               & AlexNet             &                                  & 56.5                    & 29.7                            \\
        Wang \amp\ Gupta~\cite{wang2015unsupervised}                        & AlexNet             &                                  & 58.7                    & \dash                           \\
        Donahue~\etal~\cite{donahue2016adversarial}                         & AlexNet             &                                  & 60.1                    & 35.2                            \\
        Doersch~\etal~\cite{doersch2015unsupervised,donahue2016adversarial} & AlexNet             &                                  & 65.3                    & \dash                           \\
        Zhang~\etal~(col)~\cite{zhang2016colorful}                          & AlexNet             &                                  & 65.6                    & 35.6                            \\
        Zhang~\etal~(s-b)~\cite{zhang2017split}\cmt{$^\dagger$}             & AlexNet             &                                  & 67.1                    & 36.0                            \\
        Noroozi \amp\ Favaro~\cite{noroozi2016jigsaw}                       & Modified AlexNet    &                                  & 68.6                    & \dash                           \\ \midrule
        Our method                                                          & AlexNet             &                                  & 65.9\cmt{c107d/ds-cls}  & 38.4\cmt{a222}                  \\
                                                                            & VGG-16              & {\scriptsize (+FoV)}             & \textbf{77.2}\cmt{a224} & 56.0\cmt{a138}                  \\
                                                                            & ResNet-152          & {\scriptsize (+FoV)}             & \textbf{77.3}\cmt{a246} & \textbf{60.0}\cmt{a244}         \\ \midrule
        Our ensemble                                                        & $3\times$ResNet-152 & {\scriptsize (+FoV)}             & \textbf{79.8}           & \textbf{61.6}\cmt{}             \\
        \bottomrule
    \end{tabular*}
    \caption[Comparison on VOC of self-supervised pretraining methods]{
        \textbf{VOC Comparison.} Comparison with other initialization and
        self-supervision methods on VOC 2007 Classification (test) and VOC 2012
        Segmentation (val). First, we see how poorly our autoencoder really
        performed, with greedy layer-wise pretraining outperforming it and sometimes random initialization.
        However, it is worth mentioning that we perturb detection scores before
        calculating mAP, since otherwise even a trivial solution can
        incorrectly get around 53\%; we cannot confirm that this ever happened
        to numbers that are not ours, however we do caution the reader that
        numbers in this range may not be reliable.
        Our baseline AlexNet results (38.4\%) are
        also the most competitive among AlexNet models. The use of
        a hypercolumn instead of FCN is partly responsible: running
        Zhang~\etal's colorization model with a hypercolumn yields 36.4\%\cmt{218}, only a
        slight improvement over 35.6\%.
        Switching to ResNet, adding a larger FoV, and training even longer
        yields a significantly higher result at 60.0\%\cmt{a221} mIU. This
        number can be improved further with an ensemble of three
        colorization-pretrained networks. Note, the ``+FoV'' only affects the
        segmentation results.  The modified AlexNet used by Noroozi \amp{}
        Favaro has the same number of parameters as AlexNet, with a spatial
        reduction of 2 moved from conv1 to pool5, increasing the size of the
        intermediate activations.  \cmt{($^\dagger$ Concurrent work)}
    }
    \label{tab:comparison}
\end{center}
\end{table}

\textbf{Padding.} For our ImageNet pretraining experiments, we observe that
going from a classification network to a fully convolutional network can
introduce edge effects due to each layer's zero padding. A problem not
exhibited by the original VGG-16, leading us to suspect that it may be due to
the introduction of batch normalization. For the newly trained networks,
activations increase close to the edge, even though the receptive fields
increasingly hang over the edge of the image, reducing the amount of semantic
information. Correcting for this\footnote{We pad with the bias from the
previous layer, instead of with zeros. This is an estimate of the expectation value,
since we use a parameter-free batch normalization with zero mean, leaving
only the bias.} makes activations well-behaved, which was important in order to
appropriately visualize top activations. However, it does not offer a
measurable improvement on downstream tasks, which means the network can
correct for this during the fine-tuning stage.

\textbf{Color.} Since the domain of a colorization network is grayscale, our
downstream experiments operate on grayscale input unless otherwise stated. When
colorization is re-introduced, we convert the grayscale filters in
\texttt{conv1\_1} to RGB (replicate to all three channels, divide by three) and
let them fine-tune on the downstream task.

\section{Results} \label{sec:results}

We first present results on two established PASCAL VOC benchmarks, followed in
\cref{sec:experiments} by an investigation into different design choices and
pretraining paradigms.

\subsection{PASCAL}
\label{sec:pascal}

\textbf{VOC 2012 Semantic Segmentation.} We train on the standard extended
segmentation data (10,582 samples) and test on the validation set (1,449
samples). We sample random crops at the original scale. Our AlexNet results show
that colorization is a front-runner among self-supervised methods at 38.4\% mIU. Using our
ResNet-152 model with extended field-of-view we achieve 60.0\% mIU (see
\cref{tab:comparison}), the highest reported results on this benchmark that do
not use supervised pretraining. It is noticeable that this value is
considerably higher than the AlexNet-based FCN~\cite{long2015fully} (48.0\%)
and even slightly higher than the VGG-16-based FCN (59.4\%\footnote{Both of
these values refer to VOC 2011 and evaluated on only 736 samples, which means
the comparison is imprecise.}), both methods train on ImageNet with class
annotations.

\textbf{VOC 2007 Classification.} We train on \textit{trainval} (5,011 samples) and
test on \textit{test} (4,952 samples). We use the same training procedure and loss with
10-crop testing as in~\cite{donahue2016adversarial}. Our AlexNet results at
65.9\% mAP match state-of-the-art compared to other models. Our ResNet results
improve this to 77.3\% mAP (see \cref{tab:comparison}), setting a new state-of-the-art when
no ImageNet labels are used.

These values can be improved further to 61.6\% and 79.8\%, respectively, by
forming an ensemble with our best ResNet model (trained for 35 epochs) and two separately trained ResNet models (trained for 10 epochs each).

\section{Experiments} \label{sec:experiments}

\begin{table}
    \begin{minipage}[t]{0.58\linewidth}
        \vspace{0pt}
        \begin{tabular*}{\linewidth}{lr}
        \toprule
        Pretraining Loss                              & Seg. (\%mIU) \\ \midrule
        Regression\cmt{c95}                           & 48.0\cmt{a159} \\
        Histograms (no hypercolumn)\cmt{c92b} \qquad\qquad        & 52.7\cmt{a161} \\
        Histograms\cmt{c90e batch size 7, rest are 8} & 52.9\cmt{a160} \\
        \bottomrule
    \end{tabular*}
        \caption[Self-supervision loss]{
        \textbf{Self-supervision loss.} (VGG-16) The choice of loss has a
        significant impact on downstream performance. However, pretraining with
        a hypercolumn does not seem to benefit learning. We evaluate this on
        VOC 2012 Segmentation (val) with a model that uses hypercolumns,
        regardless of whether or not it was used during pretraining.
    }\label{tab:loss}
    \end{minipage}
    \,\,
    \begin{minipage}[t]{0.39\linewidth}
        \vspace{0pt}
    \begin{tabular*}{\linewidth}{l|rr}
        \toprule
        Initialization & Grayscale & Color \\\midrule
        Classification & 66.5 \cmt{a135} & 69.5\cmt{a22 } \\
        Colorization   & 56.0 \cmt{a138} & 55.9\cmt{a173} \\
        \bottomrule
    \end{tabular*}
        \caption[Color vs.\ grayscale input]{
        \textbf{Color vs.\ grayscale input.} (VOC 2012 Segmentation, \%mIU) Even
        though our classification-based model does 3 points better using color,
        re-introducing color yields no benefit.
    }
    \label{tab:color}
    \end{minipage}
\end{table}

We present a wide range of experiments, highlighting important aspects of our
competitive results. For these studies, in addition to VOC 2012 Semantic
Segmentation, we also use two classification datasets that we constructed:

\textbf{ImNt-100k/ImNt-10k.} Similar to ImageNet classification with 1000
classes, except we have limited the training data to exactly 100 and 10
samples/class, respectively. In addition, all images are converted to
grayscale. We test on ImageNet \textit{val} with single center crops of size 224-by-224,
making the results easy to compare with full ImageNet training. For our
pretraining experiments in \cref{tab:pretraining}, we also use these datasets to
see how well they are able to substitute the entire ImageNet dataset for
representation learning.

\subsection{Loss}
As seen in \cref{tab:loss}, regressing on color in the Lab space yields a
5-point lower result (48.0\%) than predicting histograms in hue/chroma (52.9\%). This 
demonstrates that the choice of loss is of crucial importance to
representation learning. This is a much larger difference than
\cref{chp:colorization} reports in colorization performance
between the two methods ($24.25$ and  $24.45$ dB PSNR / $0.318$ and $0.299$ RMSE).
Histogram predictions are meant to address the problem of color uncertainty.
However, the way they instantiate an image by using summary statistics from the
histogram predictions, means this problem to some extent is re-introduced. Since
we do not care about instantiating images, we do not suffer this penalty and
thus see a much larger improvement using a loss based on histogram predictions.
Our choice of predicting separate histograms in the hue/chroma space also
yields an interesting finding in \cref{fig:viz}, where we seem to have
non-semantic filters that respond to input with high chromaticity as well as
low chromaticity, clearly catering to the chroma prediction.

\begin{table}
\begin{center}
    \begin{tabular*}{\linewidth}{l@{\quad}l@{\quad}r|@{\quad}r@{\quad}r@{\quad}|@{\qquad}r@{\quad}r}
        \toprule
          Architecture
        & Initialization
        & Epochs 
        & Segmentation
        & +FoV
        & ImNt-100k
        & -10k \\

        &
        &
        & \multicolumn{2}{c|@{\qquad}}{\footnotesize \%mIU}
        & \multicolumn{2}{c}{\qquad\quad\footnotesize\%top-5}
        \\ \midrule
        AlexNet
        & Random
        & -
        & 23.5\cmt{a228}
        & 24.6\cmt{a229}
        & 39.1\cmt{a143c}
        & 6.7\cmt{a91 }
        \\
        AlexNet
        & Colorization\cmt{c98}
        & 3
        & 36.2\cmt{a219}
        & 40.8\cmt{a227}
        & 48.2\cmt{a144c}
        & 17.4\cmt{a145c}
        \\
        AlexNet
        & Colorization\cmt{c107d}
        & 10
        & 38.4\cmt{a222,a260}
        & 42.5\cmt{a261}
        & 49.7\cmt{a262}
        & 18.8\cmt{a263}
        \\ \midrule
        VGG-16
        & Random
        & -
        & 32.8\cmt{a157}
        & 35.1\cmt{a156 or a157}
        & 43.2\cmt{a90 ,old42.0}
        & 8.6\cmt{a88 ,old8.0}
        \\
        VGG-16
        & Colorization\cmt{c90e}
        & 3
        & 50.7\cmt{a178}
        & 52.9\cmt{a160}
        & 59.0\cmt{a164,old59.0}
        & 23.3\cmt{a165,old23.0}
        \\
        VGG-16
        & Colorization\cmt{c90c}
        & 10
        & 55.0\cmt{a264}
        & 56.0\cmt{a138}
        & 63.1\cmt{a265}
        & 28.9\cmt{a266}
        \\ \midrule
        ResNet-152
        & Random
        & -
        & \color{gray}{*9.9\cmt{a186}}
        & \color{gray}{*10.5\cmt{a166}}
        & 42.5\cmt{a139,old43.0}
        & 8.1\cmt{a167,old8.0}
        \\
        ResNet-152
        & Colorization\cmt{c93c}
        & 3
        & 52.3\cmt{a177}
        & 53.9\cmt{a168}
        & 63.1\cmt{a169,old62.0}
        & 29.6\cmt{a170,old29.0}
        \\
        ResNet-152
        & Colorization\cmt{c93b}
        & 10
        & 57.5\cmt{a267}
        & 59.1\cmt{a268}
        & 68.1\cmt{a269}
        & 32.9\cmt{a270}
        \\
        ResNet-152
        & Colorization\cmt{c93c}
        & 35
        & 58.7\cmt{a271}
        & 60.0\cmt{a244}
        & 68.8\cmt{a272}
        & 35.0\cmt{a273}
        \\ \bottomrule
    \end{tabular*}
    \caption[Network architecture validation]{
        \textbf{Architectures.} 
        Evaluation of the impact of network architecture and pretraining time.
        The networks are evaluated on VOC 2012 semantic segmentation (val) and
        two small-sample ImageNet classification tasks.  For our segmentation
        results, we also consider the effects of increasing the receptive field
        size (+FoV). Training residuals from scratch (marked with a *) is
        possibly compromised by the hypercolumn, causing the low values.
    }
    \label{tab:archs}
\end{center}
\end{table}

\begin{table}
\begin{center}
    \begin{minipage}{0.47\linewidth}
        \begin{tabular*}{\linewidth}{lrr|r}
            \toprule
            Pretraining         & $N$ & Epochs & Seg. \scriptsize \%mIU          \\ \midrule
            None                & \ndash       & \ndash & 35.1\cmt{a156 or a157}\\ \midrule
            C1000               & 1.3M         & 80     & 66.5\cmt{a135}        \\
            C1000  \cmt{bo53}   & 1.3M         & 20     & 62.0\cmt{a134}        \\
            C1000  \cmt{bo58}   & 100k         & 250    & 57.1\cmt{a125}        \\
            C1000  \cmt{bo59}   & 10k          & 250    & 44.4\cmt{a112}        \\ \midrule
            \multicolumn{2}{l}{E10    \cmt{bo63} \quad  {\scriptsize (1.17M)} 1.3M} & 20     & 61.8\cmt{a232b}       \\
            \multicolumn{2}{l}{E50    \cmt{bo64} \quad {\scriptsize (0.65M)} 1.3M} & 20     & 59.4\cmt{a235}        \\ \midrule
            H16    \cmt{bo54}   & 1.3M         & 20     & 60.0\cmt{a121}        \\
            H2     \cmt{bo55}   & 1.3M         & 20     & 46.1\cmt{a122}        \\ \midrule
            R50    \cmt{bo56}   & 1.3M         & 20     & 57.3\cmt{a123}        \\
                   \cmt{bo56b}  &              & 40     & 59.4\cmt{a172}        \\
            R16    \cmt{bo60b}  & 1.3M         & 20     & 42.6\cmt{a243}        \\ 
                   \cmt{bo60}   &              & 40     & 53.5\cmt{a220}        \\
            \bottomrule
        \end{tabular*}
    \end{minipage}
    \,
    \begin{minipage}{0.495\linewidth}
    \includegraphics{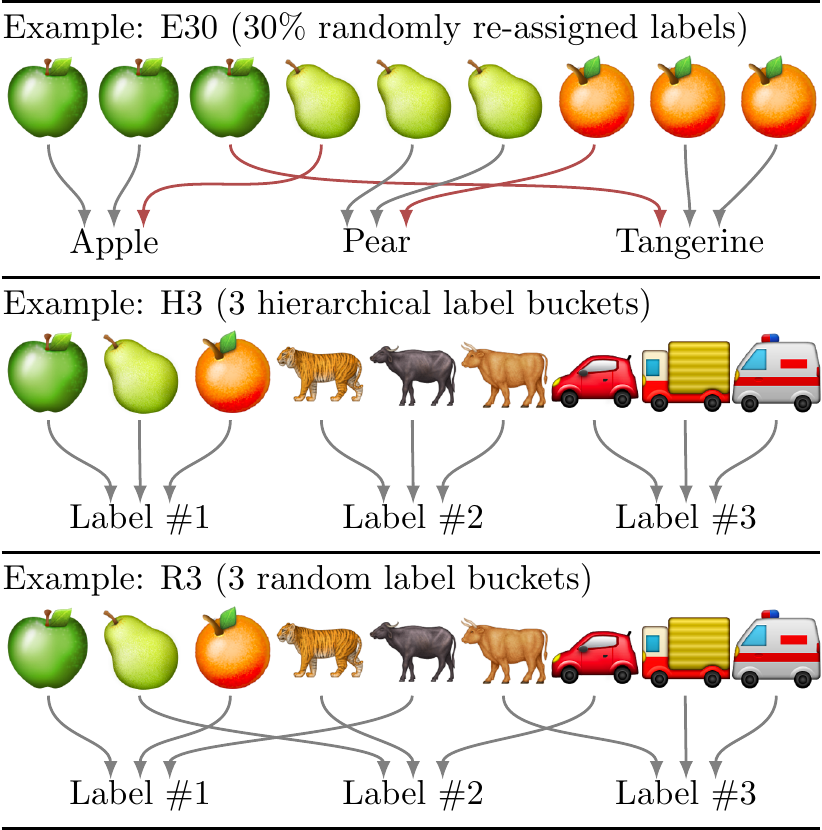}
    \end{minipage}
    \caption[ImageNet pretraining variations]{
        \textbf{ImageNet pretraining variations.} We evaluate how useful various
        modifications of ImageNet are for VOC 2012 Segmentation (val-gray). We
        create new datasets either by reducing sample size or by reducing the
        label space. The former is done simply by reducing sample size or by introducing 10\% (E10) or 50\% (E50) label noise.
        The latter is done using hierarchical label buckets
        (H16 and H2) or random label buckets (R50 and R16).
        The model trained for 80 epochs
        is the publicly available VGG-16 (trained for 76 epochs) that we
        fine-tuned for grayscale for 4 epochs.  The rest of the models were
        trained from scratch on grayscale images.
    }
    \label{tab:pretraining}
\end{center}
\end{table}

\subsection{Network Architecture}
The investigation into the impact of network architecture has been a neglected
aspect of recent self-supervision work, which has focused only on the dated AlexNet.  We
present the first detailed study into the untapped potential of using more
modern networks. These results are presented in \cref{tab:archs}.

It is not entirely obvious that an increase in model complexity will pay off,
since our focus is small-sample datasets and a smaller network may offer a
regularizing effect.  Take ImNt-100k, where AlexNet, VGG-16, and ResNet-152 all
perform similarly when trained from scratch (39.1\%, 43.2\%, 42.5\%). However,
the percentage point improvement when utilizing 10 epochs of colorization pretraining
follows a clear trend (+10.6, +19.9, +25.6).  This shows that self-supervision
allows us to benefit from higher model complexity even in small-sample regimes.
Compare this with $k$-means initialization~\cite{krahenbuhl2016datadriven},
which does not show any improvements when increasing model complexity
(\cref{tab:comparison}).

Training ResNet from scratch for semantic segmentation is an outlier value in
the table. This is the only experiment that trains a residual network from
scratch together with a hypercolumn; this could be a disruptive combination as
the low numbers suggest.

\subsection{ImageNet Pretraining} \label{sec:pretraining}

We relate self-supervised pretraining to ImageNet pretraining by revisiting and
reconsidering various aspects of this paradigm (see \cref{tab:pretraining}).
First of all, we investigate the importance of 1000 classes (C1000).  To do
this, we join ImageNet classes together based on their place in the WordNet
hierarchy, creating two new datasets with 16 classes (H16) and only two classes
(H2). We show that H16 performs only slightly short of C1000 on a downstream
task with 21 classes, while H2 is significantly worse. If we compare this to
our colorization pretraining, it is much better than H2 and only slightly worse
than H16.

Next, we study the impact of sample size, using the subsets ImNt-100k and
ImNt-10k described in \cref{sec:experiments}.  ImNt-100k does similarly well as
self-supervised colorization (57.1\% vs.\ 56.0\% for VGG-16), suggesting that our method
has roughly replaced 0.1M labeled samples with 3.7M unlabeled
samples. Reducing samples to 10 per class sees a bigger drop in downstream
results. This result is similar to H2, which is somewhat surprising: collapsing
the label space to a binary prediction is roughly as bad as using 1/100th of
the training data. Recalling the improvements going from regression to
histogram prediction for colorization, the richness of the label space seems
critical for representation learning.

We take the 1000 ImageNet classes and randomly place them in 50 (R50) or 16 (R16) buckets that we
dub our new labels. This means that we are training a highly complex
decision boundary that may dictate that a golden retriever and a minibus belong
to the same label, but a golden retriever and a border collie do not. We
consider this analogous to self-supervised colorization, since the supervisory
signal similarly considers a red car arbitrarily more similar to a red postbox
than to a blue car. Not surprisingly, our contrived dataset R50 results in a
5-point drop on our downstream task, and R16 even more so with a
20-point drop. However, we noticed that the training loss
was still actively decreasing after 20 epochs. Training instead for 40 epochs
showed an improvement by about 2 points for R50, while 11 points for R16.
In other words, complex classes can provide useful supervision for
representation learning, but training may take longer.
This is consistent with our impression of self-supervised colorization;
although it converges slowly, it keeps improving its feature generality with
more training.

Finally, we test the impact of label noise. When 10\% of the training images
are re-assigned a random label (E10), it has little impact on downstream
performance. Increasing the label noise to 50\% (E50) incurs a 2.6-point penalty, but it is
still able to learn a competitive representation.

\subsection{Training Time and Learning Rate}
\begin{figure}
\begin{center}
    \includegraphics[width=\linewidth,trim=0 0.0cm 0 0]{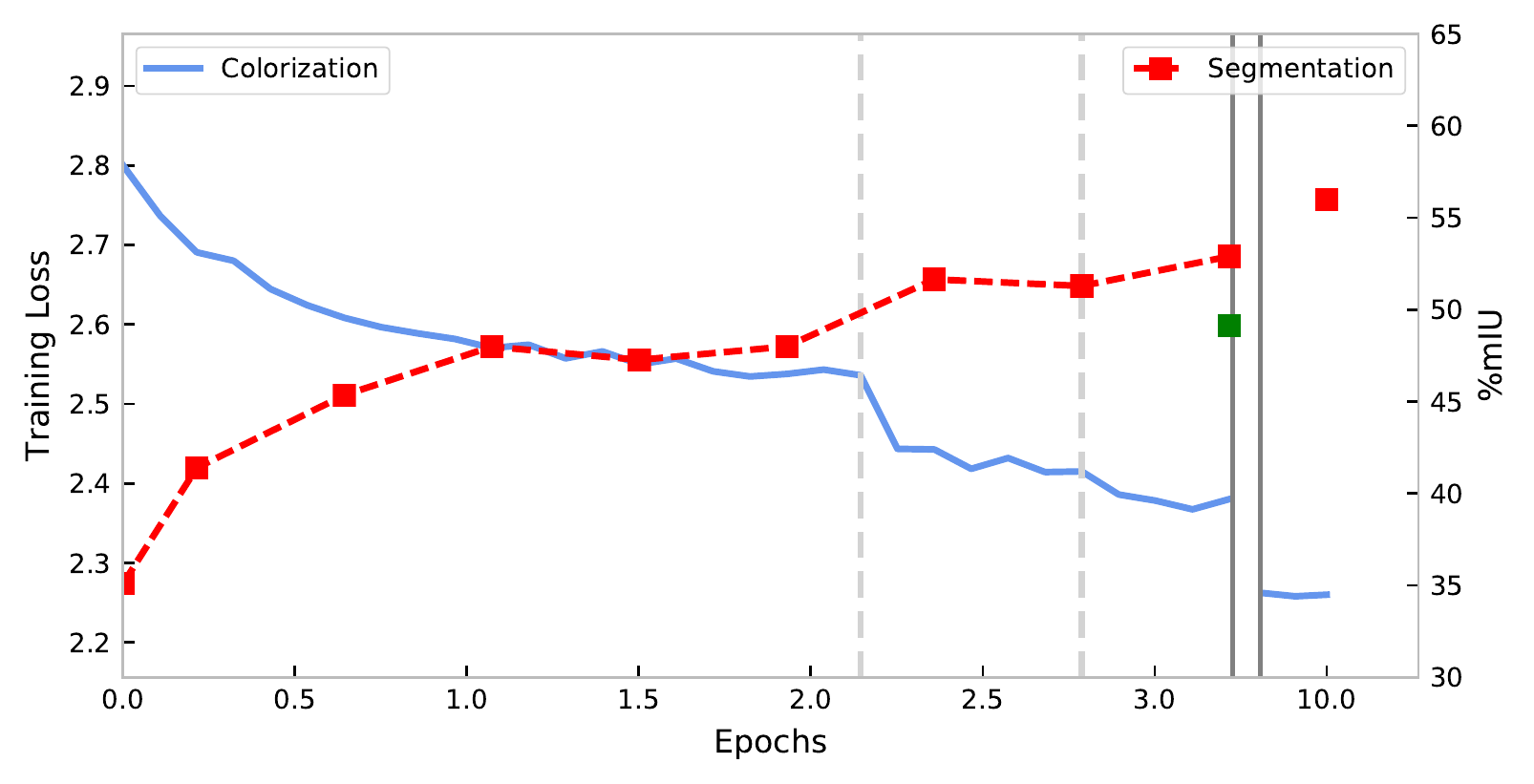}\cmt{a158}
\end{center}
\caption[Learning rate]{
\textbf{Learning rate.}
    The blue line shows colorization training loss and the vertical dashed
    lines are scheduled learning rate drops. The red squares are results on a
    downstream task (VOC 2012 Segmentation) initialized by the corresponding
    snapshot of the colorization network. Some key observations: We quickly get
    value for our money, with a 6-point improvement over random initialization
    with only 0.2 epochs of training. Furthermore, improvements on the
    downstream task do not quickly saturate, with results improving further
    when trained 10 epochs in total. Dropping the learning rate on the
    pretraining task helps the downstream task, with a similarly abrupt
    improvement as with the training loss around 2 epochs. Training the full 3
    epochs without ever dropping the learning rate results in 49.1\% (green 
    square) compared to 52.9\% mIU.
}
\label{fig:droplr}
\end{figure}

We show in \cref{fig:droplr} that it is crucial for good performance on
downstream tasks to reduce learning rate during pretraining. This result was
not obvious to us, since it is possible that the late stage of training with
low learning rate is too task-specific and will not benefit feature generality.

In addition, we show the importance of training time by demonstrating that
training for three times as long (10 epochs, 37M samples) improves results from
52.9\% to 56.0\% mIU on VOC 2012 Segmentation. Our ResNet-152 model (60.0\%
mIU) trained for 4 months on a single GPU.

\def\n{\tiny\textcolor{black}{$\square$}}
\def\y{\tiny\textcolor{black}{$\blacksquare$}}

\begin{table}
\centering
    \begin{tabular*}{\linewidth}{l@{\qquad\qquad\qquad\qquad}r|@{\qquad\qquad\,\,}r@{\quad\,\,}r@{\quad\,\,}r}
        \toprule
        \multicolumn{2}{l|@{\qquad\qquad\,\,}}{Fine-tuned layers (VGG-16)} & Random & Colorization & Classification \\ \midrule
        $\varnothing$   & \n\n\n\n\n\n\n & 3.6\cmt{a153}          & 36.5\cmt{a148} & 60.8\cmt{a147}\\
        {fc6, fc7}      & \n\n\n\n\n\y\y & \dash\cmt{a154}               & 42.6\cmt{a150} & 63.1\cmt{a149}\\
        {conv4\_1..fc7} & \n\n\n\y\y\y\y & \dash\cmt{a155}               & 53.6\cmt{a152} & 64.2\cmt{a151}\\
        {conv1\_1..fc7} & \y\y\y\y\y\y\y & 35.1\cmt{a156: or a157}& 56.0\cmt{a138} & 66.5\cmt{a135}\\
        \bottomrule
    \end{tabular*}
    \caption[End-to-end fine-tuning]{
        \textbf{End-to-end fine-tuning.} (VOC 2012 Segmentation, \%mIU) Classification-based pretraining
        needs less fine-tuning than our colorization-based method.
        This is consistent with our findings that our network experiences a
        higher level of feature shift (\cref{fig:corr-after-finetune}). We
        also include results for a randomly initialized network, which
        does not work at all without fine-tuning (3.6\%). This is to show that
        it is not simply by virtue of the hypercolumn that we are able to do
        reasonably well (36.5\%) without any fine-tuning of the base network.
    }
    \label{tab:finetune}
\end{table}

\subsection{Latent Representation}
Good results on secondary tasks only give evidence that our self-supervised network has
the potential to be shaped into a useful representation. We investigate if the
representation learned through colorization is immediately
useful or only holds a latent representation. If the latter, how is our
representation different from a good initialization scheme?

\begin{figure*}[t]
    \centering
    \includegraphics{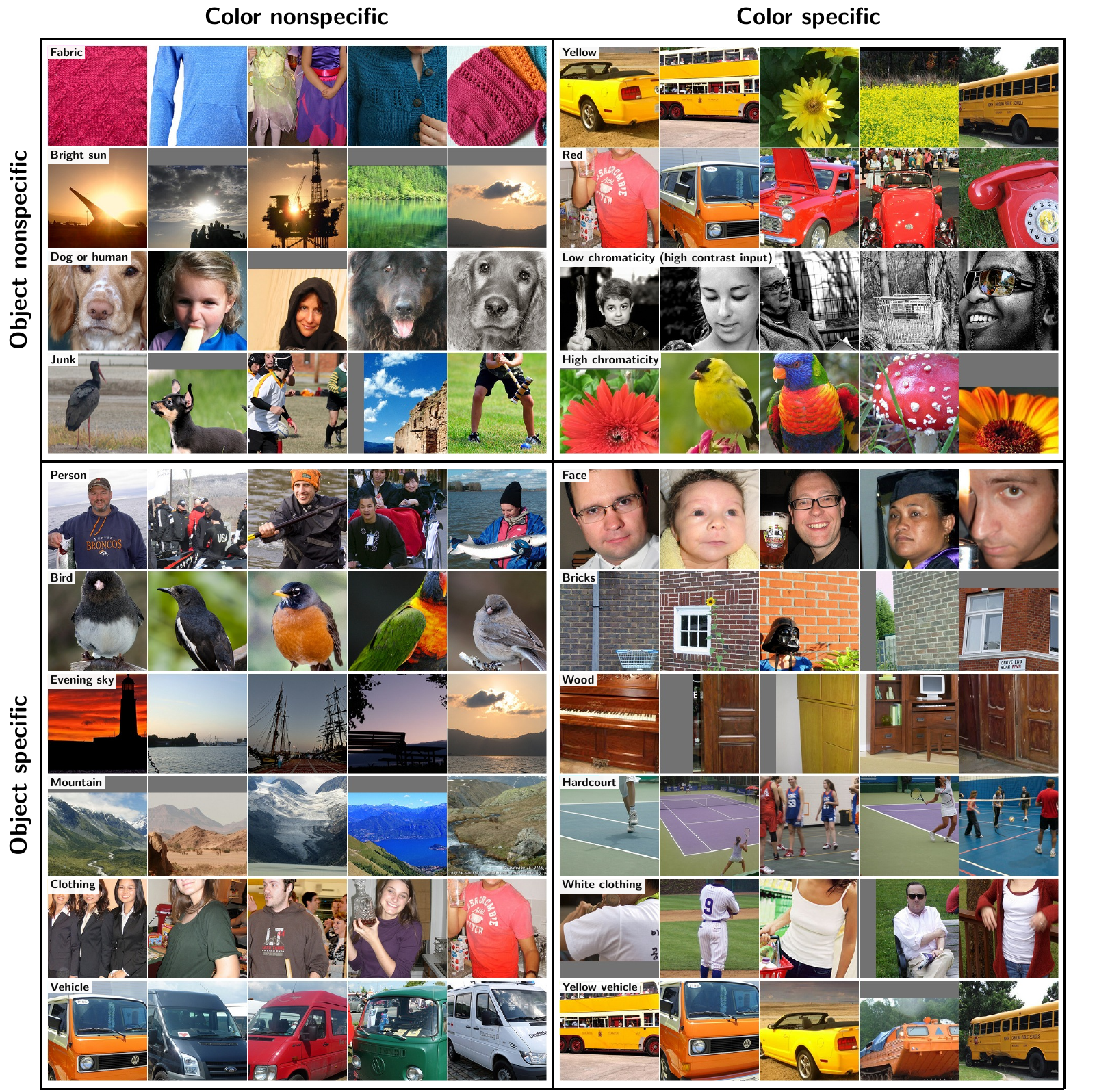}
    \caption[Feature visualization]{
        \textbf{Feature visualization.}
        Patches around activations from held-out images are visualized for a select number of \texttt{fc7} features (VGG-16). Even though the network takes only grayscale input, we visualize each patch in its original color for the benefit of the reader. As a result, if all the activations are consistent in color (right column), the feature is predictive of color. Similarly, if a feature is semantically coherent (bottom row), it means the feature is predictive of an object class. The names of each feature are manually set based on the top activations.
    }
    \label{fig:viz}
\end{figure*}

First, we visualize features to get a sense of how the colorization network has
organized the input into features. We posit that we will find features
predictive of color, since we know that the colorization network is able to
predict color with good accuracy. In \cref{fig:viz}, we visualize top
activations from the network's most high-level layer, and indeed we find
color-specific features. However, we also find semantic features that group
high-level objects with great intra-class variation (color, lighting, pose,
etc.).  This is notable, since no labeled data was used to train the network.
The notion of objects has emerged purely through their common color and visual
attributes (compare with~\cite{zhou2015deepscene}). Object-specific features
should have high task generality and be useful for downstream tasks. Features
that are specific to both object and color (bottom-right quadrant in
\cref{fig:viz}) can be divided into two categories: The first is when the
object generally has a unimodal color distribution (e.g.\ red bricks, brown
wood); the second is when the network has learned a color sub-category of an
object with multimodal color distribution (e.g.\ white clothing, yellow
vehicle). These should all have high task generality, since it is easy for a
task-specific layer to consolidate several color sub-categories into
color-invariant notions of objects.

So how much do the features change when fine-tuned? We visualize top
activations before and after in \cref{fig:viz-after-finetune} and show
in \cref{fig:corr-after-finetune} that the colorization features change
much more than supervised features. Some features are completely re-purposed,
many are only pivoted, and others remain more or less the same. These
results are consistent with the four quadrants in \cref{fig:viz}, that show
that some features are specific to colorization, while others seem to have
general purpose.

Next, we look at how much fine-tuning is required for the downstream task.
\Cref{tab:finetune} tells us that even though fine-tuning is more important
than for supervised pretraining (consistent with the correlation results
in \cref{fig:corr-after-finetune}), it is able to perform the task with the
colorization features alone similarly well as randomly initializing the network
and training it end-to-end from scratch.

Somewhat poor results without fine-tuning and a lower percentage of feature
re-use supports the notion that the colorization network in part holds {\em
latent} features. However, the visualized features and the strong results
overall suggest that we have learned something much more powerful than a good
initialization scheme.

\subsection{Color}
We show in \cref{tab:color} that re-introducing color yields no
benefit (consistent with the
findings of Zhang~\etal~\cite{zhang2016colorful}). However, concurrent work~\cite{zhang2017split}
presents a better method of leveraging the color channels by separately
training a network for the ``opposite'' task (predicting intensity from color). The
two separate networks are combined for downstream use.

\section{Conclusion}

Using the high-level semantic requirements of automatic colorization, we have
proposed it as a drop-in replacement for ImageNet pretraining. This method presents
state-of-the-art results on semantic segmentation and small-sample
classification that do not use ImageNet labels. Our investigation is focused,
but not limited in purpose, to automatic colorization. The investigation into
the benefits of network architecture, training scheduling, loss, etc., may benefit
self-supervision in general.
A detailed investigation into
self-supervised colorization shows the importance of the loss, network
architecture and training details in achieving competitive results. We
conjecture that many of these findings are relevant beyond colorization and
will benefit the broader self-supervision effort. We also draw parallels
between this and ImageNet pretraining, by exploring how robust labeled data is
for representation learning. This study shows that self-supervision is on par
with several methods using annotated data.

\setcounter{chapter}{3}
\chapter{Multi-Proxy Representation Learning}
\label{chp:multiproxy}

In this chapter we explore the possibility of combining several self-supervised
proxy tasks to evaluate if it can lead to more robust and general-purpose
representations.

\section{Introduction}

Colorization can act as a proxy task for visual representation learning, as demonstrated in \cref{chp:selfsup}.
However, we have also seen that the
representation that is learned is in many ways specialized for
colorization. It is also unable to successfully use color as a prediction input
in the final model. We note that there are other methods of
self-supervision that achieve competitive results, such as jigsaw solving~\cite{noroozi2016jigsaw} and
optical flow-based segmentation~\cite{pathak2016move}. Other methods have their own idiosyncrasies. For
instance, jigsaw solvers trained on color images learn to respond to color
only through green-purple edges (see \cref{fig:conv-filters}).
We explore the possibility of combining multiple self-supervision
methods, in order to average out these idiosyncrasies and learn a more versatile
and robust representation. As new ideas for self-supervised learning signals
are developed, they may not beat existing methods alone. However, it is still
possible that they collectively add to the improvement of self-supervised
representation learning, if there is a method of combining multiple
self-supervision proxy tasks. We evaluate several techniques, grouped into what
we call \textit{offline} and \textit{online} model aggregation methods:

\textbf{Offline.} An offline method combines
multiple already trained models into one and does not revisit the original task
loss. The main benefit of such system is that the research and engineering
effort is distributed. Taking the research community as an example, a group of
researchers, using their own favorite deep learning software, can develop a new
self-supervision method and put the model online. This could then be downloaded by separate
researchers and used to build stronger self-supervised
representations. We will consider the offline methods model distillation
(\cref{sec:distillation}) and model concatenation (\cref{sec:splitnetwork}).

\textbf{Online.} Online methods involve actively training several methods
together. Unlike offline methods, they do not have the benefits of distributed
effort. However, the potential benefits of being able to train the network with all
methods actively involved may warrant this sacrifice. It does have the
engineering downside that each method has to be re-implemented into a single
framework. We will consider online methods where base network weights from
different proxy tasks are disjoint (\cref{sec:splitweights}) or shared
(\cref{sec:joint}).

We show that offline and online methods both show promise on test benchmarks
based on supervised pretraining where the level of diversity between tasks is
controlled.  However, this does not yet translate to the self-supervised setting,
suggesting that the level of diversity between representations is perhaps not as
high as we initially conjectured, or simply not high enough for current methods
to appreciate.

Contributions in this chapter include:

\begin{itemize}
    \item A new model distillation loss that works on both probabilistic and
        non-probabilistic activations, with much less need to tune the
        temperature parameter.

    \item We formulate \textit{task-agnostic model distillation} and propose an effective
        method to learn from multiple teachers that were not trained for a single task.

    \item We diagnose the issue of imbalanced gradient norms when combining
        multiple tasks and propose an easy but effective fix using global
        gradient normalization.

    \item We implement three self-supervision methods into a single framework
        and evaluate joint training. In addition to colorization and jigsaw
        solving, we propose a new simple self-supervision method based on
        predicting optical flow in videos.
\end{itemize}


\section{Related Work}
The two primary areas that we relate our work to is model distillation (offline
method) and multi-task learning (online method).

\textbf{Model Distillation.} 
Transferring the information of a large network (often an ensemble of several networks) into a smaller one is
called \textit{model compression}, \textit{model distillation}, or
\textit{student-teacher learning}. The process involves passing samples through
both the teacher (the large network) and the student (the small network),
updating only the student so that its top-level activations closely match the
teacher's activation as specified by a distance or divergence loss. The task was introduced by
Bucilu\v{a}~\etal~\cite{bucilua2006model} in 2006. Their approach involved building a
large ensemble of networks and then using the hard predictions of the ensemble to
label a large sample of synthetic data. This new dataset was subsequently used
to train a smaller network.

The concept of model compression attracted renewed attention 
a decade later. This time, neural networks had become ubiquitous and the
ambition grew to run them on low-power devices, such as mobile phones or
hearing aids. Ba and Caruana published a follow-up work in
2014~\cite{ba2014deep} where they suggest, as a replacement for hard targets,
using squared $L_2$ or softmax and KL divergence to penalize discrepancies between the
student and the teacher. Their main conclusion is that it is possible to take a
multi-layer neural network and compress it into a single layer. However,
Urban~\etal~\cite{urban2016dodeepsfollowup} revisit this claim and demonstrate
evidence to refute it, especially in the context of convolutional neural
networks.  This does not invalidate the concept of model compression however,
since it is still possible to make student networks significantly smaller
without going all the way to single-layer networks.



In the same wave of renewed attention, Hinton~\etal~\cite{hinton2015distilling} promoted the work of
Caruana and colleagues in a NIPS workshop contribution. In
addition, they contributed the notion of temperature, a loss parameter that
allows gradually trading off between hard target predictions and squared $L_2$.




More broadly, it has also become a trend to approximate non-feedforward models into
neural networks, which can be seen as a form of model compression. The
source is typically a costly iterative algorithm and the purpose is to make it
faster and potentially generalize better. Approximating a hand-coded algorithm
by a neural network is also a way to make it more versatile, allowing
fine-tuning on downstream tasks. To mention a few,
Liang~\etal~\cite{liang2008structure} initially trains a highly structured Conditional
Random Field (CRF) with costly inference. To make it cheaper, they subsequently train a
regression model on samples labeled by the CRF. Pathak~\etal~\cite{pathak2016move}
train a CNN to predict the results of an unsupervised segmentation algorithm based
on optical flow and a graph-based algorithm.

\textbf{Multi-Task Learning.}
The method for our primary online method comes from Caruana's doctoral thesis
``Multitask Learning,''~\cite{caruana1998multitask} where he trains multiple
tasks at once using neural networks with shared weights. Much follow-up work
under this name has focused on single-layer neural networks or kernel
methods~\cite{evgeniou2004regularized,argyriou2008convex,chen2011integrating,kumar2012learning}.
There is recent work using CNNs focusing on optimizing the performance on all
tasks together, for instance by making the exact branching point more
fluid~\cite{misra2016cross} or by using tensor factorization to separate
general and task-specific features~\cite{yang2016deep}. This is different from
our goal, since we do not care about the proxy task performance (only insofar
as it is correlated with representation performance) and we prefer to keep the
base network as standard as possible. Our approach follows that of
Caruana's original and the more recent UberNet~\cite{kokkinos2016ubernet}:
training all tasks with the same base network and a minimal number of task-specific layers for each task.
UberNet uses a single network to train a wide range of supervised vision tasks,
both high-level (\eg, semantic segmentation), and more low-level (\eg, boundary
detection).

\section{Model Distillation}
\label{sec:distillation}

Student-teacher learning is the act of distilling the knowledge of a teacher
network into a, typically smaller, student network. A common choice of teacher
is an ensemble of separately trained networks:
%
%
\[
    f_\mathrm{teacher}(x) = \frac 1M \sum_m f_m(x)
\]
An ensemble often leads to better prediction results than each individual model
alone. The improvement depends on the correlation of errors between models,
with lower correlation resulting in larger gains~\cite{ueda1996generalization}.
Inference in the ensemble, however, is $M$ times as expensive as a single network. The
ambition of model distillation is to compress it back down to a single network
with fast inference, while preserving, to the extent possible, the predictive
benefits of the ensemble.

A popular loss for distillation is KL divergence after a temperature-adjusted softmax
\begin{equation} \label{eq:hinton-loss}
    L(x) = T^2 D_\mathrm{KL}(\softmax(\vy/T) \| \softmax(\vz/T)),
\end{equation}
where $\vy = f_\mathrm{teacher}(x)$ and $\vz = f_\mathrm{student}(x)$.
The temperature, $T$,
allows control over the relative importance between high scoring and low scoring classes.
In the temperature limits two special cases merge: squared $L_2$ at $T \to
\infty$ (low and high scores are equally important), and hard predictions at $T
\to 0$ (only highest scoring class matters).  At test time, we often care only
about hard predictions, which is arguably why this is the only information we
need to preserve from the teacher. However, a squared $L_2$ has shown to
provide better generalization properties for out-of-sample predictions, since
it conveys a more nuanced view of the teacher.  Temperature introduces control
over this trade-off.

\subsection{Task-Agnostic Distillation}

If we train several models on different tasks, it is not possible to average
scores into an ensemble. Is it still possible to do student-teacher type distillation? We
call this \textit{task-agnostic} student-teacher learning and we will demonstrate that it
is possible and can yield positive results. The solution that we propose is to move the target
layer from the task layer to the top non-task layer (\eg, \texttt{fc7} in VGG-16). Two
issues need to be addressed: (1) We need a distillation loss that is
appropriate and works well on non-probabilistic feature activations, and (2) we
need a method of combining multiple networks into a single teacher signal that
is compatible with the student network. The first problem is addressed by
introducing a new distillation loss in \cref{sec:copyloss} and the second by
sampling subsets of features in \cref{sec:feature-compatibility}.




\subsection{New Distillation Loss}
\label{sec:copyloss}


We propose a new distillation loss that can be used for both task-driven (distillation at task layer) and
task-agnostic (distillation at top intermediate layer) student-teacher learning. It closely resembles temperature-adjusted KL
divergence, with a few important differences. In \cref{sec:abdiv}, we further show that this is a
special case of the Alpha-Beta (AB) divergence and demonstrate how it can be
configured to achieve a variety of properties.

The new distillation loss is defined as follows:
\begin{equation} \label{eq:new-loss}
    L_\mathrm{new}(x; T) = \frac{T^2}{C} D_\mathrm{KL}(\exp(\hat{\vy}/T) \| \exp(\hat{\vz}/T))
\end{equation}
The normalized exponential (softmax) in \eqref{eq:hinton-loss} has been changed to an unnormalized
exponential. Division by the number of features, $C$, makes the scale of the loss invariant to $C$ and
the loss easily interpretable as the average loss per feature.
Using unnormalized exponentials
produces unnormalized measures, prompting the use of the \textit{generalized}
KL divergence~\cite{kullback1951information}:
\begin{equation}
    D_\mathrm{KL}(\vp \| \vq) = \sum_i \left[ p_i \log \frac{p_i}{q_i} - p_i + q_i\right] \qquad \vp, \vq \in \mathbb{R}_+^C
\end{equation}
The features are instead pre-normalized using statistics of the teacher signal:
\begin{equation} \label{eq:norm}
    \hat \vy = \left( \frac{y_1}{s_1}, \dots, \frac{y_C}{s_C} \right), \quad
    \hat \vz = \left( \frac{z_1}{s_1}, \dots, \frac{z_C}{s_C} \right), \quad
    s_c = \sqrt{\mathbb{\hat{E}}[y_c^2]}\,\,\text{ for all } c
\end{equation}
The normalization has two purposes. First, since temperature effectively
adjusts feature scale, this is a way to establish a universal temperature
scale that does not change depending on the natural scale of the network.
This can optionally be done with a single teacher statistic used for all
features, $s = \sqrt{\mathbb{\hat{E}}[y^2]}$.
However, if the teacher vector is a concatenation of several networks with
different natural scale, individual statistics are preferred. We can also
optionally remove the normalization from $\vz$, which would cause the student
to learn well-normalized top activations, without trying to mimic the teacher
precisely.

A side-by-side comparison of the change in gradients show how similar the
updated loss is ($L_\mathrm{old}$ is $L$ from \eqref{eq:hinton-loss}):
\begin{align*}
    \frac{\partial L_\mathrm{old}}{\partial z_i} = 
    \displaystyle T \left( \frac{\e^{z_i/T}}{\sum_j \e^{z_j/T}} - \frac{\e^{y_i/T}}{\sum_j \e^{y_j/T}} \right)
    \qquad\qquad
    \frac{\partial L_\mathrm{new}}{\partial z_i} = 
    \displaystyle T \left( \frac{\e^{\hat z_i / T}}{C} - \frac{\e^{\hat y_i / T}}{C} \right)
\end{align*}
Apart from feature pre-normalization, the only difference is that the sums over
all exponentials have been replaced by $C$. We note that both normalization
sums in the old loss approaches $C$ as $T \to \infty$.
Removing the cross-terms in the denominator means re-visiting the intuition for temperature. Instead of
affecting the entropy of the activation vector, it simply changes the shape of
the loss that operates on each activation separately. This makes it easy to
visualize and promotes better intuition.

The old loss was shift invariant for both student and teacher activations. This
means that if we add a constant to $\vy$, the loss does not change. This is
independently true for $\vz$, which means that the loss does not drive the
student and the teacher to have the same mean over activations. The new loss
instead forces this, which is an innocuous change for task-driven distillation that is
easily accommodated by biases in the student. For task-agnostic distillation
however, the change is important since when the network is finally used for a
downstream task, the top activations are passed through a ReLU before a task
prediction layer. The ReLU is not shift invariant, so it is important that the
student activations have the same mean as the teacher activations.

\subsection{AB Divergence}
\label{sec:abdiv}

The AB divergence is borrowed from work in non-negative matrix
factorization~\cite{abdiv}. It takes unnormalized measures and has two
parameters, $(\alpha, \beta) \in \mathbb{R}^2$, that control the properties of
the divergence. The divergence is defined as a sum over the elements of the
measures, without cross-terms, as
\begin{equation}
    D_\mathrm{AB}^{(\alpha, \beta)}(\vp \| \vq) = \sum_i d_\mathrm{AB}^{(\alpha, \beta)}(p_i, q_i),
\end{equation}
where
\begin{equation}
    d_\mathrm{AB}^{(\alpha, \beta)}(p, q) = \begin{cases}
        -\frac 1{\alpha\beta} \left( p^\alpha q^\beta - \frac{\alpha}{\alpha + \beta} p^{\alpha + \beta} - \frac{\beta}{\alpha + \beta} q^{\alpha + \beta}\right)
            & \alpha, \beta, \alpha + \beta \not= 0 \\
        \frac 1{\alpha^2} \left( p^\alpha \log \frac {p^\alpha}{q^\alpha} - p^\alpha + q^\alpha \right)
            & \alpha \not= 0, \beta = 0 \\
        \frac 1{\alpha^2} \left( \log \frac {q^\alpha}{p^\alpha} + \left( \frac{q^\alpha}{p^\alpha}\right)^{-1} -1 \right)
            & \alpha = -\beta \not= 0 \\
        \frac 1{\beta^2} \left( q^\beta \log \frac {q^\beta}{p^\beta} - q^\beta + p^\beta \right)
            & \alpha = 0, \beta \not= 0 \\
        \frac 12 (\log p - \log q)^2 & \alpha, \beta = 0. \\
    \end{cases}
\end{equation}
For readers familiar with the R\'enyi divergence (also known as the Alpha
divergence)~\cite{renyi1961measures}, the AB divergence can be interpreted as
extending it with the notion of temperature.
The sum $\alpha + \beta$ of the AB divergence is the inverse temperature and controls the
importance of large activations.

\begin{figure}
    \begin{minipage}{0.6\linewidth}
        \begin{center}

\begin{tikzpicture}[thick,>=latex,scale=4]
    \tikzset{main line/.style={very thick}}
    \def\axisfr{1.65}
    \def\axisfrb{0.4}
    \def\axisfrom{-1.65}
    \def\axisto{1.65}
    \def\diagscale{0.8}
    \def\invpos{0.55}

    \draw[main line,->] (0, -\axisto*\diagscale) to (0, \axisto) node[below right] {$\beta$};

    \draw[main line,->,color=green!50!black] (-\axisfrb,0) .. controls (0,0) .. (\axisto,0) node[above left,color=black] {$\alpha$};

    \draw[main line,color=blue](-\axisfrb,\axisfrb) .. controls (0, 0) ..
                    (\diagscale*\axisfr,-\diagscale*\axisfr)
        node[below left,pos=0.97,color=blue] {\textbf{\textsf{Scale Invariant}}};;

    \draw[main line,color=red] (-\axisfrb, -\axisfrb) .. controls (0, 0) ..
                    (\diagscale*\axisto, \diagscale*\axisto) 
        node[above left,pos=0.97,color=red] {\textbf{\textsf{Symmetric}}};

    \fill (0, 0) circle (0.55pt)
        node[below right=4.7pt,
             chamfered rectangle,
             chamfered rectangle xsep=5pt,
             fill=white,
             fill opacity=0.75,
             text opacity=1] {$\DE(\log \vp\|\log \vq)$};

    \fill (1, 0) circle (0.55pt)
        node[above] {$\DKL(\vp\|\vq)$} 
        node[below] {1}
        ;

    \fill (1, -1) circle (0.55pt)
        node[below left] {$\DIS(\vp\|\vq)$} 
        ;

    \fill (1, 1) circle (0.55pt)
        node[above left] {$\DE(\vp\|\vq)$} 
        ;

    \fill (0, 1) circle (0.55pt)
        node[left] {$\DKL(\vq\|\vp)$}
        node[right] {1};
        ;

    \fill (0, -1) circle (0.55pt)
        node[right] {-1};
        ;



    \node[double arrow,draw,fill=white!80!yellow,rotate=-45] at (\invpos, \invpos) {\textsf{\textbf{Duality}}};

    \node[double arrow,draw,fill=white!80!yellow,rotate=45] at (\invpos, -\invpos) {\textsf{\textbf{Inversion}}};

\end{tikzpicture}

        \end{center}
    \end{minipage}
    \begin{minipage}{0.4\linewidth}
    \caption[AB divergence]{
    \textbf{AB divergence.}
    Adjusting $(\alpha, \beta)$ changes the properties of the divergence. The red
        line $\alpha=\beta$ describes symmetric divergences, where $D(\vp\|\vq) = D(\vq\|\vp)$. Reflecting
        a point around this duality line causes arguments to be swapped
        (\eg, the reverse KL divergence is found at $(0, 1)$). The blue line
        $\alpha=-\beta$ describes scale invariant divergences, $D(\vp\|\vq) =
        D(c\vp\|c\vq)$, which results in shift invariance in the feature space.
        Two points reflected around this inversion line describes two scale
        variant divergences: one that penalizes larger values more and one that
        penalizes smaller values more.
        This figure is a replica of one in the original AB divergence paper~\cite{abdiv}.
}
    \label{fig:ab-div}
\end{minipage}
\end{figure}

\begin{figure}
    \centering
    \begin{tabular}{cc}
        \includegraphics[width=0.48\linewidth]{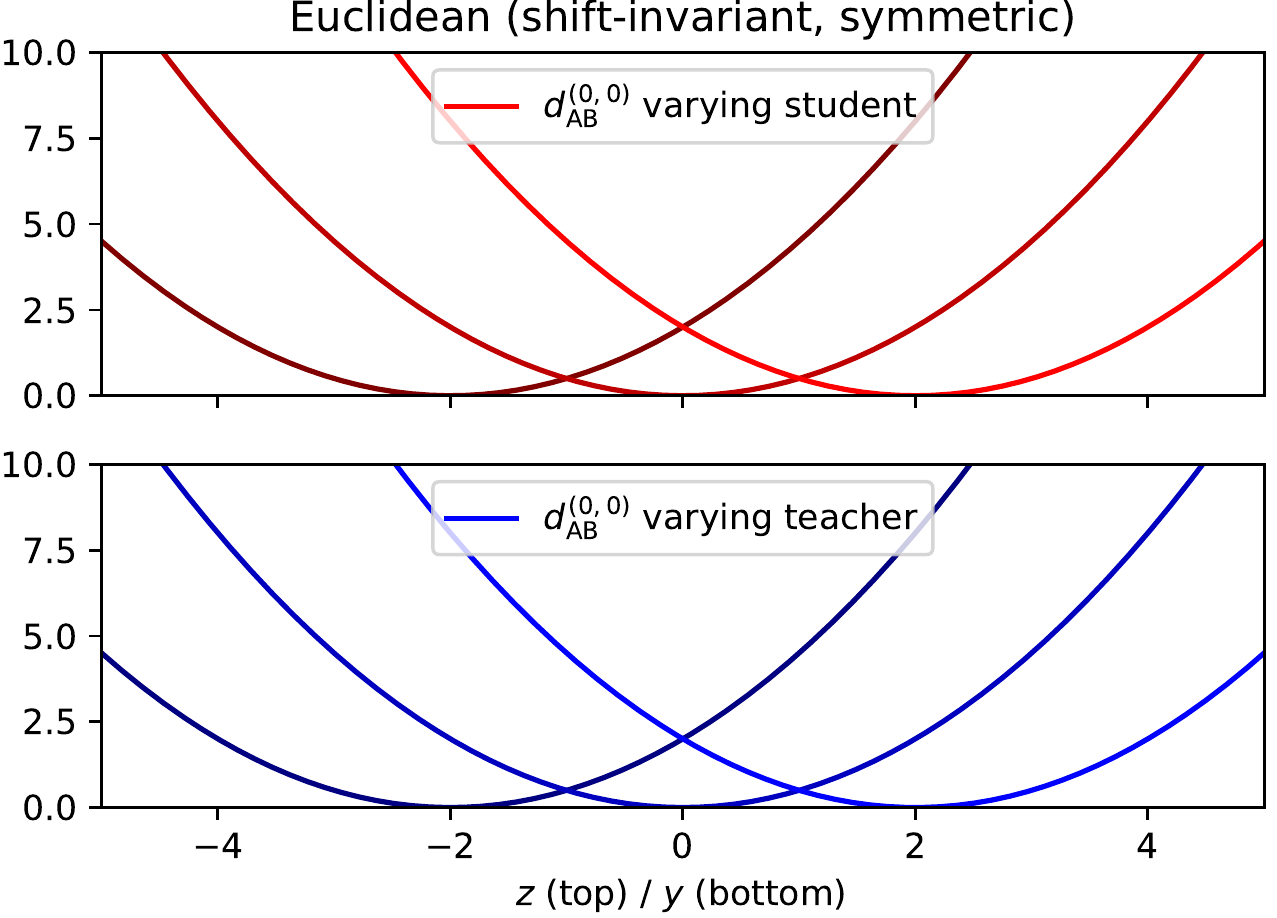}&
        \includegraphics[width=0.48\linewidth]{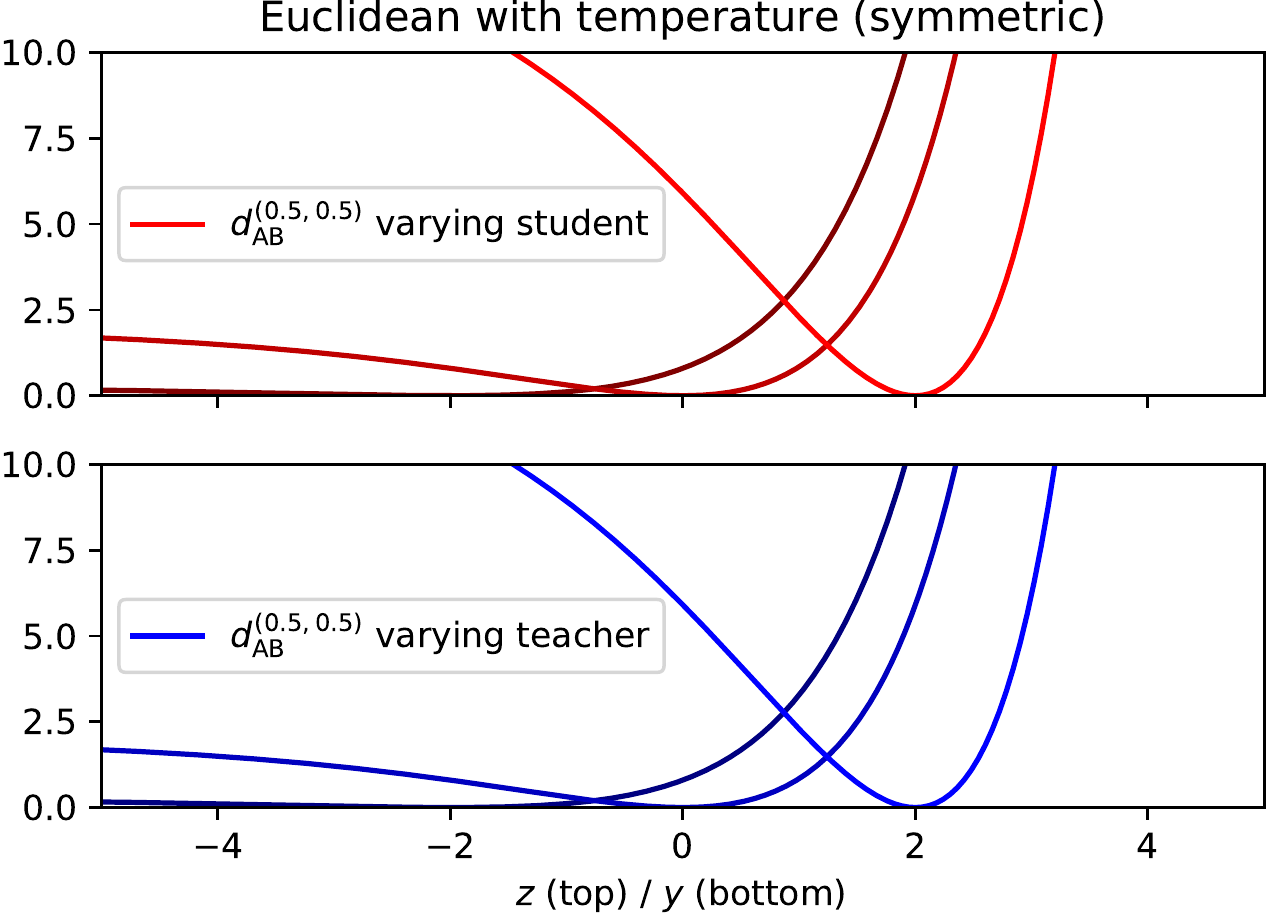}\\
        \includegraphics[width=0.48\linewidth]{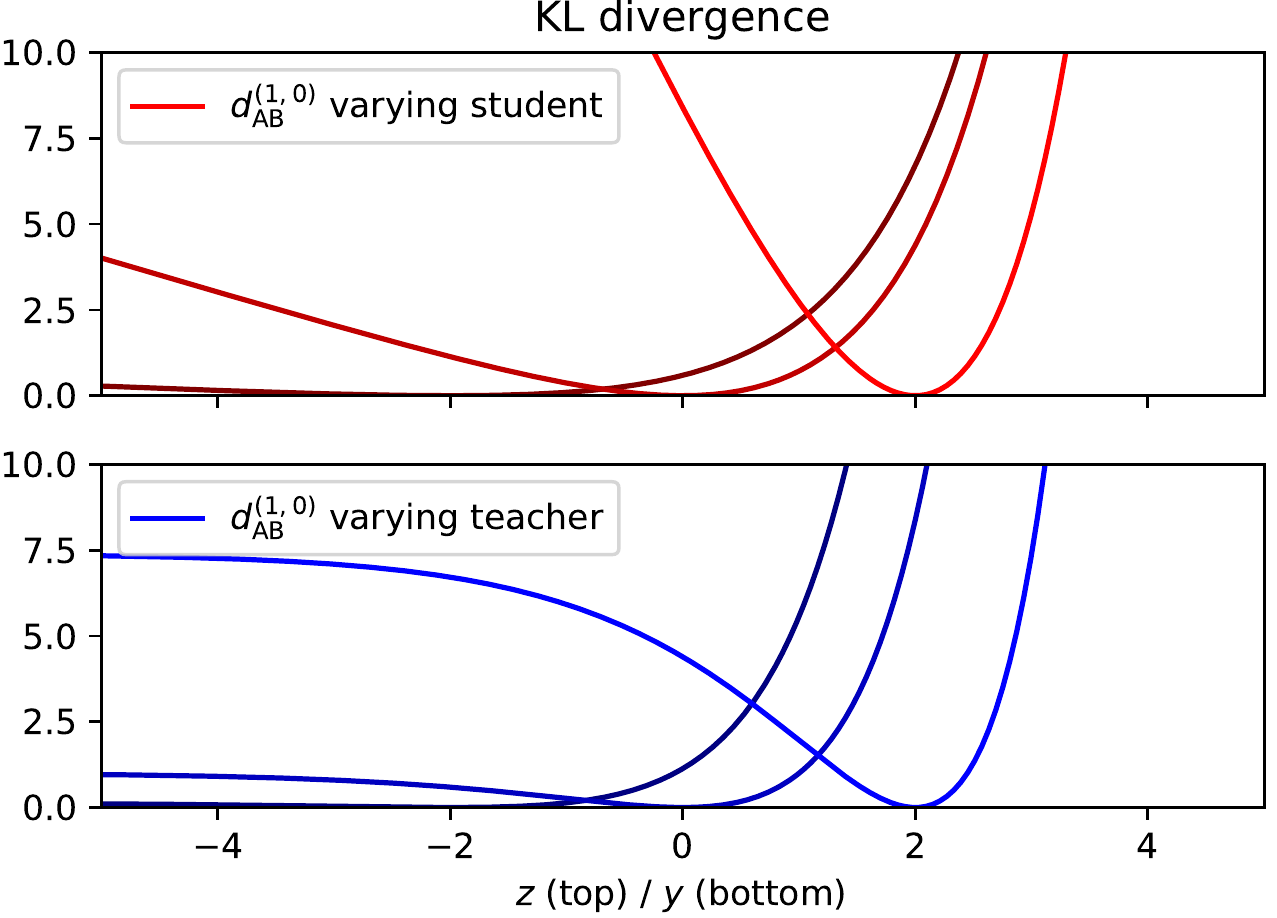}&
        \includegraphics[width=0.48\linewidth]{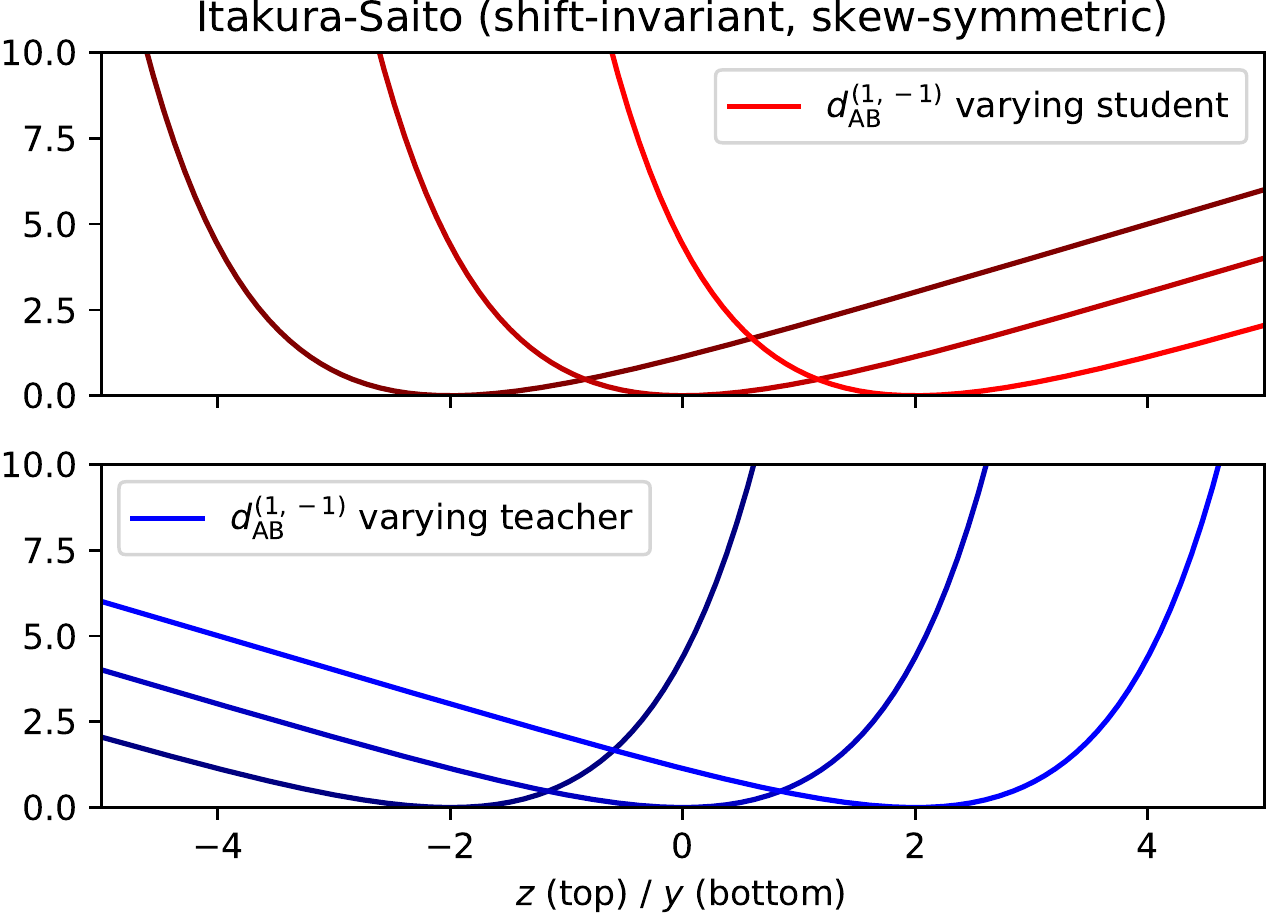}\\
    \end{tabular}
    \caption[Shape of AB divergence]{
        \textbf{Shape of AB divergence.} Four AB divergences coupled with
        exponential activations are visualized in two plots each. The top (red)
        shows three anchor points of the teacher score (-2, 0, 2) while varying
        the student score. The bottom (blue) shows three anchor points of the
        student score (-2, 0, 2) while varying the teacher score. The anchor
        points are not labeled, but are easy to identify since they reach
        exactly 0 only at the anchor value. The divergence is symmetric if the
        red and blue plots are the same ($D(p\|q) = D(q\|p)$) and
        shift-invariant if the plots for the three anchor points have the same
        shape.  These four divergences are evaluated quantitatively in
        \cref{fig:ab-div-exp}.
    }
    \label{fig:ab-div-shapes}
\end{figure}

The new loss presented in \eqref{eq:new-loss} is a special case where $\beta = 0$:
\begin{equation}
    L_\mathrm{new}(x; T) = \frac 1C D_\mathrm{AB}^{(1/T, 0)}(\exp(\hat\vy)\| \exp(\hat\vz))
\end{equation}
It even provides the $T^2$ scale correction from \eqref{eq:hinton-loss} and
\eqref{eq:new-loss}. By adjusting both $\alpha$ and $\beta$ we can explore a
variety of divergences. First of all, at $(0, 0)$, coupled with the exponential
activation function, we get the squared $L_2$ loss. For the
traditional loss \eqref{eq:hinton-loss}, this was only true in the limit $T \to
\infty$ and when both feature vectors were assumed to have zero mean. Furthermore,
by moving along the $\alpha = \beta$ line, we get a symmetric divergence with
the notion of temperature.
If we instead move along the $\alpha = -\beta$ line, we preserve
shift invariance in $\vy$ and $\vz$, while changing the balance between how
much we penalize over-estimation versus under-estimation. Since the divergence
has no cross terms and $d_\mathrm{AB}$ operates on scalars, it makes it easy to
visualize the loss. See \cref{fig:ab-div} for a schematic of the parameter space and
\cref{fig:ab-div-shapes} for concrete examples visualized.

\subsection{Feature Compatibility}
\label{sec:feature-compatibility}

Now that we have a loss that works well on intermediate activations, we move on
to the second problem of task-agnostic distillation: compatibility of the
feature vector between the student and the teacher. For instance, consider a
teacher that is the concatenation of $F$ top activations of $M$ identical
architectures trained on different tasks. Such teacher will now have $MF$ features
and thus not compatible with the student, which has only $F$. We
consider several solutions to address this.

\textbf{Adapt student.} 
A simple way is to allow the student to have the same number of features as the
teacher at the necessary layer. Since this only affects one layer, it is still
possible to make the student smaller in terms of parameters. However, this
skews the network more and more as the teacher grows (\eg, as $M$ increases),
and it makes fair comparisons between methods harder. For these reasons, we
will restrict ourselves to methods that can abide by the student architecture
without changing it.

\textbf{Auxiliary layer.}
One way of addressing this is to pass the $F$ features of the student through a
fully connected layer with $MF$ output activations. The distillation loss is
then applied on top of this additional layer, where both student and teacher
have $MF$ features. The extra layer is discarded after distillation training,
so it would be unfortunate if it has captured important high-level semantics.
Note that a ReLU is applied before the fully connected layer, since we want the
top activations of the student to be able to pass through a ReLU with
meaningful results.

\textbf{Feature subset.}
Another approach, that admittedly seems less powerful, is to take a random
subset of $F$ from the teacher's $MF$ features. This gives better results
than any auxiliary layer approach that we tried (see \cref{tab:distillation}).
Although not experimentally tested here, a slight improvement might be offered
by first training the student with a wide layer (equivalent to ``adapt
student'') and then subsetting the features to reduce it to the target network.
This would allow all features of the teacher vector to drive representation
learning for all previous layers of the network.

\subsection{Data Augmentation, Dropout, and Warm Start}

Even with the distillation loss and network compatibility addressed, it can still be a challenge to perform a
successful model distillation. Since training data for distillation does not
need annotations, this can and should be drawn from a large data source. However,
it is important to be mindful of the data distribution. For instance, a preliminary investigation
showed that ImageNet downsampled to 32-by-32 was not a good source of unlabeled distillation samples for
a model originally and eventually operating on CIFAR-100. However, using data
augmentation on CIFAR-100 during the distillation phase is helpful.

When distillation data is limited, we found that the best source of
regularization is dropout in the student network (see \cref{sec:copytasks}).
Adding dropout in the teacher has a limited effect on results, so we do not do
it for any of the experiments.

Finally, in the case where the teacher is $M$ networks of the same architecture
as the student, it may help to warm start the student with one of the teacher
networks. We will see that in our experiments in
\cref{tab:distillation} this offers only a minor improvement, which is a
good sign that our distillation training trains well on its own.




\section{Disjoint Weights Training}

We move on to other aggregation methods, namely two where each parameter in the
final network is associated with a single task. This prevents multiple tasks
from operating on the same parameter, which is associated with an issue of uneven
gradients described in \cref{sec:uneven}.


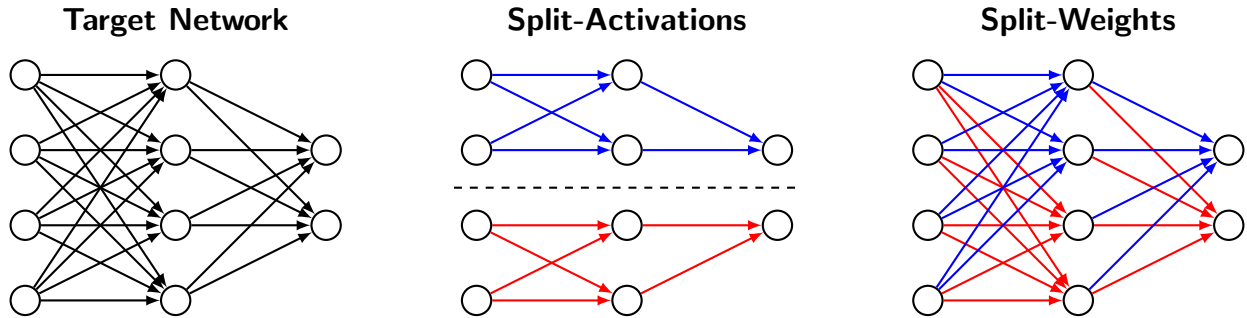
\begin{figure}
    \centering
    \begin{tikzpicture}[thick,>=latex,scale=1]
    \tikzset{main line/.style={very thick}}
    \tikzstyle{flow arrow}=[->]
    \tikzstyle{loss arrow}=[->,dashed]
    \tikzstyle{layer}=[rectangle,draw,thick,fill=purple!50,thin,minimum width=1.5cm,minimum height=5]

    \pgfmathsetmacro{\layerone}{4}
    \pgfmathsetmacro{\layeronehalf}{2}
    \pgfmathsetmacro{\layeronehalfplus}{3}
    \pgfmathsetmacro{\layertwo}{4}
    \pgfmathsetmacro{\layertwohalf}{2}
    \pgfmathsetmacro{\layertwohalfplus}{3}
    \pgfmathsetmacro{\layerthree}{2}
    \pgfmathsetmacro{\layerthreehalf}{1}
    \pgfmathsetmacro{\layerthreehalfplus}{2}
    \pgfmathsetmacro{\ytext}{2.2}
    \pgfmathsetmacro{\ydiff}{1.0}


    \pgfmathsetmacro{\basex}{0.0}

    \foreach \i in {1, ..., \layerone} {
        \pgfmathsetmacro{\x}{\basex+0.0}
        \pgfmathsetmacro{\y}{\ydiff*(\i - (\layerone+1)/2)}
        \node[circle,draw,thick,fill=white] (x-1-\i) at (\x, \y) {};
    }

    \foreach \i in {1, ..., \layertwo} {
        \pgfmathsetmacro{\x}{\basex+2.0}
        \pgfmathsetmacro{\y}{\ydiff*(\i - (\layertwo+1)/2)}
        \node[circle,draw,thick,fill=white] (x-2-\i) at (\x, \y) {};
    }

    \foreach \i in {1, ..., \layerone} {
        \foreach \j in {1, ..., \layertwo} {
            \draw[flow arrow] (x-1-\i) -- (x-2-\j);
        }
    }

    \foreach \i in {1, ..., \layerthree} {
        \pgfmathsetmacro{\x}{\basex+4.0}
        \pgfmathsetmacro{\y}{\ydiff*(\i - (\layerthree+1)/2)}
        \node[circle,draw,thick,fill=white] (x-3-\i) at (\x, \y) {};
    }

    \foreach \i in {1, ..., \layertwo} {
        \foreach \j in {1, ..., \layerthree} {
            \draw[flow arrow] (x-2-\i) -- (x-3-\j);
        }
    }
    
    \node at (\basex+2, \ytext) {\textbf{\textsf{Target Network}}};


    \pgfmathsetmacro{\basex}{6}

    \foreach \i in {1, ..., \layerone} {
        \pgfmathsetmacro{\x}{\basex+0.0}
        \pgfmathsetmacro{\y}{\ydiff*(\i - (\layerone+1)/2)}
        \node[circle,draw,thick,fill=white] (x-1-\i) at (\x, \y) {};
    }

    \foreach \i in {1, ..., \layertwo} {
        \pgfmathsetmacro{\x}{\basex+2.0}
        \pgfmathsetmacro{\y}{\ydiff*(\i - (\layertwo+1)/2)}
        \node[circle,draw,thick,fill=white] (x-2-\i) at (\x, \y) {};
    }

    \foreach \i in {1, ..., \layeronehalf} {
        \foreach \j in {1, ..., \layertwohalf} {
            \draw[red,flow arrow] (x-1-\i) -- (x-2-\j);
        }
    }

    \foreach \i in {\layeronehalfplus, ..., \layerone} {
        \foreach \j in {\layertwohalfplus, ..., \layertwo} {
            \draw[blue,flow arrow] (x-1-\i) -- (x-2-\j);
        }
    }

    \foreach \i in {1, ..., \layerthree} {
        \pgfmathsetmacro{\x}{\basex+4.0}
        \pgfmathsetmacro{\y}{\ydiff*(\i - (\layerthree+1)/2)}
        \node[circle,draw,thick,fill=white] (x-3-\i) at (\x, \y) {};
    }

    \foreach \i in {1, ..., \layertwohalf} {
        \foreach \j in {1, ..., \layerthreehalf} {
            \draw[red,flow arrow] (x-2-\i) -- (x-3-\j);
        }
    }

    \foreach \i in {\layertwohalfplus, ..., \layertwo} {
        \foreach \j in {\layerthreehalfplus, ..., \layerthree} {
            \draw[blue,flow arrow] (x-2-\i) -- (x-3-\j);
        }
    }

    \draw[dashed] (\basex-0.3, 0.0) -- (\basex + 4.3, 0.0);

    \node at (\basex+2, \ytext) {\textbf{\textsf{Split-Activations}}};


    \pgfmathsetmacro{\basex}{12}

    \foreach \i in {1, ..., \layerone} {
        \pgfmathsetmacro{\x}{\basex+0.0}
        \pgfmathsetmacro{\y}{\ydiff*(\i - (\layerone+1)/2)}
        \node[circle,draw,thick,fill=white] (x-1-\i) at (\x, \y) {};
    }

    \foreach \i in {1, ..., \layertwo} {
        \pgfmathsetmacro{\x}{\basex+2.0}
        \pgfmathsetmacro{\y}{\ydiff*(\i - (\layertwo+1)/2)}
        \node[circle,draw,thick,fill=white] (x-2-\i) at (\x, \y) {};
    }

    \foreach \i in {1, ..., \layerone} {
        \foreach \j in {1, ..., \layertwohalf} {
            \draw[red,flow arrow] (x-1-\i) -- (x-2-\j);
        }
    }

    \foreach \i in {1, ..., \layerone} {
        \foreach \j in {\layertwohalfplus, ..., \layertwo} {
            \draw[blue,flow arrow] (x-1-\i) -- (x-2-\j);
        }
    }

    \foreach \i in {1, ..., \layerthree} {
        \pgfmathsetmacro{\x}{\basex+4.0}
        \pgfmathsetmacro{\y}{\ydiff*(\i - (\layerthree+1)/2)}
        \node[circle,draw,thick,fill=white] (x-3-\i) at (\x, \y) {};
    }

    \foreach \i in {1, ..., \layertwo} {
        \foreach \j in {1, ..., \layerthreehalf} {
            \draw[red,flow arrow] (x-2-\i) -- (x-3-\j);
        }
    }

    \foreach \i in {1, ..., \layertwo} {
        \foreach \j in {\layerthreehalfplus, ..., \layerthree} {
            \draw[blue,flow arrow] (x-2-\i) -- (x-3-\j);
        }
    }

    \node at (\basex+2, \ytext) {\textbf{\textsf{Split-Weights}}};

\end{tikzpicture}
    \caption[Schematic of disjoint weights training]{
        \textbf{Schematic of disjoint weights training.}
        The target network represents the base neural network architecture.
        The split-activations method (\cref{sec:splitnetwork}) concatenates separately trained networks that have been made thinner in order to conform to the target network. In this example there are two tasks,
        visualized as red and blue, respectively. The thinness of the two networks means we lose the
        weights that would ordinarily cross the dashed line (cross-weights).
        The split-weights method (\cref{sec:splitweights}) is similar to
        split-activations, except it assigns those cross-weights to the
        task associated with the destination node.
    }
    \label{fig:split-schematic}
\end{figure}

\subsection{Split-Activations}
\label{sec:splitnetwork}

The first method we consider is performed offline and simply concatenates
multiple separately trained networks. This technique has been successfully
applied to complementary self-supervision tasks, where it is called the
\textit{split-brain autoencoder}~\cite{zhang2017split}.

For the final concatenated network to resemble our target student, each task network must be
thinner than the original by a factor $1/M$ (see \cref{fig:split-schematic}). This becomes a subset of the
student network, since cross-connections between task columns are removed.
This does not scale well in $M$, since the number of parameters in the combined network will change by
a factor $1/M$. Of course, we could widen the networks accordingly to preserve
the total number of parameters. This is an avenue that makes fair network
comparisons harder, so we do not explore it here.
Note, in our implementation during downstream training, once the network has
been combined, we allow cross-weights to update freely after a minimally
disruptive random initialization.


\subsection{Split-Weights}
\label{sec:splitweights}

Inspired by split-activations, we propose split-weights. Instead of partitioning
the activations at each layer into separate columns, we partition the weights.
This means that, similar to split-activations, no weight in the base network receives
gradients from more than one task. The cross-weights that were
previously discarded are now assigned to the column that they go into. This
means that a task column can look and use activations from other tasks 
and no weights are left untrained (see \cref{fig:split-schematic}).

This offers an improvement over split-activations, even at $M=2$. The gap
widens as $M$ is increased and split-weights loses more and more parameters to
cross-terms.
The downside is that it no longer enjoys the benefits of an offline method and
needs to be trained jointly.


\section{Shared Weights Training}
\label{sec:joint}

Finally, we discuss the method proposed by Caruana~\cite{caruana1998multitask},
to add the losses and use weight sharing for the base network.
We can think of this as keeping separate copies of a base network, each with a task-specific
appendage. A forward-backward pass is performed on each with their respective
task loss. The gradients are finally aggregated and applied to all networks by
element-wise summation. Using this formulation, each task can have its own data
source and different tasks can even operate on differently sized input.\footnote{Fully connected layers are re-interpreted as 1-by-1 convolutional layers.}

\subsection{Uneven Gradients Problem}
\label{sec:uneven}

Initial attempts at training a colorizer and a jigsaw solver together,
resulted in features reminiscent of a network trained only for the jigsaw
task. Upon closer inspection, it turned out that the gradients coming from the
jigsaw loss dominated the colorization loss in terms of size of gradient norms. A consequence of
this can be seen in \cref{fig:conv-filters}, where the first convolutional
filters only resemble those from a jigsaw solver and not from a colorizer.

There are many reasons
why this might happen, especially with such network diversity where one uses a
hypercolumn and the other one does not. Here is a far more simple example
that illustrates how easily this can happen.
Imagine two linear classifiers sitting atop a base network with similar types of tasks.
The weight gradients are directly proportional to the loss, 
$
    \| \nabla_{\mathbf{w}} L \|  \propto L
$, and
it is easy to show that if we change the loss to $L' = \alpha L$, each element
of the gradient is proportionally changed as well:
\begin{equation} \label{eq:loss-multiplier}
    \frac{\partial L'}{\partial w_i} = \frac{\partial L'}{\partial L} \frac{\partial L}{\partial w_i} = \alpha \frac{\partial L}{\partial w_i}
\end{equation}
As a result, we can change the influence of the two losses. However, this should
not lead to the conclusion that because two losses are similarly scaled so will
their gradients be. There are many aspects underlying this discrepancy, such as
choice of loss and structure and initialization of task-specific appendages.

We explore two options for making sure the gradients are well-balanced and that
the contributions are close to equal. Both solutions involve optimization
algorithms that are \textit{loss scale invariant}, meaning the gradients do not
change if we multiply the loss by an arbitrary factor.

\subsection{Separate Adaptive Optimizers} \label{sec:sep-opt}

There are many popular choices of gradient optimizers (\eg, SGD, Adadelta,
Adam). We will phrase these optimizers as being pre-processing routines of the
gradient vector. The network weights are then updated based on the pre-processed gradient
and the learning rate. For instance, in this view of optimizers,
Stochastic Gradient Descent (SGD) is the identity function, since all it does is perform the update step with the raw gradients. The
routine does not need to be a mathematical function and can be a subroutine with internal state.
For instance, if we add
momentum~\cite{qian1999momentum}, the routine is an exponential smoother with an
internal state that keeps track of the gradient of the last time frame. An
adaptive optimizer is when the routine normalizes the gradient and thus becomes
invariant to the loss scale.

In multi-task learning, separate networks are maintained for each task. Forward-backward passes compute the raw gradients for each instance of the base network and these are then aggregated
(by summation or averaging) into a single gradient. This gradient is then passed
through an optimizer pre-processing routine and finally applied to update the weights
of each instance of the base network. If $\optimizer$ is the gradient pre-processing
routine, $\vg_m$ the gradient of task $m$, the final gradient $\vg$ is computed as:
\begin{equation}
    \vg \leftarrow \optimizer\left(\frac 1M \sum_t \vg_m\right)
\end{equation}
One solution to the uneven gradients problem is to first change the order
around, so that the gradient pre-processing is performed separately for each
task before the aggregation:
\begin{equation} \label{eq:sep-opt}
    \vg \leftarrow \frac 1T \sum_m \optimizer_m(\vg_m)
\end{equation}
Now there are $M$ separate optimizers, one for each task $m$. If adaptive
optimizers are used, then any imbalance will be normalized away before
aggregation. We refer to this solution to the uneven gradients problem as using separate
adaptive optimizers.

There are potential issues with this. First, let us discuss some of the popular
adaptive optimizers, such as 
    Adagrad~\cite{duchi2011adaptive},
    Adadelta~\cite{zeiler2012adadelta},
    RMSprop~\cite{tieleman2012rmsprop}, and
    Adam~\cite{kingma2015adam}.
What all these optimizers have in common, apart from being loss scale invariant,
is that they in various ways normalize the size of gradients \textit{per
feature}. This may in some cases cause unwanted acceleration of learning in certain areas of the network. For
instance, in the beginning of training, it is common for early layer gradients
to be particularly small, since changes are dampened by the randomly
initialized network. As a result, it can take a while until the early layers
start updating properly. Adaptive optimizers are more aggressive and will take
small gradients and amplify them. In our experience, this has both the
potential to benefit \textit{and} damage representation learning. We observe this in \cref{tab:cifar}, where Adam
consistently under-performs compared to SGD across all tasks. The aggressiveness
can be softened by introducing a small constant, $\epsilon$, in the gradient
normalization denominator. However, this $\epsilon$ technically breaks loss scale
invariance and the greater the $\epsilon$ is (more softening), the more this
is broken. Typical values for $\epsilon$ are extremely small (\eg, $10^{-5}$) and only meant
to avoid numerical instabilities.



Additionally, as with momentum, many optimizers use an internal state. If we
require a separate optimizer state for each task, we incur additional memory
overhead. For instance, the parameters of VGG-16 takes roughly 0.5 GiB and
corresponding state for an Adam solver takes 1 GiB. To use separate Adam
instances for each task, this number needs to be multiplied by $M$, which
quickly becomes a significant portion of the GPU memory.


\subsection{Globally Normalized SGD (NormSGD)}
\label{sec:norm-sgd}


We propose another solution that makes sure that gradients are balanced between tasks,
using a single optimizer and not requiring an adaptive optimizer. Although
it can be coupled with any optimizer, we will use it with mini-batch
momentum-SGD and refer to the combination as NormSGD.  The trick is to globally normalize
the gradients using the pre-processing function
\begin{equation} \label{eq:global-norm}
    \texttt{norm}(\vg) = \frac{\vg}{\frac 1P \| \vg \|_1} = \frac{\vg}{\mathbb{E}[|g|]} = \frac{\vg}{\frac 1P \sum_p |g_p|},
\end{equation}
where $P$ is the total number of parameters for the entire base network.
This can be combined with other optimizer pre-processing routines (\eg,
momentum) by making subsequent calls.
This normalization is reminiscent of RMSprop~\cite{tieleman2012rmsprop} and Adadelta~\cite{zeiler2012adadelta}, where each gradient
element is divided by its absolute value. However, we instead compute a single
normalization statistic and use it for all elements of the gradient. This preserves the
gradient direction and does not boost individual features with low gradient.

\begin{figure}
    \centering
    \includegraphics[width=0.7\linewidth]{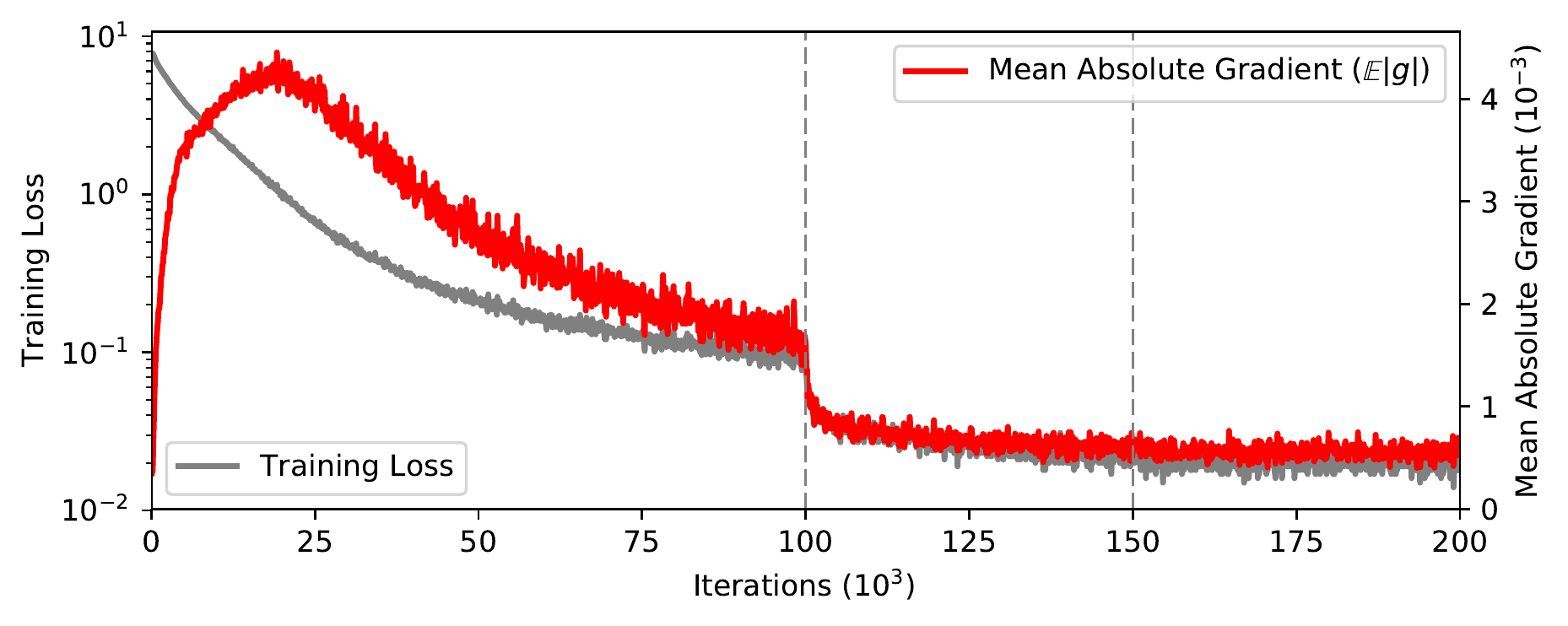}
    \caption[Gradient norm]{
        \textbf{Gradient norm.}
        This shows the loss (in gray) and gradient norm (in red) over time for supervised training
        on CIFAR-100. We draw attention to two observations. First, there is a
        warm-up phase, where gradients quickly increase in norm. Second, after
        this phase the norm drops over time, especially when the learning rate
        is dropped by a factor 10 at 100,000.
    }
    \label{fig:loss-norm}
\end{figure}

However, this still changes the nature of SGD. To see why, we plot in \cref{fig:loss-norm} the size
of the gradient over time when training with standard
SGD. We first observe that there is an acceleration phase where gradients increase in
size as the network starts finding the quickest paths to loss reduction. Next,
as the loss keeps dropping, so does the size of the gradients. In other words,
two things may happen differently when globally normalizing the gradients: (1) It may be a bit too
aggressive in the early stages when the gradient norm is low, and (2) the
learning rate only drops according to schedule without also being naturally
diminished as the loss approaches zero. We offer corrections to these to bring
it even closer to SGD.

The first issue of aggressive starts is minor and can be solved by using a
brief period of learning rate warm up. A fix to the second problem is much more
pressing, since without it we have observed occasional training instabilities
and divergence. This is perhaps because gradients with small norm and low
signal-to-noise ratio get boosted, increasing the risk of jolting the network
into a bad state. The fix is to freeze the gradient norm statistics before
reaching this point, which in our experience can be anytime before dropping the
initial learning rate, but after the initial warm-up phase.

NormSGD can be beneficial for single-task training. Loss scale and learning rate are
according to \cref{eq:loss-multiplier} two equivalent ways to control the size
of gradient steps. NormSGD eliminates this over-parameterization, giving the
control rightly to the learning rate alone. We have observed that it is much
easier to switch between losses, without having to re-calibrate the learning
rate. Furthermore, it can be applied prior to gradient aggregation when training
$T$ tasks simultaneously, as in \eqref{eq:sep-opt}.

What guarantees do we have about the ``contribution'' of each task under this
gradient transformation? By design, all pre-processed task gradients have the same $L_1$ norm, \ie~$\|\optimizer(\vg_i)\|_1 =
\|\optimizer(\vg_j)\|_1$, for all $i, j$. However, if we consider contribution to mean how much of
a single task gradient is preserved in the aggregated gradient (for instance quantified by $\vg \cdot \vg_i$), then we
have no guarantees beyond trivial upper bounds. For instance, in a three-way multi-task loss, we could
theoretically end up having two sets of perfectly canceling gradients, in which case the
aggregated gradient would be only a scale factor different from the third one. Seen in a different way,
each task has the same vote, but a vote does not guarantee influence. This can
be seen as a feature, since it will favor updates with a quorum. We present
evidence in \cref{tab:cifar} supporting the benefits of NormSGD, along with qualitative
results in \cref{fig:conv-filters} that suggest even contributions of tasks.




\subsection{Round-Robin Scheduling}


Despite memory saving of not having separate instances of an adaptive
optimizer, training multiple losses still does not scale well in the number of
tasks $M$. There is work to make back-propagation on computational graphs more
memory efficient~\cite{chen2016training,gruslys2016memory}, used for instance
by UberNet~\cite{kokkinos2016ubernet}.
As an alternative, we can keep only one task active, switching task
in a circular order after a fixed time slice or \textit{quantum} ($Q$).  If we set the
quantum to one iteration ($Q=1$), it will be close to joint training.
However, constant context switching will significantly slow down training.
Ideally, we set the quantum high enough for the context switch to become a negligible or
at least justifiable cost.


A context switch involves marshaling the compute graph for one task and
transferring it from the GPU memory to the CPU memory, as well as reviving another task's
state from the CPU into the GPU. Even if your favorite deep learning software does not
explicitly allow such operations, standard snapshotting to file can be used on a
RAM-mounted file system (\eg, \texttt{ramfs} or \texttt{tmpfs} on Linux).

When using round-robin scheduling, we never perform gradient
aggregation (eq.~\ref{eq:sep-opt}). Nevertheless, similar issues of unbalanced
gradients still apply. The memory overhead of multiple adaptive optimizers is no longer an
issue, which makes them viable. However, if adaptive optimizers do not perform well (as for
us), NormSGD still works well for round-robin scheduling.

\section{Evaluation}

We evaluate the aggregation techniques on a variety of supervised pretraining
tasks created from CIFAR-100. We use supervised tasks because they require less
training and give a clearer signal of the benefits since we can carefully
control the level of potential benefits from multi-task learning.  Once we have developed a method that works
well in this setting, we apply it to self-supervision in
\cref{sec:multiproxy-selfsup}.

\subsection{Distillation}
\label{sec:copytasks}

\begin{table}
    \begin{minipage}[t]{0.49\columnwidth}
        \vspace{0pt}
        \begin{tabular*}{\linewidth}{lllll}
            \multicolumn{5}{l}{Name: \textit{Deep}} \\
            \toprule
            Type    & Size/Stride    & Channels & Dropout \\ \midrule
            Conv    & $5\times 5$/1 & 64       & \\
            Conv    & $3\times 3$/1 & 64       & \\
            Pool    & $2\times 2$/2 & \dash \\
            Conv    & $3\times 3$/1 & 128                 & 0.1 \\
            Conv    & $3\times 3$/1 & 128                 & 0.1 \\
            Pool    & $2\times 2$/2 & \dash \\
            Conv    & $3\times 3$/1 & 192                 & 0.2 \\
            Conv    & $3\times 3$/1 & 192                 & 0.2 \\
            Pool    & $2\times 2$/2 & \dash \\
            Conv    & $3\times 3$/1 & 192                 & 0.3 \\
            Conv    & $3\times 3$/1 & 192                 & 0.3 \\
            Pool    & $2\times 2$/2 & \dash \\
            Flatten & \dash         & 3072                & \\
            Task    & \dash         & 100                 & \\
            \bottomrule
        \end{tabular*}
    \end{minipage}
    ~
    \begin{minipage}[t]{0.49\columnwidth}
        \vspace{0pt}
        \begin{tabular*}{\linewidth}{lllll}
            \multicolumn{5}{l}{Name: \textit{Shallow}} \\
            \toprule
            Type    & Size/Stride   & Channels & Dropout \\ \midrule
            Conv    & $5\times 5$/1 & 8       & \\
            Pool    & $2\times 2$/2 & \dash \\
            Conv    & $3\times 3$/1 & 16                 & 0.1 \\
            Pool    & $2\times 2$/2 & \dash \\
            Conv    & $3\times 3$/1 & 32 & 0.2 \\
            Pool    & $2\times 2$/2 & \dash \\
            Conv    & $3\times 3$/1 & 32                 & 0.3 \\
            Conv    & $3\times 3$/1 & 192                & 0.3 \\
            Pool    & $2\times 2$/2 & \dash \\
            Flatten & \dash         & 3072                & \\
            Task    & \dash         & 100                 & \\
            \bottomrule
        \end{tabular*}
        \caption[Test architectures]{\textbf{Test architectures.} Description
        of architectures \textit{Deep} and \textit{Shallow} for distillation
        and multi-task experiments.
        }
        \label{tab:cnet}
    \end{minipage}
\end{table}

We evaluate our ability to do task-driven and task-agnostic distillation.  For this,
we use two typical convolutional
networks, \textit{Deep} (8 layers) and \textit{Shallow} (5 thin layers)
described in detail in \cref{tab:cnet}. The \textit{top layer} has 3,072
features and feeds into a \textit{task layer} with typically 100 features (depending on the task).
Five distillation scenarios are formulated and described in
\cref{tab:dist-probs}. The prefix \textit{Task} denotes traditional
distillation that occurs
at the task layer, while \textit{Top} indicates task-agnostic distillation at
the top layer. For the latter, we use a held-out portion of the training data to
see how well it can train a linear classifier on the distilled network's top layer. The
problems are briefly summarized as \textit{Copy} (make an exact copy),
\textit{Compress} (compress into smaller network), and \textit{Ensemble}
(compress multiple networks into one).

\begin{table}
\begin{center}
    \begin{tabular*}{\linewidth}{ll@{\,\,\,}p{12cm}}
    \toprule
        Problem & \multicolumn{2}{l}{Description} \\ \midrule
    \textit{Task-Copy}
        & 1. & Train \textit{Deep} on all 50,000 samples \\
        & 2. & Duplicate model at task layer using all 50,000 samples \\
        \midrule
    \textit{Task-Compress}
        & 1. & Train \textit{Deep} on all 50,000 samples \\
        & 2. & Compress into \textit{Shallow} at task layer using all 50,000 samples \\
        \midrule
    \textit{Task-Ensemble}
        & 1. & Train six \textit{Deep} on all 50,000 samples and ensemble \\
        & 2. & Compress into \textit{Deep} at task layer using all 50,000 samples (with data augmentation described in~\cite{larsson2017fractalnet} and no student dropout) \\
        \midrule
    \textit{Top-Copy}
        & 1. & Train \textit{Deep} on first the 40,000 samples \\
        & 2. & Duplicate model at top layer using the first 40,000 samples \\
        & 3. & Add a task layer and train a linear classifier (do not fine-tune end-to-end) using the last 10,000 samples \\
        \midrule
    \textit{Top-Compress}
        & 1. & Train \textit{Deep} on first the 40,000 samples \\
        & 2. & Compress into \textit{Shallow} at top layer using the first 40,000 samples \\
        & 3. & Add a task layer and train a linear classifier (do not fine-tune end-to-end) using the last 10,000 samples \\
        \bottomrule
    \end{tabular*}
    \caption[Distillation problems]{\textbf{Distillation problems.}
        Problems to tests various model distillation scenarios. The
        \textit{Copy} problems are not what we typically refer to as distillation,
        since the information is not transferred into a smaller network.
        However, it still uses the same training mechanisms and it is
        interesting since a perfect copy should be possible, so any in performance
        loss is caused entirely by the training procedure and not by reduced
        model complexity. All problems are evaluated on the
        CIFAR-100 test set. See \cref{fig:ab-div-exp,tab:copyloss}.
    }
    \label{tab:dist-probs}
\end{center}
\end{table}

All models are evaluated on the CIFAR-100 test set.
The distillation loss is evaluated in \cref{tab:copyloss}. Temperature generally offers an
improvement. The gains are greater when the task is harder and we are
compressing into a smaller network. For \textit{Top-Copy}, the network
copy is close to perfect even for squared $L_2$, so temperature offers no
improvement. For \textit{Top-Compress}, temperature helps over squared $L_2$,
but not as much as for \textit{Task-Compress}. 

The results for \textit{Task-Ensemble} show that we can succesfully improve
single-model results from $59.7 \pm 0.3$ to $62.5 \pm 0.1$ by ensemble
training followed by model distillation with temperature. Without temperature,
the improvements are somewhat lower, but still better than the baseline, at 
$61.7 \pm 0.1$.

One of the primary
benefits of the new distillation loss is that it works well in diverse settings
without having to re-tune the parameters $\alpha$ and $\beta$. Across all five
problems, we found that $D_\mathrm{AB}^{(0.25, 0)}$ worked best. The old loss
\eqref{eq:hinton-loss} can in each situation be made close to equivalent, but
it would result in widely different temperatures.

Student dropout may help if training data is scarce. For instance, the two numbers
for \textit{Task-Copy} in \cref{tab:copyloss} drop significantly without
dropout ($57.8 \pm 0.2 \rightarrow 54.4 \pm 0.4$ and $58.6 \pm 0.4
\rightarrow 54.7 \pm 0.3$). However, since we use data augmentation for
\textit{Task-Ensemble}, dropout can cause over-regularization which is why we
do not use it. If we do use it, accuracy goes down ($61.7 \pm 0.1 \rightarrow
57.2 \pm 0.1$ and $62.5 \pm 0.1 \rightarrow 59.0 \pm 0.2$).


\begin{table}
    \newcommand{\xa}[3]{$\num[round-mode=places,round-precision=1]{#1}\pm\num[round-mode=places,round-precision=1]{#2}$}
    \newcommand{\xb}[2]{$\num[round-mode=places,round-precision=1]{#1}\phantom{^{}\pm0.0}$}
    \newcommand{\tb}{\textbf}
    \begin{center}
        \begin{tabular*}{\linewidth}{l|@{\extracolsep{\fill}}rrrrr}
        \toprule Loss
            & \textit{Task-Ensemble}
            & \textit{Task-Copy}
            & \textit{Task-Compress}
            & \textit{Top-Copy}
            & \textit{Top-Compress}
            \\ \midrule
            $D_\mathrm{AB}^{(0, 0)}$ ($L_2$)
            & \xa{61.6566666667}{0.0821921867063}{pc378}
            & \xa{57.82}{0.17029}{pc236} 
            & \xa{36.06}{0.30047}{pc293} 
            & \xa{52.454}{0.11253}{pc218b} 
            & \xa{38.3675}{0.3355}{pc341} 
            \\
            $D_\mathrm{AB}^{(\alpha, 0)}$
            & \xa{62.49}{0.0588784057755}{pc383}
            & \xa{58.59}{0.35922}{pc242} 
            & \xa{37.988}{0.30695}{pc294} 
            & \xa{52.57}{0.161245154966}{pc215b} 
            & \xa{39.0425}{0.5948}{pc340} 
            \\
            \midrule
        Teacher
            & \xb{66.01}{pc322}
            & \xa{59.66}{0.252586618806}{pc120} 
            & \xa{59.66}{0.252586618806}{pc120} 
            & \xa{52.208}{0.460104335993}{pc121} 
            & \xa{52.208}{0.460104335993}{pc121} 
            \\
        \bottomrule
        \end{tabular*}
    \end{center}
    \caption[Distillation loss]{
        \textbf{Distillation loss.} (Test accuracy, \%)
        We evaluate on five student-teacher tasks (see
        section~\ref{sec:copytasks}) the importance of temperature-adjusted KL
        divergence ($\alpha \not= 0$) as opposed to a squared $L_2$ loss
        ($\alpha = 0$).  Temperature becomes more important with higher degree
        of compression and less important at non-task layers (\textit{Top-}
        tasks).  Parameters of $D_\mathrm{AB}$ work well across situations,
        demonstrated here by the fact that we ended up using $\alpha = \frac
        14$ for all tasks. More values of $(\alpha, \beta)$ are explored in \cref{fig:ab-div-exp}.
    }
    \label{tab:copyloss}
\end{table}


The methods of creating a teacher signal compatible with the student are
evaluated in \cref{tab:distillation}. Feature subsets offer a significant
improvement over the use of auxiliary layers (see \cref{sec:feature-compatibility}). Warm-starting the distillation
with a single-task network offers faster convergence but modest benefits to the
end results.

\begin{table}
\begin{center}

\newcommand{\xa}[2]{$\num[round-mode=places,round-precision=1]{#1}$}
\newcommand{\tb}{\textbf}
    \begin{minipage}[t]{0.43\linewidth}
        \vspace{0pt}
        \begin{tabular*}{\linewidth}{l|r}
            \toprule
            Method                      & $\dotcup25\%$ \\ \midrule
            Auxiliary layer             & \xa{40.89}{pc334} \\
            Auxiliary layer (warm start)& \xa{41.69}{pc333} \\
            Feature subset              & \xa{43.56}{pc325b} \\  
            Feature subset (warm start) & \xa{43.70}{pc326b} \\
            \bottomrule
        \end{tabular*}
    \end{minipage}
    \qquad
    \begin{minipage}[t]{0.46\linewidth}
        \vspace{0pt}
        \includegraphics[width=\linewidth]{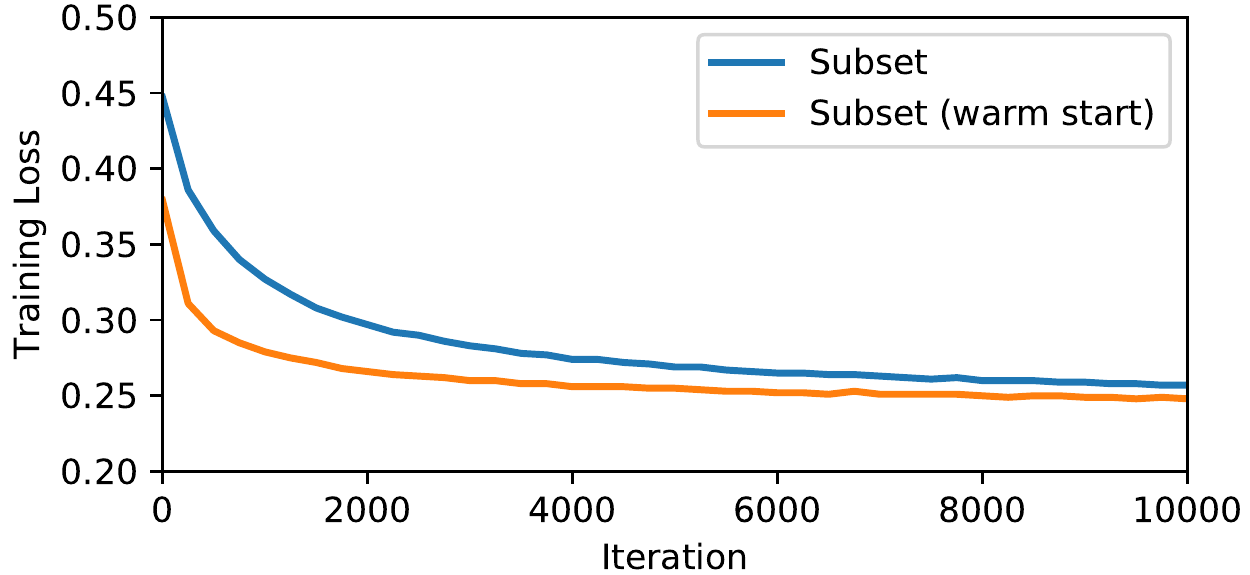}
    \end{minipage}
\end{center}
\vspace{-10pt}
\caption[Distillation methods]{
    \textbf{Distillation methods.} (Test accuracy, \%) Distillation works well, offering a dramatic
    improvement over a single model (32.4\%). Using an auxiliary layer to
    make student and teacher compatible for training works reasonably well.
    However, the simpler approach of taking a subset of the teacher's
    features work even better. In these experiments, there is little gain
    in warm-starting the network with a single network's weights. However, it
    does converge faster (\textit{right}) and benefits may be more prominent on larger network
    architectures that are harder to train well from scratch.
}
\label{tab:distillation}
\end{table}

\begin{figure}
    \centering
    \includegraphics[width=0.8\linewidth]{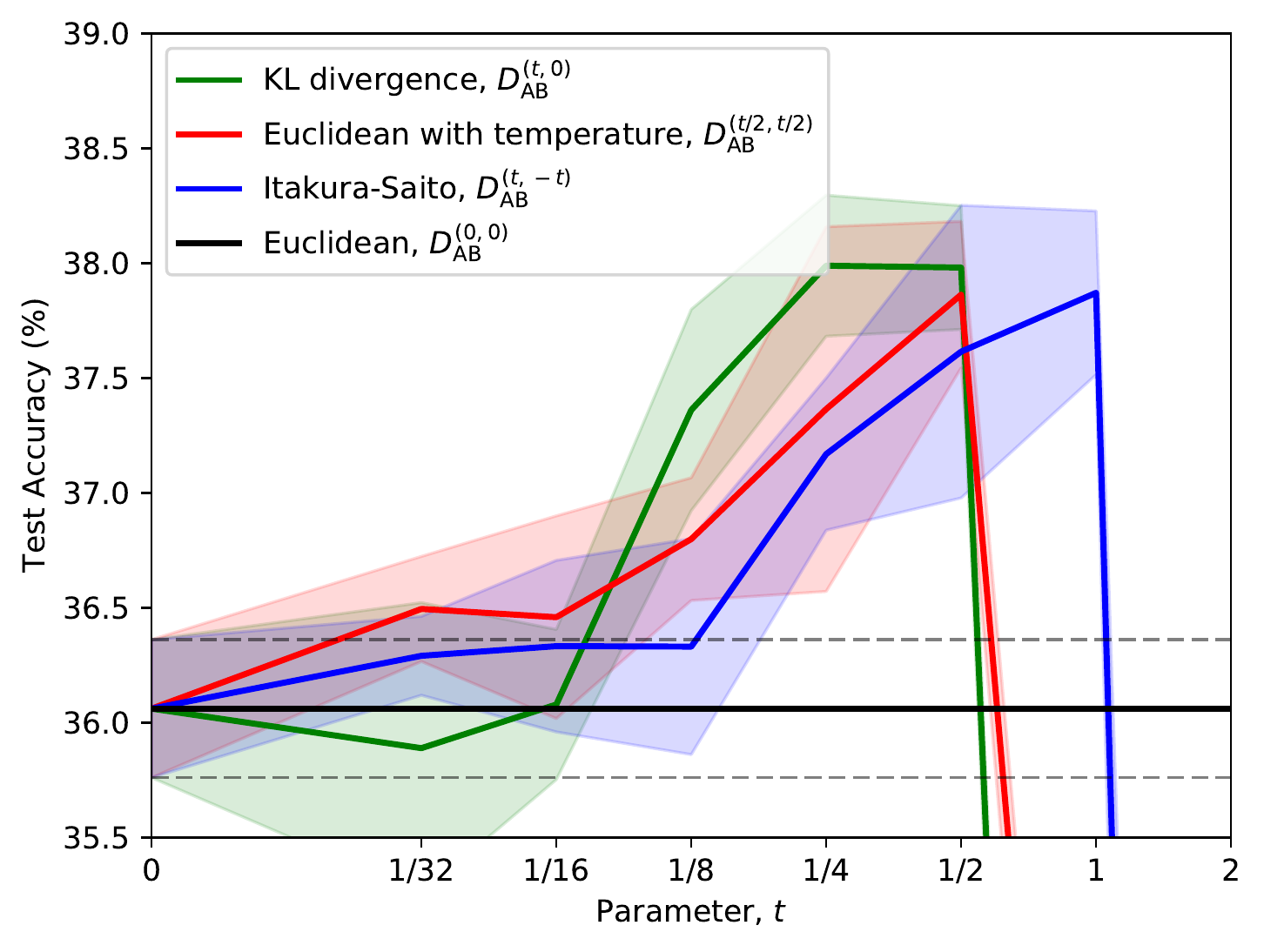}
    \caption[AB divergence as a distillation loss]{
        \textbf{AB divergence as a distillation loss.}
        We evaluate three types of AB divergences as a distillation loss on the
        \textit{Task-Compress} problem (see \cref{sec:copytasks}). The
        parameter $t$ moves along different straight lines on the $(\alpha,
        \beta)$ parameter space, with colors corresponding to
        \cref{fig:ab-div}. For KL divergence and Euclidean, $t$ corresponds to
        inverse temperature ($T^{-1}$). All values improve over $L_2$ loss ($t
        = 0$), but reach abrupt instabilities as they go too far away from the
        origin. There is no statistically significant winner among the top
        values ($38.0 \pm 0.3$, $37.9 \pm 0.3$, and $37.9 \pm 0.4$). However,
        KL divergence remains a natural choice, in addition to having a broader
        peak. Visualization of these losses can be seen in \cref{fig:ab-div-shapes}
    }
    \label{fig:ab-div-exp}
\end{figure}

Furthermore, we evaluate AB divergences where $\beta \not= 0$ on the
\textit{Task-Compress} problem. The results are compiled in \cref{fig:ab-div-exp} and
show to our surprise that several variants improve over the baseline with
results statistically indistinguishable from temperature-adjusted KL
divergence. This includes the symmetric $D_\mathrm{AB}^{(t, t)}$ and the scale
invariant $D_\mathrm{AB}^{(t, -t)}$. Our recommendation is to use the KL
divergence line ($D_\mathrm{AB}^{(t, 0)}$), however the AB divergence does
offer fertile ground for future experimentation.

\subsection{Multi-Task Representation Learning}

For multi-task representation learning, we split the CIFAR-100 training
data into 40,000 samples for pretraining and 10,000 for fine-tuning. The
40,000, are split further into a wide variety of different tasks that 
are designated as unrelated in the multi-task regime. Each of 4 tasks can for instance
randomly sample 50\% of the data, which we denote $\cup50\%$, where $\cup$ denotes
that each sample is independent. Alternatively, if the data is split into 4
pieces, we denote it $\dotcup25\%$, where the added dot indicates the tasks for a
\textit{disjoint} partition. The splits can be done by random samples or by
class. For instance, 4 tasks split using $\cup50\%$ by class, means that each
task gets a random subset of 50\% of the classes and all the samples that go
with it. This particular situation may result in classes lacking any
pretraining representation at all.

The generated datasets are now considered separate tasks and will serve as
analogs to different self-supervision methods. A representation is now learned
using any of method outlined in this chapter. For instance, in the case of model distillation,
each task is trained separately, the task layer is discarded, and model distillation is performed at the top layer.
For joint training, all tasks are attached to the top layer of the base network
and trained together. Regardless of the method used, it will produce a trained
base network without a task layer.
A task layer is then attached to this base network and trained on the 10,000 held out for fine-tuning.
In order to emphasize the representation learning, we do not use
end-to-end fine-tuning and only the new task layer is allowed to update.

Results are presented in \cref{tab:cifar}. First, we note that distillation
does quite well, offering substantial improvements over the baseline of using
only a single task for representation learning. It is often on par with
split-weights training and sometimes does better, even though it is an online
method. Split-activations, the only other method that enjoys the benefits of
an offline method, fails to train at all for 10 tasks and is strongest for two tasks (since this
eliminates the least amount of parameters from pretraining).
However, we see no benefits of this model since at two tasks the distillation
is still superior with a 7-point margin. In fact, distillation is as good as
the best online method when combining only two tasks.

Joint training consistently does well, although our fears of adaptive methods are
confirmed by consistently under-performing compared to SGD by about 2 points. It may seem
that using 4 tasks of 50\% overlapping classes should be better
than 4 tasks with only 25\% disjoint classes. However, there is only a slight trend
in this direction, which may be explained by the fact that in the overlapping case,
a few classes fail to get representation at all.

The right-most column is labeled \textit{Mixed}. For these four tasks, we
use four distinct losses for the task: multi-class softmax KL divergence,
squared $L_2$, multi-label sigmoid KL divergence, and multi-label hinge loss.
The softmax and sigmoid both train a decent model alone, but $L_2$ and
multi-label hinge loss struggle and bring the single task mean down. Since they
struggled individually, we did not evaluate this for offline methods; it would
not be fair to give them failed models that they are not allowed to change.
Online methods on the other hand can benefit from joint training, in that maybe the failed
losses simply had a hard time getting close to a good solution or were too
prone to diverge into failed states.  Training them together seems to alleviate
this, since most methods are able to train reasonable models. First, we see that
Adam does better than SGD. It is not entirely clear why, since Adam does not
individually normalize the tasks. Even more surprisingly is that using separate Adam optimizers (``Multi-Adam'')
as described in \cref{sec:sep-opt} does \textit{worse} than Adam. Perhaps it is boosting poor gradients from the
struggling methods too much. NormSGD addresses and resolves the issues of SGD
to the surprising extent that it becomes on par with results using the same loss.

Are there still settings where Multi-Adam may be beneficial over Adam? Yes, if
we do a more contrived mixed loss, by simply using softmax KL divergence for
all, but with each loss multiplied by a random factor. In this setting, Multi-Adam
and NormSGD are equivalent to their same-loss counter-parts, while Adam and
SGD do take a performance hit. Even though there are situations where
Multi-Adam does solve a problem that Adam does not, we still prefer NormSGD due to
the trend toward better results for representation learning.

\Cref{tab:roundrobin} shows that joint training with NormSGD can
be implemented using round-robin scheduling. This approach scales better in $M$ in terms
of memory use. If tasks are sufficiently different, the time costs of a forward-backward pass
may vary. In joint training, this means that all other tasks must wait for the
bottleneck task; an issue that is avoided using round-robin scheduling.




\begin{table}
\begin{center}

\newcommand{\xa}[2]{$\num[round-mode=places,round-precision=1]{#1}$}
\newcommand{\xb}[3]{$\mathbf{#3}$}
\newcommand{\tb}{\textbf}
    \begin{tabular*}{\linewidth}{l|@{\extracolsep{\fill}}rrrrrr}
        \toprule
        Tasks
            & 4
            & 4
            & 2
            & 4
            & 10
            & 4
            \\
        Task loss
            & Same
            & Same
            & Same
            & Same
            & Same
            & Mixed
            \\
        Classes
            & 
            & $\cup50\%$
            & $\dotcup50\%$
            & $\dotcup25\%$
            & $\dotcup10\%$
            & $\dotcup25\%$
            \\
        Sample size
            & $\dotcup25\%$
            & 
            & 
            & 
            & 
            & 
            \\
            \midrule
        Single task (mean)
            & \xa{38.4025}{pc180[,b,c,d]}
            & \xa{42.0575}{pc125[,b,c,d]}
            & \xa{41.62}{pc186[,b]}
            & \xa{32.3825}{pc124[,b,c,d]}
            & \xa{24.416}{pc364,res-*}
            & \xa{16.8275}{pc124,pc159f,pc160d,pc161e}
            \\
        Single task (max)
            & \xa{38.84}{pc180[,b,c,d]}
            & \xa{43.03}{pc125[,b,c,d]}
            & \xa{42.09}{pc186[,b]}
            & \xa{33.29}{pc124[,b,c,d]}
            & \xa{26.54}{pc364,res-7}
            & \xa{32.23}{pc124,pc159f,pc160d,pc161e}
            \\
            \midrule
        Distillation (subset)
            & \xa{43.95}{pc330}
            & \xa{46.65}{pc331}
            & \xa{54.24}{pc329}
            & \xa{47.22}{pc326}
            & \xa{42.16}{pc357}
            & \dash 
            \\
        Split-activations
            & \xa{29.37}{pc172b}
            & \xa{39.84}{pc167}
            & \xa{47.26}{pc178b}
            & \xa{32.98}{pc151b}
            & \xa{1.00}{pc358}
            & \dash 
            \\ 
        Split-weights
            & \xa{46.96}{pc173}
            & \xa{50.30}{pc324}
            & \xa{52.29}{pc179}
            & \xa{45.98}{pc152}
            & \xa{25.52}{pc359d}
            & \dash 
            \\
        Joint (Adam)
            & \xa{52.14}{pc168}
            & \xa{49.22}{pc162}
            & \xa{51.74}{pc174} 
            & \xa{49.34}{pc148b}
            & \xa{45.18}{pc360}
            & \xa{46.61}{pc182}
            \\
        Joint (Multi-Adam)
            & \xa{52.94}{pc169}
            & \xa{50.69}{pc163}
            & \xa{52.76}{pc175}
            & \xa{49.50}{pc150}
            & \xa{44.34}{pc361}
            & \xa{44.32}{pc355c}
            \\
        Joint (SGD)
            & \xa{54.07}{pc170f}
            & \xa{52.16}{pc164}
            & \xa{53.99}{pc176}
            & \xa{51.27}{pc122c}
            & \xa{46.48}{pc362}
            & \xa{38.49}{pc146} 
            \\
        Joint (NormSGD)
            & \xa{54.44}{pc354}
            & \xa{51.43}{pc343e}
            & \xa{54.06}{pc337d}
            & \xa{51.54}{pc344g}
            & \xa{47.93}{pc363}
            & \xb{51.05}{pc353}{51.1}
            \\
            \midrule
        Pretrained on full dataset
            & \xa{55.46}{pc121}
            & \xa{55.46}{pc121}
            & \xa{55.46}{pc121}
            & \xa{55.46}{pc121}
            & \xa{55.46}{pc121}
            & \xa{55.46}{pc121}
            \\ 
        \bottomrule
    \end{tabular*}
\end{center}
\caption[Multi-task learning]{
    \textbf{Multi-task learning.} (Test accuracy, \%)
    Each task is a subset of CIFAR-100, either by class or by sample size. The
    subset can be a disjoint partition ($\dotcup$) or independently sampled per
    task allowing overlaps ($\cup$).
    The \textit{single task} results should be considered a baseline; a method
    fails to have practical use if it falls below this value.
    On the other end, we do not expect methods to exceed pretraining on
    the entire dataset, so this can be considered an upper
    bound. Two of the mixed losses did not train well individually, causing the
    single task mean to be low. It is unfair to evaluate them for offline
    methods. However, online methods have the potential to prop up poor choices
    of losses, which is exactly what happened with NormSGD (shown in bold) and
    to a lesser degree with Adam. Multi-Adam did worse than Adam, ruling it out as a
    solution to uneven losses. Joint training can also be substituted by round-robin scheduling
    (see \cref{tab:roundrobin}).
%
%
}
\label{tab:cifar}
\end{table}

\begin{table}
\begin{center}

\newcommand{\xa}[2]{$\num[round-mode=places,round-precision=1]{#1}$}
\newcommand{\xb}[3]{$\mathbf{#3}$}
\newcommand{\tb}{\textbf}
    \begin{tabular*}{\linewidth}{l|rr|rrrrrr}
        \toprule
            & $Q$
            & $I$
            & 4 Same
            & 4 Same
            & 2 Same
            & 4 Same
            & 10 Same
            & 4 Mixed
            \\
        Method
            & \multicolumn{2}{c|}{\footnotesize($\times 10^3$)}
            & $N\dotcup25\%$
            & $C{\cup}50\%$
            & $C\dotcup50\%$
            & $C\dotcup25\%$
            & $C\dotcup10\%$
            & $C\dotcup25\%$
        \\
            \midrule
        Joint
            & \ndash
            & 200
            & \xa{54.44}{pc354}
            & \xa{51.43}{pc343e}
            & \xa{54.06}{pc337d}
            & \xa{51.54}{pc344g}
            & \xa{47.93}{pc363}
            & \xa{51.05}{pc353}
            \\
        Round-robin
            & 1
            & 200
            & \xa{54.35}{pc365}
            & \xa{52.70}{pc366}
            & \xa{53.63}{pc367}
            & \xa{52.20}{pc368}
            & \xa{47.67}{pc369}
            & \xa{51.49}{pc370}
            \\
        Round-robin
            & 10
            & 200
            & \xa{54.19}{pc365b}
            & \xa{51.32}{pc366b}
            & \xa{52.20}{pc367b}
            & \xa{51.04}{pc368b}
            & \xa{48.19}{pc369b}
            & \xa{51.19}{pc370b}
            \\
        \bottomrule
        \multicolumn{9}{l}{$Q$: Quantum (time between task switch), $I$: Iterations, $M$: Tasks}
    \end{tabular*}
\end{center}
\caption[Round-robin task scheduling]{
    \textbf{Round-robin task scheduling.} (Test accuracy, \%)
    Keeping only a single task active at any time is an
    alternative to joint training.
    This often works equally well as joint training, even when the quantum is
    high. For instance, when the quantum is $Q=\num[group-separator={,}]{10000}$, the total iterations
    $I=\num[group-separator={,}]{200000}$ (learning rate drops at $\num[group-separator={,}]{100000}$ and $\num[group-separator={,}]{150000}$) and tasks $M=10$,
    then each task only gets a single segment at the original learning rate and
    only half at once-dropped and the other half at twice-dropped
    learning rate. Despite such high quantum, the last row has almost no
    effect on the results. The only task that suffers a bit is when $M=2$.
    All methods use NormSGD.
}
\label{tab:roundrobin}
\end{table}

\section{Multi-Proxy Self-Supervision}
\label{sec:multiproxy-selfsup}

We finally turn our attention to combining multiple self-supervision methods.
Since joint training with NormSGD shows the most promise for multi-task
representation learning, we now apply it to multi-proxy self-supervised
representation learning. We use three self-supervised tasks outlined
below. Our implementation does not use round-robin scheduling, although it
would be an appropriate strategy for a future implementation, especially if
$M$ is increased further.

Following \cref{chp:unsupervised,chp:selfsup}, we evaluate the representation
learned from multi-proxy training on classification and semantic segemntation
from the PASCAL VOC challenges (see \cref{sec:pascal}).

\subsection{Proxy Tasks}
\label{sec:proxy-tasks}

A brief summary of the different self-supervision methods that are combined.

\textbf{Colorization.} The colorization method is described in
\cref{chp:colorization,chp:selfsup}. We use a re-implementation in
TensorFlow (as opposed to Caffe used in earlier chapters), and this might be a reason why it tests 
lower than our Caffe implementation with the same amount of training.
When training jointly, the
first convolutional layer must have three-channel filters, so we instead feed it
an RGB-encoded grayscale version, where the same intensity image is replaced
three times.

\textbf{Jigsaw.} The original jigsaw method~\cite{noroozi2016jigsaw} uses a
modified version of AlexNet and does not pretrain \texttt{fc6} and \texttt{fc7}
at all. This is to allow each patch to be rather small (128-by-128) so
that a 3-by-3 puzzle can be used. We want the jigsaw solver to contribute to
all layers, so we opted for a 2-by-2 puzzle that are fed 500-by-500 images,
with each puzzle piece the canonical input size at 227-by-227 (the extra pixels
are necessary for wiggle room between the pieces). At 2-by-2, we use all 24
possible permutations of the puzzle as the output of the prediction task.
We use ImageNet~\cite{imagenet} and Places205~\cite{zhou2014learning} as a
combined source of unlabeled images for both Colorization and Jigsaw.

\textbf{Video.} For video, we use a novel approach developed for the purpose of
this experiment. It is inspired by Pathak~\etal~\cite{pathak2016move}, however
we take a simpler approach of leveraging optical flow for ease of
implementation.\footnote{Simply using their pre-processed data would be easier,
but it had not been publicly available yet.} We compute Farneb{\"a}ck optical
flow~\cite{farneback2003two} and associate it with a single frame. We sample
random locations uniformly from the image. Next, for each location we sample another
location anchored to the first by two independent Gaussian distributions. For these pairs of locations,
the relative optical flow is computed and its amplitude is placed into one of
32 bins.  Sparse hypercolumns (see \cref{sec:sparse-hypercolumns}) are
extracted and the paired hypercolumns are concatenated and fed to a prediction
layer that outputs a distribution over the 32 bins of relative flow amplitude.
Note, since we are only considering relative flow, this method is invariant to camera translation and therefore to some extent invariant to camera rotation.
We use the Hollywood2~\cite{marszalek2009actions} dataset as source of unlabeled
video input. Since this is a small dataset and we do not plan to use any of the
labels, we use both the training and testing data. This consists of 2517 clips
of median length 10 seconds.

\subsection{Method}

The three methods share the same base network. We implement this by keeping
separate instances of this network with tied weights. This is necessary since
we feed the networks differently sized inputs from different data sources. If
two methods use the same data source and size, they can use a single network
instance with two sets of task-specific appendages. Keeping separate network
instances also allows placing them on different GPUs for parallel processing, with
a synchronization phase at the end of each iteration.

The base network is AlexNet and we use NormSGD (\cref{sec:norm-sgd}) for
training. Colorization and Video both use complete hypercolumns, extracted from
all layers, and a fully connected layer with 1024 channels prior to the task
layer.  We train for 2M iterations, with factor 10 drops of the learning rate at 1M and 1.5M.  Batch
sizes are 8 when tasks are trained individually and 4 for each task when
trained jointly. This means that ``C+J'' sees the same amount of data, but the
gradients are likely noisier. To avoid making the gradients even noisier, we did not
drop this further for ``C+J+V,'' even though this technically means more data
was fed during training. The code for jointly training models, with
additional implementation details for all three methods, is made publicly
available.\footnote{\url{https://github.com/gustavla/self-supervision}}


\begin{figure}
\begin{minipage}[t]{0.24\linewidth}
  \vspace{0pt}
  \centering
  \includegraphics[width=.9\linewidth]{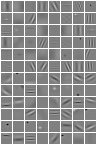}\\
    C (Caffe)
    \\
  \vspace{0.4cm}
  \includegraphics[width=.9\linewidth]{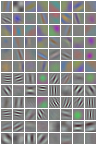}\\
    ImageNet
\end{minipage}%
\begin{minipage}[t]{0.24\linewidth}
  \vspace{0pt}
  \centering
  \includegraphics[width=.9\linewidth]{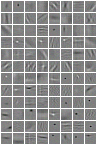}\\
    C\phantom{q} \\
  \vspace{0.4cm}
  \includegraphics[width=.9\linewidth]{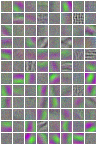}\\
    C+J (SGD)
\end{minipage}
\begin{minipage}[t]{0.24\linewidth}
  \vspace{0pt}
  \centering
  \includegraphics[width=.9\linewidth]{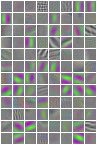}\\
    J\phantom{q} \\
  \vspace{0.4cm}
  \includegraphics[width=.9\linewidth]{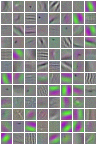}\\
    C+J (NormSGD)
\end{minipage}
\begin{minipage}[t]{0.24\linewidth}
  \vspace{0pt}
  \centering
  \includegraphics[width=.9\linewidth]{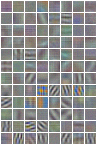}\\
    V\phantom{q} \\
  \vspace{0.4cm}
  \includegraphics[width=.9\linewidth]{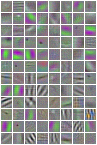}\\
    C+J+V (NormSGD)
\end{minipage}
\vspace{0.4cm}
\caption[First convolutional filters]{\textbf{First convolutional filters.}
    The first filters of AlexNet are large enough for meaningful visual
    inspection. We note that all three self-supervised methods have very
    distinct characteristics. 
    Colorization (C) is grayscale and has Gabor-like
    filters as well as blobs.  We also show our Caffe implementation from
    \cref{chp:selfsup}, which was trained longer and is therefore a bit
    smoother.
    Jigsaw (J) oddly has a lot of green-purple edges, although this is
    consistent with the original paper~\cite{noroozi2016jigsaw}.  Video (V) has
    more color
    with broad edges and many high-frequency Fourier filters (without tapering
    off as Gabor filters).  Our initial attempt at combining two methods (C+J)
    resulting in one dominating the other and filters indistinguishable from J
    alone (except a bit grainier here, due to shorter training).  This was
    resolved by NormSGD and we get a more even mix of green-purple edges
    (from J) and blobs/Gabor filters (from C). With three proxy tasks we see characteristics of all 
    methods, suggesting a fairly even focus on all three tasks. The diversity
    of filters for ``C+J+V'' makes it the closest to the filters trained on ImageNet.
    Each set of filters is individually stretched out in scale, in a symmetric
    way such that middle-gray represents 0.
}
\label{fig:conv-filters}
\end{figure}

\subsection{Results}

Results are seen in \cref{tab:multiproxy}. We try combining both two tasks
(Colorization and Jigsaw), and three tasks (Colorization, Jigsaw, and Video).
However, the downstream performance results neither increase nor decrease as a
result of multi-proxy learning. 

Training Colorization and Jigsaw together with SGD, we observe that Jigsaw
gradients dominate the Colorization gradients. As a result, the filters of the
first convolutional layer were similar to the filters of Jigsaw trained alone.
Using NormSGD, although it does not result in improved downstream results, it
produces filters that have the appearance of being more evenly mixed between
both tasks (see \cref{fig:conv-filters}).

\begin{figure}
    \centering
    \includegraphics[width=1.0\linewidth]{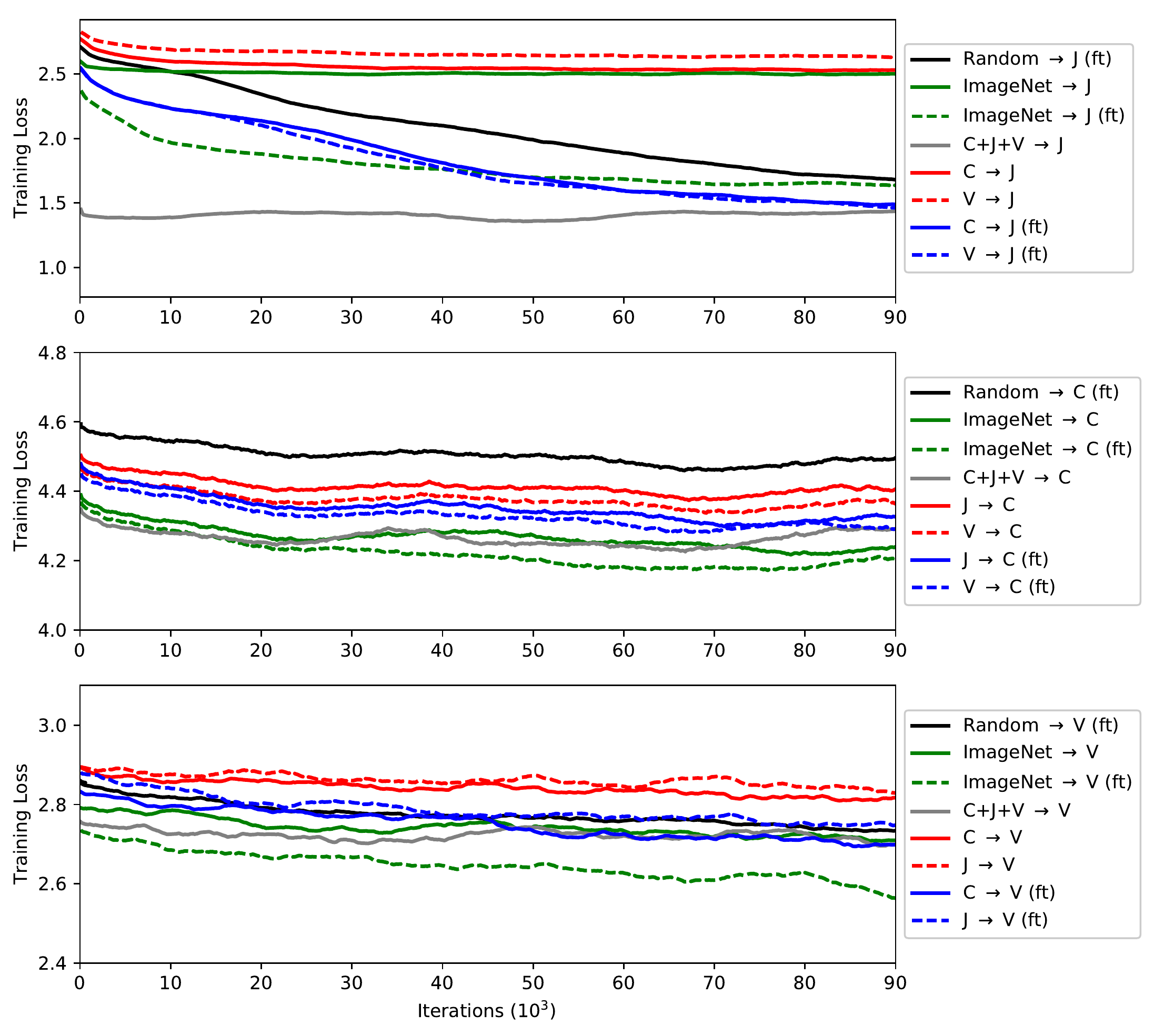}
    \caption[Inter-proxy training loss]{
        \textbf{Inter-proxy training loss.}
        We plot the initial learning curves for the tasks C (Colorization), J
        (Jigsaw), and V (Video). The notation ``ImageNet $\rightarrow$ C
        (ft)'' means that Colorization was trained after initializing from an
        ImageNet pretrained network, with ``ft'' indicating end-to-end
        fine-tuning as opposed to task-specific fine-tuning only. The first
        observation we make is that initializing with other self-supervised
        tasks is almost always beneficial, with Video the only exception
        where it did not help. Moreover, Jigsaw, which is the only one without
        hypercolumns, does not do well without fine-tuning, even when
        initializing with ImageNet. We also try initializing with ``C+J+V'' and
        we see that the network as expected has learned all three tasks to some
        degree. Finally, it is hard to get clear signals from C and V, since
        they are difficult tasks and the dynamic range of the loss is small
        even over long training periods. We improved this by turning off
        randomizing samples for Colorization, which is why all lines share
        trends.
    }
    \label{fig:jcv-loss}
\end{figure}

To further investigate what might be happening, we try pretraining on one proxy
task to see if it helps another proxy task. These results are plotted in
\cref{fig:jcv-loss} and indicate that tasks generally make the other
tasks converge faster. However, we also see that initializing with a jointly trained
network makes all three tasks quickly learn the task much better than using
either of the other self-supervised tasks. In summary, we have trained a more
general network by training jointly and it is conceivable that there exist tasks
where this would be expressed. However, our tests on classification and
semantic segmentation do not give expression to these potential benefits.

\section{Conclusion}

In this chapter we have considered a wide range of methods for combining proxy
tasks that learn a single representation. This goal is different from
multi-task learning, since we do not really care how we perform on the
throw-away proxy tasks. We consider both offline methods (train
separately and combine) and online methods (train together). For offline methods,
task-agnostic distillation by taking feature subsets at the top layer is a
competitive method. In developing this approach, we also develop a new
distillation loss that can be used for at task layers as well as task-agnostic
layers. For online methods, we show that joint training is effective, but it 
is important to be aware of the possibility of uneven gradients. We address
this problem by proposing NormSGD, which is invariant to loss scale and thus 
suited when combining disparate losses. If the memory overhead of joint
training becomes a problem, we show that training a single task at a time using
round-robin scheduling is also fine for representation learning.

Despite promising results on test benchmarks, we show no improved results by
jointly training self-supervision methods. This is a curious negative result
that could suggest that there is not much to gain from the apparent diversity
of the different self-supervision methods. However, it could also mean that
harnessing the diversity is more difficult than we envisioned, and the methods
outlined here is only a first step in achieving this long-term goal. If this is
the case, we hope that our work can offer a good starting point.

\begin{table}
\begin{center}

\newcommand{\xa}[2]{$\num[round-mode=places,round-precision=1]{#1}$}
    \begin{tabular*}{\linewidth}{l@{\qquad\qquad\quad\,\,\,}|@{\qquad\quad}r@{\qquad\quad}r}
        \toprule
        Method           & Classification (\%mAP) & Segmentation (\%mIU) \\ \midrule
        No Pretraining   & \xa{46.15}{a231}       & \xa{23.5}{a228}      \\ \midrule
        C (Colorization) & \xa{61.13}{mu124}      & \xa{36.34}{a258}     \\
        J (Jigsaw)       & \xa{60.02}{mu101}      & \xa{33.72}{a252}     \\
        V (Video)        & \xa{60.64}{mu98}       & \xa{35.94}{a253}     \\ \midrule
        C+J              & \xa{59.16}{mu100}      & \xa{35.52}{a254}     \\
        C+J+V            & \xa{60.23}{mu107}      & \xa{36.08}{a259}     \\
        \bottomrule
    \end{tabular*}
\end{center}
\caption[Multi-proxy self-supervision]{
    \textbf{Multi-proxy self-supervision.} (AlexNet)
    We implement three entirely separate self-supervision methods (see \cref{sec:proxy-tasks}) with similar
    pretraining benefits. Two or three methods are then trained jointly using NormSGD. 
    Note, colorization results for AlexNet are even better in
    \cref{tab:comparison} due to longer training and possibly subtle implementation
    details.
}
\label{tab:multiproxy}
\end{table}

\chapter*{Conclusion}
\addcontentsline{toc}{chapter}{Conclusion}

We have demonstrated two unsupervised methods for representation learning,
highlighting limitations of traditional methods of unsupervised learning. The
first method failed due to the limitations of greedy training, which makes it
difficult to leverage the model complexity of deep networks. The second method
overcomes this issue through end-to-end training. However, it still falls short
since it uses an unsupervised loss based on reconstruction error that fails to
emphasize high-level semantics.

Instead, we use a supervised loss on a task manufactured from unlabeled data
(self-supervised). We propose to do this with automatic colorization, which we
separately develop into a graphics application with state-of-the-art results.
Colorization turns out to be an
excellent vehicle for representation learning. The primary reason for its success is
that the task requires high-level visual understanding, unlike the
reconstruction error used in traditional unsupervised methods. We show that
this new proxy task leads to significant progress in closing the gap between
supervised and unsupervised pretraining.

We hope that this gap can be further closed by combining multiple
self-supervision techniques through model aggregation or joint training.
However, we have shown that this is nontrivial as is evidenced by our extensive
investigation showing no improvement over single-model results.
Further work in this direction may still turn out to be worthwhile, in which
case we hope that our progress offers a foundation for future work.



\appendix

\chapter{Permutation EM} \label{apx:permutation-em}

\section{Introduction}

We present further details and derivations of Permutation EM described in~\cref{sec:permutation-em}.

\subsection{Cyclic Groups}
Even though the algorithm is general to any set of permutations, it is particularly suited for products of cyclic groups. Consider for instance two cyclic groups of image transformations: one group of rotations $(G_\rot, \circ)$ of size $R$ and one group of edge polarities $(G_\pol, \circ)$ of size $T = 2$. The symbol $\circ$ denotes function composition of the transformations. Either of these two cyclic groups can be used as a latent transformation, as well as their direct product $G = G_\rot \times G_\pol$. Let the transformations be denoted $(g_\rot, g_\pol) \in G$ and $g = g_\rot \circ g_\pol$. Let $g_p$ denote the transformation $p$ in $g$.
For each sample, a complete sample can be constructed by concatenating all transformations applied to the sample:
\begin{equation} \label{eq:training-X}
    \vx = [g_1(x), g_2(x), \dots, g_P(x)]\ \\
\end{equation}
Each cyclic group is generated by a single transformation, such as $\Phi_\mathrm{rot}(x)$ (rotate $\frac{360}R$ degrees) and $\Phi_\mathrm{pol}(x)$ (flip intensity).

\subsection{Permutations}
Let $\sigma_N = (0\, 1\, \dots\, N-1)$ be a cyclic permutation.\footnote{This is written in cycle notation and means that $\sigma_N(n) = (n + 1) \bmod N$.} We create the permutation matrix by letting $\sigma_R \times \sigma_T$ generate pairs of indices, where one cycles through the rotations and one through the polarities. Finally, we create a bijection between these pairs into a flattened index space, corresponding to the order in which we did the transformations. 

As an example, let $R = 3$ and $T = 2$, and let the flattened index be $i = Rt + r$, so that indices $\{0, 1, 2\}$ are all rotations of one polarity and $\{3, 4, 5\}$ are all rotations of the other. The matrix becomes
\begin{equation} \label{eq:A-ex}
    A = \begin{bmatrix}
        \sigma_R & \sigma_R + R \\
        \sigma_R + R & \sigma_R \\
    \end{bmatrix} = \begin{pmatrix}
        0 & 1 & 2 & 3 & 4 & 5 \\
        1 & 2 & 0 & 4 & 5 & 3 \\
        2 & 0 & 1 & 5 & 3 & 4 \\
        3 & 4 & 5 & 0 & 1 & 2 \\
        4 & 5 & 3 & 1 & 2 & 0 \\
        5 & 3 & 4 & 2 & 0 & 1 \\
    \end{pmatrix}, 
\end{equation}
where we allow $\sigma_R$ to be treated as an $R \times R$ matrix. 
From here on, we will not mention $R$ and $T$ anymore since they are abstracted away through the total number of permutations $P = RT$ and the permutation matrix $A$.

\subsection{Notation}
If $F$ is the size of a set, $\mathcal{F} = \{0, \dots, F - 1\}$ is the set of indices and $f \in \mathcal{F}$ is an index. Zero-based indexing will be particularly useful for our cyclic groups. I will follow this notation for $P$ (permutations), $D$ (features, both spatial and incoming channels), $F$ (canonical parts) and $N$ (samples).
Let $X$ denote by context either a random vector corresponding to a complete sample ($\vx$
from \eqref{eq:training-X}) or a sample. Let $\vX = (X^{(1)}, \dots, X^{(N)})$ denote the entire training dataset. Similar notation follows for the latent space using $Z$ and $W$.
To improve readability, the $(i,j)$ index of the permutation matrix $A$ will be denoted as $A(i,j)$, to avoid small fonts. 


\section{Model}
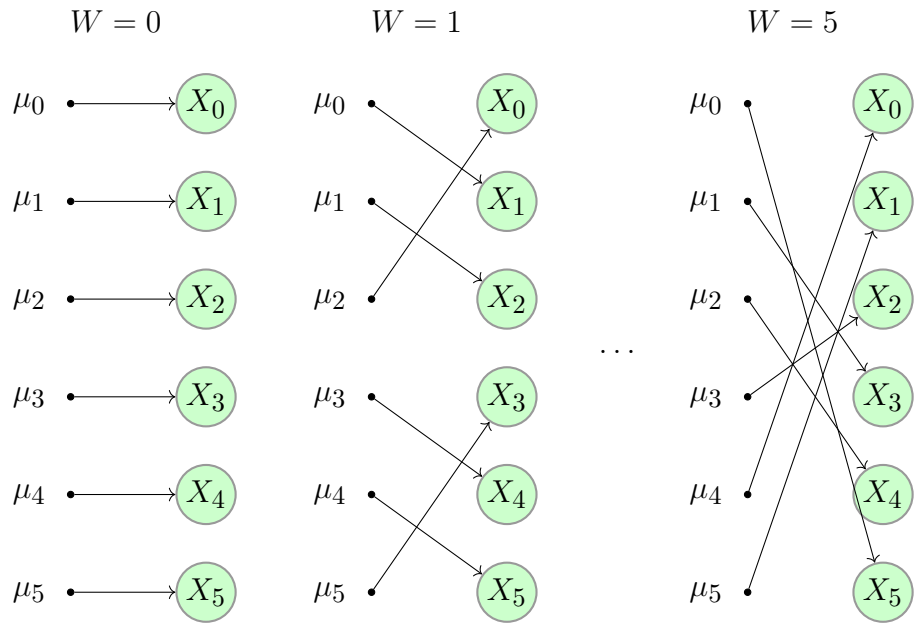
\begin{figure}
\begin{center}
\begin{tikzpicture}[
  font=\sffamily,
  every matrix/.style={ampersand replacement=\&,column sep=2cm,row sep=2cm},
  param/.style={draw,fill=black,circle,inner sep=0.03cm},
  param caption/.style={node distance=0.55cm},
  var/.style={draw=black!40,thick,circle,fill=green!20,inner sep=.05cm},
  to/.style={->,>=stealth',shorten >=1pt,semithick,font=\sffamily\footnotesize},
  every node/.style={align=center},
  node distance=1.2cm]

  \node  (pol1) at (1, 7.6) {$W=0$};
  \node[param] (0mu0)  at (0.4, 6.5) {};
  \node[param] (0mu1)  at (0.4, 5.2) {};
  \node[param] (0mu2)  at (0.4, 3.9) {};
  \node[param] (0mu3)  at (0.4, 2.6) {};
  \node[param] (0mu4)  at (0.4, 1.3) {};
  \node[param] (0mu5)  at (0.4, 0.0) {};
  \node[var] (0X0) at (2.2, 6.5) {$X_0$};
  \node[var] (0X1) at (2.2, 5.2) {$X_1$};
  \node[var] (0X2) at (2.2, 3.9) {$X_2$};
  \node[var] (0X3) at (2.2, 2.6) {$X_3$};
  \node[var] (0X4) at (2.2, 1.3) {$X_4$};
  \node[var] (0X5) at (2.2, 0.0) {$X_5$};

  \node[param caption] [left of=0mu0] {$\mu_0$};
  \node[param caption] [left of=0mu1] {$\mu_1$};
  \node[param caption] [left of=0mu2] {$\mu_2$};
  \node[param caption] [left of=0mu3] {$\mu_3$};
  \node[param caption] [left of=0mu4] {$\mu_4$};
  \node[param caption] [left of=0mu5] {$\mu_5$};

  \draw[->] (0mu0) -- (0X0);
  \draw[->] (0mu1) -- (0X1);
  \draw[->] (0mu2) -- (0X2);
  \draw[->] (0mu3) -- (0X3);
  \draw[->] (0mu4) -- (0X4);
  \draw[->] (0mu5) -- (0X5);

  \node  (pol5) at (5, 7.6) {$W=1$};
  \node[param] (1mu0)  at (4.4, 6.5) {};
  \node[param] (1mu1)  at (4.4, 5.2) {};
  \node[param] (1mu2)  at (4.4, 3.9) {};
  \node[param] (1mu3)  at (4.4, 2.6) {};
  \node[param] (1mu4)  at (4.4, 1.3) {};
  \node[param] (1mu5)  at (4.4, 0.0) {};
  \node[var] (1X0) at (6.2, 6.5) {$X_0$};
  \node[var] (1X1) at (6.2, 5.2) {$X_1$};
  \node[var] (1X2) at (6.2, 3.9) {$X_2$};
  \node[var] (1X3) at (6.2, 2.6) {$X_3$};
  \node[var] (1X4) at (6.2, 1.3) {$X_4$};
  \node[var] (1X5) at (6.2, 0.0) {$X_5$};

  \node[param caption] [left of=1mu0] {$\mu_0$};
  \node[param caption] [left of=1mu1] {$\mu_1$};
  \node[param caption] [left of=1mu2] {$\mu_2$};
  \node[param caption] [left of=1mu3] {$\mu_3$};
  \node[param caption] [left of=1mu4] {$\mu_4$};
  \node[param caption] [left of=1mu5] {$\mu_5$};

  \draw[->] (1mu0) -- (1X1);
  \draw[->] (1mu1) -- (1X2);
  \draw[->] (1mu2) -- (1X0);
  \draw[->] (1mu3) -- (1X4);
  \draw[->] (1mu4) -- (1X5);
  \draw[->] (1mu5) -- (1X3);

  \node  (text) at (7.7, 3.2) {$\dots$};

  \node  (pol0) at (10, 7.6) {$W=5$};
  \node[param] (5mu0)  at (9.4, 6.5) {};
  \node[param] (5mu1)  at (9.4, 5.2) {};
  \node[param] (5mu2)  at (9.4, 3.9) {};
  \node[param] (5mu3)  at (9.4, 2.6) {};
  \node[param] (5mu4)  at (9.4, 1.3) {};
  \node[param] (5mu5)  at (9.4, 0.0) {};
  \node[var] (5X0) at (11.2, 6.5) {$X_0$};
  \node[var] (5X1) at (11.2, 5.2) {$X_1$};
  \node[var] (5X2) at (11.2, 3.9) {$X_2$};
  \node[var] (5X3) at (11.2, 2.6) {$X_3$};
  \node[var] (5X4) at (11.2, 1.3) {$X_4$};
  \node[var] (5X5) at (11.2, 0.0) {$X_5$};

  \node[param caption] [left of=5mu0] {$\mu_0$};
  \node[param caption] [left of=5mu1] {$\mu_1$};
  \node[param caption] [left of=5mu2] {$\mu_2$};
  \node[param caption] [left of=5mu3] {$\mu_3$};
  \node[param caption] [left of=5mu4] {$\mu_4$};
  \node[param caption] [left of=5mu5] {$\mu_5$};

  \draw[->] (5mu0) -- (5X5);
  \draw[->] (5mu1) -- (5X3);
  \draw[->] (5mu2) -- (5X4);
  \draw[->] (5mu3) -- (5X2);
  \draw[->] (5mu4) -- (5X0);
  \draw[->] (5mu5) -- (5X1);

\end{tikzpicture}
\end{center}
\caption[Latent permutations]{\textbf{Latent permutations.} This illustrates how the latent permutation parameter $W$ controls how the blocks of the parameter $\mu$ are paired with the blocks of the data $X$. In this case $R=3$ and $T=2$ and the permutations follow the matrix in \eqref{eq:A-ex}.} \label{fig:polarity-graph}
\end{figure}

Let $Z \in \mathcal{F}$ be a latent variable indicating which cluster component or ``part'' and let $W \in \mathcal{P}$ denote the latent permutation. Then, parts are assumed to be generated from the following distribution
\begin{align*}
    (Z, W) &\sim \Cat{\pi}, \\
    X_{w',d} \mid Z=z, W=w &\sim \Bern{ \mu_{z, A({w,w'}), d}}, \quad\quad \forall w', d 
\end{align*}
where $\mu \in [0, 1]^{F \times P \times D}$ represents the parts model and $\pi \in [0, 1]^{F \times P}$ the joint priors over $(Z, W)$.
As we can see, the permutation parameter $W$ dictates which segment of $\mu$ a block of $X$ is generated from. An illustration of this is seen in fig.~\ref{fig:polarity-graph}.

\section{Expectation Maximization}
The
parameters $\theta = (\pi, \mu)$ are trained using the Expectation Maximization
(EM) algorithm.
Starting with the E-step:
\begin{align*}
    Q(\theta , \theta^\old) &= \mathbb{E}_{\vZ, \vW \mid \vX; \theta^\old} [ \log \Pr(\vX, \vZ, \vW ; \theta) ] \\
    &= \sum_n \mathbb{E}_{Z^{(n)}, W^{(n)} \mid X^{(n)}; \theta^\old} [ \log \Pr(X^{(n)}, Z^{(n)}, W^{(n)} ; \theta) ] \\
    &= \sum_{n,Z,W}  \underbrace{\Pr(Z, W \mid X^{(n)} ;\theta^\old)}_{\gamma_{n,Z,W}} \log \Pr(X^{(n)}, Z, W ; \theta)
\end{align*}
The responsibilities, $\gamma \in [0, 1]^{N \times F \times P}$, are computed as follows
\begin{align*}
    \gamma^{(n)}_{Z,W} &= \Pr(Z, W \mid X^{(n)}; \theta^\old) \\
        &= \frac {\Pr(X^{(n)} \mid Z, W; \mu^\old) \Pr(Z,W;  \pi^\old)}{ \sum_{Z',W'} \Pr(X^{(n)} \mid Z', W'; \mu^\old) \Pr(Z', W';  \pi^\old)} 
\end{align*}
Introducing unnormalized responsibilities $\gamma'$ and expanding the probabilities gives
\begin{empheq}[box={\mybluebox[1pt]}]{align*}
    \gamma^{(n)}_{z,w}    = \frac{\ungamma^{(n)}_{z,w}}{ \displaystyle\sum_{z',w'} \ungamma^{(n)}_{z',w'}}, \qquad
     \quad\ungamma^{(n)}_{z,w} = \pi_{z,w} \prod_{w',d} p(X^{(n)}_{A(w,w'),d}; \mu_{z,w',d}),
\end{empheq}
where, $p(X; \mu) = \mu^X (1 - \mu)^{1 - X}$. When implementing this, it is better to calculate the log of the unnormalized responsibilities and use the log-sum-exp trick to avoid underflow.
The complete log-likelihood is given by
\begin{align*}
    \log \Pr(X, Z, W ; \theta)
    &=
    \log \Pr(X \mid Z, W ; \mu) \Pr(Z, W ; \pi) \\
    &=\log \pi_{Z,W} + \sum_{W',d} \log p(X_{A({W,W'}),d}; \mu_{Z,W',d}),
\end{align*}
where we note that $\log p(x; \mu) = x \log \mu + (1 - x) \log(1 - \mu)$.

The M-step is
\[
    \theta^\new = \argmax_\theta Q(\theta, \theta^\old)
\]
Starting with $\pi$, this follows the same update as a regular Bernoulli EM. This is given by adding a Lagrange multiplier that ensures that $\sum_{z',w'} \pi_{z',w'} = 1$, deriving and setting the result to zero:
\begin{empheq}[box={\mybluebox[1pt]}]{align*}
    \pi^\new_{z,w} = \frac {\sum_n \gamma^{(n)}_{z,w}}{N}.
\end{empheq}
For $\mu$, we have 
\begin{align*}
    \frac{\partial Q(\theta, \theta^\old) }{\partial \mu_{f,w,d}} =
        \sum_{n,w'}\gamma^{(n)}_{f,w'}  \left[   
        \frac{X^{(n)}_{A(w, w'),d}}{\mu_{f,w,d}} - 
        \frac{1 - X^{(n)}_{A(w, w'),d}}{1 - \mu_{f,w,d}} 
            \right].
\end{align*}
Setting this to zero and re-arranging yields
\begin{empheq}[box={\mybluebox[1pt]}]{align*}
    \mu^\new_{f,w,d} = \frac{\sum_{n,w'} \gamma^{(n)}_{f,w'} X^{(n)}_{A(w, w'),d}}{\sum_{n,w'} \gamma^{(n)}_{f,w'}}
\end{empheq}

\section{Coding Features}

Once the parameters $\mu$ have been determined, we code a certain feature over
a patch by finding the maximum likelihood part over it. If we were true
to the model, we would set 
\[
    (Z^\mathrm{(ML)}, W^\mathrm{(ML)}) = \argmax_{Z,W} \Pr(X \mid Z, W; \mu),
\]
where $X$ is arranged as in \eqref{eq:training-X}. However, this requires the tedious and computationally intensive task of rotating and polarity-flipping the patch. Instead, we rely on the nearly perfect correlation between the transformed patches and match only the original patch to one of the transformations. This is equivalent of taking the parts model and considering it to be $\mu \in [0, 1]^{FP \times D}$, where $FP$ now is the number of parts. In this formulation, the coding of the parts is identical to the original model without a latent permutation parameter. 


\include{apx2-training}

\makebibliography


\begin{thebibliography}{100}

\bibitem{tensorflow2015-whitepaper-full}
Mart\'{\i}n Abadi, Ashish Agarwal, Paul Barham, Eugene Brevdo, Zhifeng Chen,
  Craig Citro, Greg~S. Corrado, Andy Davis, Jeffrey Dean, Matthieu Devin,
  Sanjay Ghemawat, Ian Goodfellow, Andrew Harp, Geoffrey Irving, Michael Isard,
  Yangqing Jia, Rafal Jozefowicz, Lukasz Kaiser, Manjunath Kudlur, Josh
  Levenberg, Dan Man\'{e}, Rajat Monga, Sherry Moore, Derek Murray, Chris Olah,
  Mike Schuster, Jonathon Shlens, Benoit Steiner, Ilya Sutskever, Kunal Talwar,
  Paul Tucker, Vincent Vanhoucke, Vijay Vasudevan, Fernanda Vi\'{e}gas, Oriol
  Vinyals, Pete Warden, Martin Wattenberg, Martin Wicke, Yuan Yu, and Xiaoqiang
  Zheng.
\newblock {TensorFlow}: Large-scale machine learning on heterogeneous systems,
  2015.
\newblock Software available from tensorflow.org.

\bibitem{amit2007pop}
Yali Amit and Alain Trouv{\'e}.
\newblock Pop: Patchwork of parts models for object recognition.
\newblock {\em International Journal of Computer Vision}, 75(2):267--282, 2007.

\bibitem{argyriou2008convex}
Andreas Argyriou, Theodoros Evgeniou, and Massimiliano Pontil.
\newblock Convex multi-task feature learning.
\newblock {\em Machine Learning}, 73(3):243--272, 2008.

\bibitem{ba2014deep}
Jimmy Ba and Rich Caruana.
\newblock Do deep nets really need to be deep?
\newblock In {\em Advances in Neural Information Processing Systems (NIPS)},
  pages 2654--2662, 2014.

\bibitem{ballard1987modular}
Dana~H Ballard.
\newblock Modular learning in neural networks.
\newblock In {\em AAAI}, pages 279--284, 1987.

\bibitem{bearman2016point}
Amy Bearman, Olga Russakovsky, Vittorio Ferrari, and Li~Fei-Fei.
\newblock {What's the Point: Semantic Segmentation with Point Supervision}.
\newblock In {\em European Conference on Computer Vision (ECCV)}, 2016.

\bibitem{bengio2007greedy}
Yoshua Bengio, Pascal Lamblin, Dan Popovici, Hugo Larochelle, et~al.
\newblock Greedy layer-wise training of deep networks.
\newblock {\em Advances in Neural Information Processing Systems (NIPS)},
  19:153, 2007.

\bibitem{bengio2014deep}
Yoshua Bengio, Eric Thibodeau-Laufer, Guillaume Alain, and Jason Yosinski.
\newblock Deep generative stochastic networks trainable by backprop.
\newblock In {\em International Conference on Machine Learning (ICML)}, pages
  226--234, 2014.

\bibitem{bernstein2005part}
Elliot~Joel Bernstein and Yali Amit.
\newblock Part-based statistical models for object classification and
  detection.
\newblock In {\em Conference on Computer Vision and Pattern Recognition
  (CVPR)}, volume~2, pages 734--740. IEEE, 2005.

\bibitem{BST:CVPR:2015}
Gedas Bertasius, Jianbo Shi, and Lorenzo Torresani.
\newblock Deepedge: A multi-scale bifurcated deep network for top-down contour
  detection.
\newblock In {\em Conference on Computer Vision and Pattern Recognition
  (CVPR)}, 2015.

\bibitem{bucilua2006model}
Cristian Bucilu\v{a}, Rich Caruana, and Alexandru Niculescu-Mizil.
\newblock Model compression.
\newblock In {\em Proceedings of the 12th ACM SIGKDD international conference
  on Knowledge discovery and data mining}, pages 535--541. ACM, 2006.

\bibitem{caruana1998multitask}
Rich Caruana.
\newblock Multitask learning.
\newblock In {\em Learning to learn}, pages 95--133. Springer, 1998.

\bibitem{charpiat2008automatic}
Guillaume Charpiat, Matthias Hofmann, and Bernhard Sch{\"o}lkopf.
\newblock Automatic image colorization via multimodal predictions.
\newblock In {\em European Conference on Computer Vision (ECCV)}, 2008.

\bibitem{chen2011integrating}
Jianhui Chen, Jiayu Zhou, and Jieping Ye.
\newblock Integrating low-rank and group-sparse structures for robust
  multi-task learning.
\newblock In {\em Proceedings of the 17th ACM SIGKDD international conference
  on Knowledge discovery and data mining}, pages 42--50. ACM, 2011.

\bibitem{chen2014semantic}
Liang-Chieh Chen, George Papandreou, Iasonas Kokkinos, Kevin Murphy, and Alan~L
  Yuille.
\newblock Semantic image segmentation with deep convolutional nets and fully
  connected crfs.
\newblock In {\em International Conference on Learning Representations (ICLR)},
  2015.

\bibitem{chen2016training}
Tianqi Chen, Bing Xu, Chiyuan Zhang, and Carlos Guestrin.
\newblock Training deep nets with sublinear memory cost.
\newblock {\em arXiv preprint arXiv:1604.06174}, 2016.

\bibitem{cheng2016rifd}
Gong Cheng, Peicheng Zhou, and Junwei Han.
\newblock Rifd-cnn: Rotation-invariant and fisher discriminative convolutional
  neural networks for object detection.
\newblock In {\em Proceedings of the IEEE Conference on Computer Vision and
  Pattern Recognition}, pages 2884--2893, 2016.

\bibitem{cheng2015deep}
Zezhou Cheng, Qingxiong Yang, and Bin Sheng.
\newblock Deep colorization.
\newblock In {\em ICCV}, 2015.

\bibitem{chia2011semantic}
Alex Yong-Sang Chia, Shaojie Zhuo, Raj~Kumar Gupta, Yu-Wing Tai, Siu-Yeung Cho,
  Ping Tan, and Stephen Lin.
\newblock Semantic colorization with internet images.
\newblock {\em ACM Transactions on Graphics (TOG)}, 30(6), 2011.

\bibitem{abdiv}
Andrzej Cichocki, Sergio Cruces, and Shun-ichi Amari.
\newblock Generalized alpha-beta divergences and their application to robust
  nonnegative matrix factorization.
\newblock {\em Entropy}, 13(1):134--170, 2011.

\bibitem{coates2010analysis}
Adam Coates, Honglak Lee, and Andrew~Y Ng.
\newblock An analysis of single-layer networks in unsupervised feature
  learning.
\newblock In {\em Artificial Intelligence and Statistics Conference (AISTATS)},
  2011.

\bibitem{cohen2016group}
Taco~S Cohen and Max Welling.
\newblock Group equivariant convolutional networks.
\newblock In {\em International Conference on Machine Learning (ICML)}, 2016.

\bibitem{deshpande2015learning}
Aditya Deshpande, Jason Rock, and David Forsyth.
\newblock Learning large-scale automatic image colorization.
\newblock In {\em ICCV}, 2015.

\bibitem{dieleman2016exploiting}
Sander Dieleman, Jeffrey De~Fauw, and Koray Kavukcuoglu.
\newblock Exploiting cyclic symmetry in convolutional neural networks.
\newblock In {\em International Conference on Machine Learning (ICML)}, 2016.

\bibitem{doersch2015unsupervised}
Carl Doersch, Abhinav Gupta, and Alexei~A. Efros.
\newblock Unsupervised visual representation learning by context prediction.
\newblock In {\em International Conference on Computer Vision (ICCV)}, 2015.

\bibitem{contextpred}
Carl Doersch, Abhinav Gupta, and Alexei~A Efros.
\newblock Unsupervised visual representation learning by context prediction.
\newblock In {\em International Conference on Computer Vision (ICCV)}, pages
  1422--1430, 2015.

\bibitem{donahue2016adversarial}
Jeff Donahue, Philipp Kr{\"{a}}henb{\"{u}}hl, and Trevor Darrell.
\newblock Adversarial feature learning.
\newblock In {\em International Conference on Learning Representations (ICLR)},
  2017.

\bibitem{duchi2011adaptive}
John Duchi, Elad Hazan, and Yoram Singer.
\newblock Adaptive subgradient methods for online learning and stochastic
  optimization.
\newblock {\em Journal of Machine Learning Research (JMLR)},
  12(Jul):2121--2159, 2011.

\bibitem{dumoulin2016adversarially}
Vincent Dumoulin, Ishmael Belghazi, Ben Poole, Alex Lamb, Martin Arjovsky,
  Olivier Mastropietro, and Aaron Courville.
\newblock Adversarially learned inference.
\newblock In {\em International Conference on Learning Representations (ICLR)},
  2017.

\bibitem{pascal-voc-2012}
M.~Everingham, L.~Van~Gool, C.~K.~I. Williams, J.~Winn, and A.~Zisserman.
\newblock The {PASCAL} {V}isual {O}bject {C}lasses {C}hallenge 2012 {(VOC2012)}
  {R}esults.
\newblock
  http://www.pascal-network.org/challenges/VOC/voc2012/workshop/index.html.

\bibitem{evgeniou2004regularized}
Theodoros Evgeniou and Massimiliano Pontil.
\newblock Regularized multi--task learning.
\newblock In {\em Proceedings of the tenth ACM SIGKDD international conference
  on Knowledge discovery and data mining}, pages 109--117. ACM, 2004.

\bibitem{farabet2013learning}
Clement Farabet, Camille Couprie, Laurent Najman, and Yann LeCun.
\newblock Learning hierarchical features for scene labeling.
\newblock {\em IEEE Transactions on Pattern Analysis and Machine Intelligence
  (PAMI)}, 35(8), 2013.

\bibitem{farneback2003two}
Gunnar Farneb{\"a}ck.
\newblock Two-frame motion estimation based on polynomial expansion.
\newblock {\em Image analysis}, pages 363--370, 2003.

\bibitem{GL:ACCV:2014}
Yaroslav Ganin and Victor~S. Lempitsky.
\newblock N\({}^{\mbox{4}}\)-fields: Neural network nearest neighbor fields for
  image transforms.
\newblock In {\em Asian Conference on Computer Vision (ACCV)}, 2014.

\bibitem{glorot2010understanding}
Xavier Glorot and Yoshua Bengio.
\newblock Understanding the difficulty of training deep feedforward neural
  networks.
\newblock In {\em Artificial Intelligence and Statistics Conference (AISTATS)},
  2010.

\bibitem{grandvalet2004semi}
Yves Grandvalet and Yoshua Bengio.
\newblock Semi-supervised learning by entropy minimization.
\newblock In {\em Advances in Neural Information Processing Systems (NIPS)},
  2004.

\bibitem{gruslys2016memory}
Audrunas Gruslys, Remi Munos, Ivo Danihelka, Marc Lanctot, and Alex Graves.
\newblock Memory-efficient backpropagation through time.
\newblock In {\em Advances in Neural Information Processing Systems (NIPS)},
  pages 4125--4133, 2016.

\bibitem{gupta2012image}
Raj~Kumar Gupta, Alex Yong-Sang Chia, Deepu Rajan, Ee~Sin Ng, and Huang
  Zhiyong.
\newblock Image colorization using similar images.
\newblock In {\em ACM international conference on Multimedia}, 2012.

\bibitem{hariharan2015hypercolumns}
B.~Hariharan, P.~Aberl\'aez an~R.~Girshick, and J.~Malik.
\newblock Hypercolumns for object segmentation and fine-grained localization.
\newblock {\em Conference on Computer Vision and Pattern Recognition (CVPR)},
  2015.

\bibitem{he2015msra}
Kaiming He, Xiangyu Zhang, Shaoqing Ren, and Jian Sun.
\newblock Delving deep into rectifiers: Surpassing human-level performance on
  imagenet classification.
\newblock {\em CoRR}, abs/1502.01852, 2015.

\bibitem{resnet}
Kaiming He, Xiangyu Zhang, Shaoqing Ren, and Jian Sun.
\newblock Deep residual learning for image recognition.
\newblock In {\em Conference on Computer Vision and Pattern Recognition
  (CVPR)}, pages 770--778, 2016.

\bibitem{hinton2011transforming}
Geoffrey Hinton, Alex Krizhevsky, and Sida Wang.
\newblock Transforming auto-encoders.
\newblock {\em Artificial Neural Networks and Machine Learning (ICANN)}, pages
  44--51, 2011.

\bibitem{hinton2015distilling}
Geoffrey Hinton, Oriol Vinyals, and Jeff Dean.
\newblock Distilling the knowledge in a neural network.
\newblock In {\em NIPS (workshop)}, 2015.

\bibitem{hinton2002cd}
Geoffrey~E Hinton.
\newblock Training products of experts by minimizing contrastive divergence.
\newblock {\em Neural computation}, 14(8):1771--1800, 2002.

\bibitem{hinton2007recognize}
Geoffrey~E Hinton.
\newblock To recognize shapes, first learn to generate images.
\newblock {\em Progress in brain research}, 165:535--547, 2007.

\bibitem{hinton2006reducing}
Geoffrey~E Hinton and Ruslan~R Salakhutdinov.
\newblock Reducing the dimensionality of data with neural networks.
\newblock {\em science}, 313(5786):504--507, 2006.

\bibitem{hinton1986learning}
Geoffrey~E Hinton and Terrence~J Sejnowski.
\newblock Learning and releaming in boltzmann machines.
\newblock {\em Parallel Distrilmted Processing}, 1, 1986.

\bibitem{huang2005adaptive}
Yi-Chin Huang, Yi-Shin Tung, Jun-Cheng Chen, Sung-Wen Wang, and Ja-Ling Wu.
\newblock An adaptive edge detection based colorization algorithm and its
  applications.
\newblock In {\em ACM international conference on Multimedia}, 2005.

\bibitem{hurri2003simple}
Jarmo Hurri and Aapo Hyv{\"a}rinen.
\newblock Simple-cell-like receptive fields maximize temporal coherence in
  natural video.
\newblock {\em Neural Computation}, 15(3):663--691, 2003.

\bibitem{hyvarinen2016unsupervised}
Aapo Hyvarinen and Hiroshi Morioka.
\newblock Unsupervised feature extraction by time-contrastive learning and
  nonlinear ica.
\newblock In {\em Advances in Neural Information Processing Systems (NIPS)},
  2016.

\bibitem{IizukaSIGGRAPH2016}
Satoshi Iizuka, Edgar Simo-Serra, and Hiroshi Ishikawa.
\newblock {Let there be Color!: Joint End-to-end Learning of Global and Local
  Image Priors for Automatic Image Colorization with Simultaneous
  Classification}.
\newblock {\em ACM Transactions on Graphics (SIGGRAPH)}, 35(4), 2016.

\bibitem{iizuka2016color}
Satoshi Iizuka, Edgar Simo-Serra, and Hiroshi Ishikawa.
\newblock {Let there be Color!: Joint End-to-end Learning of Global and Local
  Image Priors for Automatic Image Colorization with Simultaneous
  Classification}.
\newblock {\em ACM Transactions on Graphics (SIGGRAPH)}, 35(4), 2016.

\bibitem{ioffe2015batch}
Sergey Ioffe and Christian Szegedy.
\newblock Batch normalization: Accelerating deep network training by reducing
  internal covariate shift.
\newblock In {\em ICML}, 2015.

\bibitem{irony2005colorization}
Revital Irony, Daniel Cohen-Or, and Dani Lischinski.
\newblock Colorization by example.
\newblock In {\em Eurographics Symp. on Rendering}, 2005.

\bibitem{isola2015learning}
Phillip Isola, Daniel Zoran, Dilip Krishnan, and Edward~H Adelson.
\newblock Learning visual groups from co-occurrences in space and time.
\newblock {\em arXiv preprint arXiv:1511.06811}, 2015.

\bibitem{jayaraman2015slow}
Dinesh Jayaraman and Kristen Grauman.
\newblock Slow and steady feature analysis: higher order temporal coherence in
  video.
\newblock In {\em Conference on Computer Vision and Pattern Recognition
  (CVPR)}, 2016.

\bibitem{jia2014caffe}
Yangqing Jia, Evan Shelhamer, Jeff Donahue, Sergey Karayev, Jonathan Long, Ross
  Girshick, Sergio Guadarrama, and Trevor Darrell.
\newblock Caffe: Convolutional architecture for fast feature embedding.
\newblock {\em arXiv preprint arXiv:1408.5093}, 2014.

\bibitem{kingma2015adam}
Diederik~P. Kingma and Jimmy Ba.
\newblock Adam: {A} method for stochastic optimization.
\newblock In {\em International Conference on Learning Representations (ICLR)},
  2015.

\bibitem{kingma2014auto}
Diederik~P Kingma and Max Welling.
\newblock Auto-encoding variational bayes.
\newblock In {\em International Conference on Learning Representations (ICLR)},
  2014.

\bibitem{kokkinos2016ubernet}
Iasonas Kokkinos.
\newblock Ubernet: Training a universal convolutional neural network for low-,
  mid-, and high-level vision using diverse datasets and limited memory.
\newblock {\em arXiv preprint arXiv:1609.02132}, 2016.

\bibitem{krahenbuhl2016datadriven}
Philipp Kr{\"{a}}henb{\"{u}}hl, Carl Doersch, Jeff Donahue, and Trevor Darrell.
\newblock Data-dependent initializations of convolutional neural networks.
\newblock In {\em International Conference on Learning Representations (ICLR)},
  2016.

\bibitem{alexnet}
Alex Krizhevsky, Ilya Sutskever, and Geoffrey~E Hinton.
\newblock Imagenet classification with deep convolutional neural networks.
\newblock In {\em Advances in Neural Information Processing Systems (NIPS)},
  pages 1097--1105, 2012.

\bibitem{kullback1951information}
Solomon Kullback and Richard~A Leibler.
\newblock On information and sufficiency.
\newblock {\em The annals of mathematical statistics}, 22(1):79--86, 1951.

\bibitem{kumar2012learning}
Abhishek Kumar and Hal Daum{\'e}~III.
\newblock Learning task grouping and overlap in multi-task learning.
\newblock In {\em International Conference on Machine Learning (ICML)}, 2012.

\bibitem{larochelle2008classification}
Hugo Larochelle and Yoshua Bengio.
\newblock Classification using discriminative restricted boltzmann machines.
\newblock In {\em International Conference on Machine Learning (ICML)}, pages
  536--543. ACM, 2008.

\bibitem{larochelle2007empirical}
Hugo Larochelle, Dumitru Erhan, Aaron Courville, James Bergstra, and Yoshua
  Bengio.
\newblock An empirical evaluation of deep architectures on problems with many
  factors of variation.
\newblock In {\em International Conference on Machine Learning (ICML)}, pages
  473--480. ACM, 2007.

\bibitem{larsson2016learning}
Gustav Larsson, Michael Maire, and Gregory Shakhnarovich.
\newblock Learning representations for automatic colorization.
\newblock In {\em European Conference on Computer Vision (ECCV)}, 2016.

\bibitem{larsson2017colorproxy}
Gustav Larsson, Michael Maire, and Gregory Shakhnarovich.
\newblock Colorization as a proxy task for visual understanding.
\newblock In {\em Conference on Computer Vision and Pattern Recognition
  (CVPR)}, 2017.

\bibitem{larsson2017fractalnet}
Gustav Larsson, Michael Maire, and Gregory Shakhnarovich.
\newblock Fractalnet: Ultra-deep neural networks without residuals.
\newblock In {\em International Conference on Learning Representations (ICLR)},
  2017.

\bibitem{lee2011unsupervised}
Honglak Lee, Roger Grosse, Rajesh Ranganath, and Andrew~Y Ng.
\newblock Unsupervised learning of hierarchical representations with
  convolutional deep belief networks.
\newblock {\em Communications of the ACM}, 54(10):95--103, 2011.

\bibitem{levin2004colorization}
Anat Levin, Dani Lischinski, and Yair Weiss.
\newblock Colorization using optimization.
\newblock {\em ACM Transactions on Graphics (TOG)}, 23(3), 2004.

\bibitem{liang2008structure}
Percy Liang, Hal Daum{\'e}~III, and Dan Klein.
\newblock Structure compilation: trading structure for features.
\newblock In {\em International Conference on Machine Learning (ICML)}, pages
  592--599. ACM, 2008.

\bibitem{liu2015parsenet}
Wei Liu, Andrew Rabinovich, and Alexander~C Berg.
\newblock Parsenet: Looking wider to see better.
\newblock {\em arXiv preprint arXiv:1506.04579}, 2015.

\bibitem{long2015fully}
Jonathan Long, Evan Shelhamer, and Trevor Darrell.
\newblock Fully convolutional networks for semantic segmentation.
\newblock In {\em Conference on Computer Vision and Pattern Recognition
  (CVPR)}, 2015.

\bibitem{luan2007natural}
Qing Luan, Fang Wen, Daniel Cohen-Or, Lin Liang, Ying-Qing Xu, and Heung-Yeung
  Shum.
\newblock Natural image colorization.
\newblock In {\em Eurographics conference on Rendering Techniques}, 2007.

\bibitem{MYP:ACCV:2014}
Michael Maire, Stella~X. Yu, and Pietro Perona.
\newblock Reconstructive sparse code transfer for contour detection and
  semantic labeling.
\newblock In {\em Asian Conference on Computer Vision (ACCV)}, 2014.

\bibitem{mao2016image}
Xiaojiao Mao, Chunhua Shen, and Yu-Bin Yang.
\newblock Image restoration using very deep convolutional encoder-decoder
  networks with symmetric skip connections.
\newblock In {\em Advances in Neural Information Processing Systems (NIPS)},
  pages 2802--2810, 2016.

\bibitem{marszalek2009actions}
Marcin Marszalek, Ivan Laptev, and Cordelia Schmid.
\newblock Actions in context.
\newblock In {\em Computer Vision and Pattern Recognition, 2009. CVPR 2009.
  IEEE Conference on}, pages 2929--2936. IEEE, 2009.

\bibitem{mishkin2015all}
Dmytro Mishkin and Jiri Matas.
\newblock All you need is a good init.
\newblock {\em arXiv preprint arXiv:1511.06422}, 2015.

\bibitem{misra2016cross}
Ishan Misra, Abhinav Shrivastava, Abhinav Gupta, and Martial Hebert.
\newblock Cross-stitch networks for multi-task learning.
\newblock In {\em Proceedings of the IEEE Conference on Computer Vision and
  Pattern Recognition}, pages 3994--4003, 2016.

\bibitem{misra2016unsupervised}
Ishan Misra, C~Lawrence Zitnick, and Martial Hebert.
\newblock Unsupervised learning using sequential verification for action
  recognition.
\newblock 2016.

\bibitem{mobahi2009deep}
Hossein Mobahi, Ronan Collobert, and Jason Weston.
\newblock Deep learning from temporal coherence in video.
\newblock In {\em International Conference on Machine Learning (ICML)}, 2009.

\bibitem{morimoto2009automatic}
Yuji Morimoto, Yuichi Taguchi, and Takeshi Naemura.
\newblock Automatic colorization of grayscale images using multiple images on
  the web.
\newblock In {\em SIGGRAPH: Posters}, 2009.

\bibitem{mostajabi2015feedforward}
Mohammadreza Mostajabi, Payman Yadollahpour, and Gregory Shakhnarovich.
\newblock Feedforward semantic segmentation with zoom-out features.
\newblock In {\em Conference on Computer Vision and Pattern Recognition
  (CVPR)}, 2015.

\bibitem{nair2009implicit}
Vinod Nair and Geoffrey~E Hinton.
\newblock Implicit mixtures of restricted boltzmann machines.
\newblock In {\em Advances in Neural Information Processing Systems (NIPS)},
  pages 1145--1152, 2009.

\bibitem{nair2010rectified}
Vinod Nair and Geoffrey~E Hinton.
\newblock Rectified linear units improve restricted boltzmann machines.
\newblock In {\em International Conference on Machine Learning (ICML)}, pages
  807--814, 2010.

\bibitem{ng2011sparse}
Andrew Ng.
\newblock Sparse autoencoder.
\newblock {\em CS294A Lecture notes}, 72(2011):1--19, 2011.

\bibitem{noroozi2016jigsaw}
Mehdi Noroozi and Paolo Favaro.
\newblock Unsupervised learning of visual representations by solving jigsaw
  puzzles.
\newblock In {\em European Conference on Computer Vision (ECCV)}, 2016.

\bibitem{owens2016ambient}
Andrew Owens, Jiajun Wu, Josh~H. McDermott, William~T. Freeman, and Antonio
  Torralba.
\newblock Ambient sound provides supervision for visual learning.
\newblock In {\em European Conference on Computer Vision (ECCV)}, 2016.

\bibitem{pathak2016move}
Deepak Pathak, Ross~B. Girshick, Piotr Doll{\'{a}}r, Trevor Darrell, and
  Bharath Hariharan.
\newblock Learning features by watching objects move.
\newblock In {\em Conference on Computer Vision and Pattern Recognition
  (CVPR)}, 2017.

\bibitem{pathak2016context}
Deepak Pathak, Philipp Kr\"ahenb\"uhl, Jeff Donahue, Trevor Darrell, and Alexei
  Efros.
\newblock Context encoders: Feature learning by inpainting.
\newblock In {\em Conference on Computer Vision and Pattern Recognition
  (CVPR)}, 2016.

\bibitem{contextencoders}
Deepak Pathak, Philipp Krahenbuhl, Jeff Donahue, Trevor Darrell, and Alexei~A
  Efros.
\newblock Context encoders: Feature learning by inpainting.
\newblock In {\em Conference on Computer Vision and Pattern Recognition
  (CVPR)}, 2016.

\bibitem{patterson2014sun}
Genevieve Patterson, Chen Xu, Hang Su, and James Hays.
\newblock The sun attribute database: Beyond categories for deeper scene
  understanding.
\newblock {\em International Journal of Computer Vision}, 108(1-2), 2014.

\bibitem{qian1999momentum}
Ning Qian.
\newblock On the momentum term in gradient descent learning algorithms.
\newblock {\em Neural networks}, 12(1):145--151, 1999.

\bibitem{qu2006manga}
Yingge Qu, Tien-Tsin Wong, and Pheng-Ann Heng.
\newblock Manga colorization.
\newblock {\em ACM Transactions on Graphics (TOG)}, 25(3), 2006.

\bibitem{ranzato2014video}
Marc'Aurelio Ranzato, Arthur Szlam, Joan Bruna, Micha{\"{e}}l Mathieu, Ronan
  Collobert, and Sumit Chopra.
\newblock Video (language) modeling: a baseline for generative models of
  natural videos.
\newblock {\em arXiv preprint arXiv:1412.6604}, 2014.

\bibitem{rasmus2015semi}
Antti Rasmus, Mathias Berglund, Mikko Honkala, Harri Valpola, and Tapani Raiko.
\newblock Semi-supervised learning with ladder networks.
\newblock In {\em Advances in Neural Information Processing Systems (NIPS)},
  pages 3546--3554, 2015.

\bibitem{renyi1961measures}
Alfr{\'e}d R{\'e}nyi et~al.
\newblock On measures of entropy and information.
\newblock In {\em Proceedings of the Fourth Berkeley Symposium on Mathematical
  Statistics and Probability, Volume 1: Contributions to the Theory of
  Statistics}. The Regents of the University of California, 1961.

\bibitem{romero2014fitnets}
Adriana Romero, Nicolas Ballas, Samira~Ebrahimi Kahou, Antoine Chassang, Carlo
  Gatta, and Yoshua Bengio.
\newblock Fitnets: Hints for thin deep nets.
\newblock In {\em International Conference on Learning Representations (ICLR)},
  2015.

\bibitem{imagenet}
Olga Russakovsky, Jia Deng, Hao Su, Jonathan Krause, Sanjeev Satheesh, Sean Ma,
  Zhiheng Huang, Andrej Karpathy, Aditya Khosla, Michael Bernstein,
  Alexander~C. Berg, and Li~Fei-Fei.
\newblock {ImageNet Large Scale Visual Recognition Challenge}.
\newblock {\em International Journal of Computer Vision (IJCV)}, 115(3), 2015.

\bibitem{sajjadi2016semi}
Mehdi Sajjadi, Mehran Javanmardi, and Tolga Tasdizen.
\newblock Regularization with stochastic transformations and perturbations for
  deep semi-supervised learning.
\newblock {\em Advances in Neural Information Processing Systems (NIPS)}, 2016.

\bibitem{salakhutdinov2009deep}
Ruslan Salakhutdinov and Geoffrey Hinton.
\newblock Deep boltzmann machines.
\newblock In {\em Artificial Intelligence and Statistics}, pages 448--455,
  2009.

\bibitem{sapiro2005inpainting}
Guillermo Sapiro.
\newblock Inpainting the colors.
\newblock In {\em ICIP}, 2005.

\bibitem{SWWBZ:CVPR:2015}
Wei Shen, Xinggang Wang, Yan Wang, Xiang Bai, and Zhijiang Zhang.
\newblock Deepcontour: A deep convolutional feature learned by positive-sharing
  loss for contour detection.
\newblock In {\em Conference on Computer Vision and Pattern Recognition
  (CVPR)}, 2015.

\bibitem{shi2000normalized}
Jianbo Shi and Jitendra Malik.
\newblock Normalized cuts and image segmentation.
\newblock {\em Pattern Analysis and Machine Intelligence, IEEE Transactions
  on}, 22(8), 2000.

\bibitem{vgg16}
K.~Simonyan and A.~Zisserman.
\newblock Very deep convolutional networks for large-scale image recognition.
\newblock In {\em International Conference on Learning Representations (ICLR)},
  2015.

\bibitem{smolensky1986information}
Paul Smolensky.
\newblock Information processing in dynamical systems: Foundations of harmony
  theory.
\newblock Technical report, DTIC Document, 1986.

\bibitem{socher2011semi}
Richard Socher, Jeffrey Pennington, Eric~H Huang, Andrew~Y Ng, and
  Christopher~D Manning.
\newblock Semi-supervised recursive autoencoders for predicting sentiment
  distributions.
\newblock In {\em Proceedings of the conference on empirical methods in natural
  language processing}, pages 151--161. Association for Computational
  Linguistics, 2011.

\bibitem{sohn2012learning}
Kihyuk Sohn and Honglak Lee.
\newblock Learning invariant representations with local transformations.
\newblock 2012.

\bibitem{sohn2013pgbm}
Kihyuk Sohn, Guanyu Zhou, Chansoo Lee, and Honglak Lee.
\newblock Learning and selecting features jointly with point-wise gated
  boltzmann machines.
\newblock In {\em International Conference on Machine Learning (ICML)}, pages
  217--225, 2013.

\bibitem{srivastava2014dropout}
Nitish Srivastava, Geoffrey Hinton, Alex Krizhevsky, Ilya Sutskever, and Ruslan
  Salakhutdinov.
\newblock Dropout: A simple way to prevent neural networks from overfitting.
\newblock {\em Journal of Machine Learning Research (JMLR)}, 15(1), 2014.

\bibitem{srivastava2015video}
Nitish Srivastava, Elman Mansimov, and Ruslan Salakhutdinov.
\newblock Unsupervised learning of video representations using lstms.
\newblock In {\em International Conference on Machine Learning (ICML)}, 2015.

\bibitem{sykora2004unsupervised}
Daniel S{\`y}kora, Jan Buri{\'a}nek, and Ji{\v{r}}{\'\i} {\v{Z}}{\'a}ra.
\newblock Unsupervised colorization of black-and-white cartoons.
\newblock In {\em International symposium on Non-photorealistic animation and
  rendering}, 2004.

\bibitem{tai2005local}
Yu-Wing Tai, Jiaya Jia, and Chi-Keung Tang.
\newblock Local color transfer via probabilistic segmentation by
  expectation-maximization.
\newblock In {\em Conference on Computer Vision and Pattern Recognition
  (CVPR)}, 2005.

\bibitem{tieleman2012rmsprop}
Tijmen Tieleman and Geoffrey Hinton.
\newblock Lecture 6.5 - rmsprop, coursera: Neural networks for machine
  learning.
\newblock 2012.

\bibitem{tola2008fast}
Engin Tola, Vincent Lepetit, and Pascal Fua.
\newblock A fast local descriptor for dense matching.
\newblock In {\em Conference on Computer Vision and Pattern Recognition
  (CVPR)}, 2008.

\bibitem{tsaftaris2014novel}
Sotirios~A Tsaftaris, Francesca Casadio, Jean-Louis Andral, and Aggelos~K
  Katsaggelos.
\newblock A novel visualization tool for art history and conservation:
  Automated colorization of black and white archival photographs of works of
  art.
\newblock {\em Studies in Conservation}, 59(3), 2014.

\bibitem{ueda1996generalization}
Naonori Ueda and Ryohei Nakano.
\newblock Generalization error of ensemble estimators.
\newblock In {\em Neural Networks, 1996., IEEE International Conference on},
  volume~1, pages 90--95. IEEE, 1996.

\bibitem{urban2016dodeepsfollowup}
G.~{Urban}, K.~J. {Geras}, S.~{Ebrahimi Kahou}, O.~{Aslan}, S.~{Wang},
  R.~{Caruana}, A.~{Mohamed}, M.~{Philipose}, and M.~{Richardson}.
\newblock Do deep convolutional nets really need to be deep (or even
  convolutional)?
\newblock In {\em International Conference on Learning Representations (ICLR)},
  2017.

\bibitem{valpola2015neural}
Harri Valpola.
\newblock From neural pca to deep unsupervised learning.
\newblock {\em Advances in Independent Component Analysis and Learning
  Machines}, pages 143--171, 2015.

\bibitem{van1998independent}
J~Hans van Hateren and Dan~L Ruderman.
\newblock Independent component analysis of natural image sequences yields
  spatio-temporal filters similar to simple cells in primary visual cortex.
\newblock {\em Proceedings of the Royal Society of London B: Biological
  Sciences}, 265(1412):2315--2320, 1998.

\bibitem{vincent2008extracting}
Pascal Vincent, Hugo Larochelle, Yoshua Bengio, and Pierre-Antoine Manzagol.
\newblock Extracting and composing robust features with denoising autoencoders.
\newblock In {\em Proceedings of the 25th international conference on Machine
  learning}, pages 1096--1103. ACM, 2008.

\bibitem{vincent2010stacked}
Pascal Vincent, Hugo Larochelle, Isabelle Lajoie, Yoshua Bengio, and
  Pierre-Antoine Manzagol.
\newblock Stacked denoising autoencoders: Learning useful representations in a
  deep network with a local denoising criterion.
\newblock {\em Journal of Machine Learning Research (JMLR)},
  11(Dec):3371--3408, 2010.

\bibitem{wang2015unsupervised}
Xiaolong Wang and Abhinav Gupta.
\newblock Unsupervised learning of visual representations using videos.
\newblock In {\em International Conference on Computer Vision (ICCV)}, 2015.

\bibitem{welsh2002transferring}
Tomihisa Welsh, Michael Ashikhmin, and Klaus Mueller.
\newblock Transferring color to greyscale images.
\newblock {\em ACM Transactions on Graphics (TOG)}, 21(3), 2002.

\bibitem{worrall2017harmonic}
Daniel~E Worrall, Stephan~J Garbin, Daniyar Turmukhambetov, and Gabriel~J
  Brostow.
\newblock Harmonic networks: Deep translation and rotation equivariance.
\newblock In {\em Conference on Computer Vision and Pattern Recognition
  (CVPR)}, 2017.

\bibitem{wu2016resnet}
Zifeng Wu, Chunhua Shen, and Anton van~den Hengel.
\newblock High-performance semantic segmentation using very deep fully
  convolutional networks.
\newblock {\em CoRR}, abs/1604.04339, 2016.

\bibitem{xiao2010sun}
Jianxiong Xiao, James Hays, Krista~A Ehinger, Aude Oliva, and Antonio Torralba.
\newblock Sun database: Large-scale scene recognition from abbey to zoo.
\newblock In {\em Conference on Computer Vision and Pattern Recognition
  (CVPR)}, 2010.

\bibitem{xie2015holistically}
Saining Xie and Zhuowen Tu.
\newblock Holistically-nested edge detection.
\newblock In {\em International Conference on Computer Vision (ICCV)}, 2015.

\bibitem{xu2015weak}
Jia Xu, Alexander~G. Schwing, and Raquel Urtasun.
\newblock Learning to segment under various forms of weak supervision.
\newblock In {\em Conference on Computer Vision and Pattern Recognition
  (CVPR)}, 2015.

\bibitem{yang2016deep}
Yongxin Yang and Timothy Hospedales.
\newblock Deep multi-task representation learning: A tensor factorisation
  approach.
\newblock In {\em International Conference on Learning Representations (ICLR)},
  2017.

\bibitem{yatziv2006fast}
Liron Yatziv and Guillermo Sapiro.
\newblock Fast image and video colorization using chrominance blending.
\newblock {\em Image Processing, IEEE Transactions on}, 15(5), 2006.

\bibitem{yu2015multiscale}
Fisher Yu and Vladlen Koltun.
\newblock Multi-scale context aggregation by dilated convolutions.
\newblock In {\em International Conference on Learning Representations (ICLR)},
  2016.

\bibitem{zeiler2012adadelta}
Matthew~D Zeiler.
\newblock Adadelta: an adaptive learning rate method.
\newblock {\em arXiv preprint arXiv:1212.5701}, 2012.

\bibitem{zhang2016colorful}
Richard Zhang, Phillip Isola, and Alexei~A Efros.
\newblock Colorful image colorization.
\newblock In {\em European Conference on Computer Vision (ECCV)}, 2016.

\bibitem{zhang2017split}
Richard Zhang, Phillip Isola, and Alexei~A Efros.
\newblock Split-brain autoencoders: Unsupervised learning by cross-channel
  prediction.
\newblock In {\em Conference on Computer Vision and Pattern Recognition
  (CVPR)}, 2017.

\bibitem{zhao2015whatwhere}
Junbo Zhao, Michael Mathieu, Ross Goroshin, and Yann LeCun.
\newblock Stacked what-where auto-encoders.
\newblock {\em arXiv preprint arXiv:1506.02351}, 2015.

\bibitem{zhou2015deepscene}
Bolei Zhou, Aditya Khosla, {\`{A}}gata Lapedriza, Aude Oliva, and Antonio
  Torralba.
\newblock Object detectors emerge in deep scene cnns.
\newblock In {\em International Conference on Learning Representations (ICLR)},
  2015.

\bibitem{zhou2014learning}
Bolei Zhou, Agata Lapedriza, Jianxiong Xiao, Antonio Torralba, and Aude Oliva.
\newblock Learning deep features for scene recognition using places database.
\newblock In {\em Advances in Neural Information Processing Systems (NIPS)},
  pages 487--495, 2014.

\bibitem{zhou2017oriented}
Yanzhao Zhou, Qixiang Ye, Qiang Qiu, and Jianbin Jiao.
\newblock Oriented response networks.
\newblock In {\em Conference on Computer Vision and Pattern Recognition
  (CVPR)}, 2017.

\end{thebibliography}

\end{document}